\documentclass[10pt,twocolumn,reqno]{article}

\usepackage{cvpr}
\usepackage{times}
\usepackage{epsfig}
\usepackage{graphicx}
\usepackage{amsmath}
\usepackage{amssymb}
\usepackage{booktabs}
\usepackage{multirow}
\usepackage{caption}
\usepackage{placeins} 
\usepackage{float}
\usepackage{multirow}
\usepackage{colortbl}
\usepackage{array}
\usepackage{tabularx}
\usepackage{pdflscape}
\usepackage{adjustbox}
\usepackage{longtable}
\usepackage{lipsum}
\usepackage{comment}
\usepackage{subcaption}
\usepackage{adjustbox}
\usepackage{cite}

\usepackage[symbol]{footmisc}

\usepackage[table,xcdraw]{xcolor}

\usepackage{subfiles}

\graphicspath{
 {./images/}
}

\usepackage{rotating}
\usepackage{float}
\usepackage{adjustbox}
\usepackage{enumitem}

\setlist[1]{itemsep=-4pt, leftmargin=10pt}\usepackage{array}
\newcolumntype{P}[1]{>{\centering\arraybackslash}p{#1}}
\usepackage[pagebackref=true,breaklinks=true,letterpaper=true,colorlinks,bookmarks=false]{hyperref}

\def\cvprPaperID{****} 

\ifcvprfinal\fi

\cvprfinalcopy 

\def\cvprPaperID{****} 

\title{A-BDD: Leveraging Data Augmentations for Safe Autonomous Driving in Adverse Weather and Lighting}

\author{Felix Assion\thanks{Felix, Florens, and Nitin are the main contributors to this article, responsible for the conceptual ideas, data generation, and the experimental study. The remaining authors mainly contributed to the development of the visually appealing augmentation algorithms.},\hspace{0.5em} Florens Gressner\footnotemark[1],\hspace{0.5em} Nitin Augustine\footnotemark[1],
\\[0.2em] Jona Klemenc, Ahmed Hammam, Alexandre Krattinger, \\[0.2em] Holger Trittenbach, Anja Philippsen, Sascha Riemer
\\[1em] neurocat}

\begin{document}
\maketitle

\begin{abstract}
High-autonomy vehicle functions rely on machine learning (ML) algorithms to understand the environment.
Despite displaying remarkable performance in fair weather scenarios, perception algorithms are heavily affected by adverse weather and lighting conditions. To overcome these difficulties, ML engineers mainly rely on comprehensive real-world datasets. 
However, the difficulties in real-world data collection for critical areas of the operational design domain (ODD) often means synthetic data is required for perception training and safety validation.
Thus, we present A-BDD\footnotemark[2], a large set of over $60{,}000$ synthetically augmented images based on BDD100K that are equipped with semantic segmentation and bounding box annotations (inherited from the BDD100K dataset). The dataset contains augmented data for rain, fog, overcast and sunglare/shadow with varying intensity levels.
We further introduce novel strategies utilizing feature-based image quality metrics like FID and CMMD, which help identify useful augmented and real-world data for ML training and testing.
By conducting experiments on A-BDD, we provide evidence that data augmentations can play a pivotal role in closing performance gaps in adverse weather and lighting conditions.
\end{abstract}

\footnotetext[2]{The \textsc{A-BDD} dataset can be found via \texttt{https://doi.org/10.5281/zenodo.13301383}. 
For more details on the dataset, including usage guidelines and terms, please refer to the provided website.}
\section{Introduction} \label{sec: introduction}

The realization of autonomous driving (AD), in particular high and full driving automation (Level 4 \& 5), hinges on the development of robust ML-based perception algorithms. Recent DMV reports indicate that perception failure is still a core driver for advanced driver assistance systems (ADAS) disengagements \cite{DMV}. \newline
In past years, developers have tried to tackle these performance insufficiencies by incorporating ever-growing, diverse image datasets during training and testing of ML components \cite{Geiger2012CVPR, Cordts2016Cityscapes, caesar2020nuscenes, sun2020waymo}.
However, existing annotated real-world datasets lack sufficient data for critical ODD scenarios. Traditional data collection approaches struggle to capture the `long tail' of the data distribution due to the lack of controllability of the ego vehicle's environment \cite{chen2024endtoendautonomousdrivingchallenges, hallgarten2024vehiclemotionplanninggeneralize}.
Outlier scenarios, like extreme weather and lighting conditions, are heavily underrepresented in state-of-the-art automotive datasets, which, in the end, leads to the aforementioned perception vulnerabilities \cite{ACDC2021}. 
At the same time, detecting critical outlier scenarios within the vast volume of raw fleet data remains challenging, complicating efforts to address the limitations of existing annotated datasets \cite{zhang2021finding}. \newline 

As a consequence, researchers and practitioners increasingly rely on synthetic data to train, test and validate perception models \cite{burdorf2022reducingrealworlddata, tremblay2018trainingdeepnetworkssynthetic, lu2024machinelearningsyntheticdata}.
Fully-synthetic data, generated by simulation engines, has become an integral part in Software-in-the-Loop (SiL) and Hardware-in-the-Loop (HiL) testing. This data type is poised to become even more critical, with its utilization expected to expand significantly in ML training \cite{pfeffer2019tradeoff}. Yet, in the context of safety-critical applications, thorough method validity argumentation remains a challenge for simulation engines \cite{schwalbe2020structuring}. Rigorous strategies and experiments are required to demonstrate that the fully-synthetic datasets adequately reflect the true real-world data distribution. \newline

\begin{figure*}[htbp]
    \centering
    \captionsetup{skip=0.5\abovecaptionskip, font=footnotesize} 

    \begin{subfigure}{0.3333\textwidth}
        \centering
        \caption*{BDD100k - Clear}
        \includegraphics[width=\linewidth]{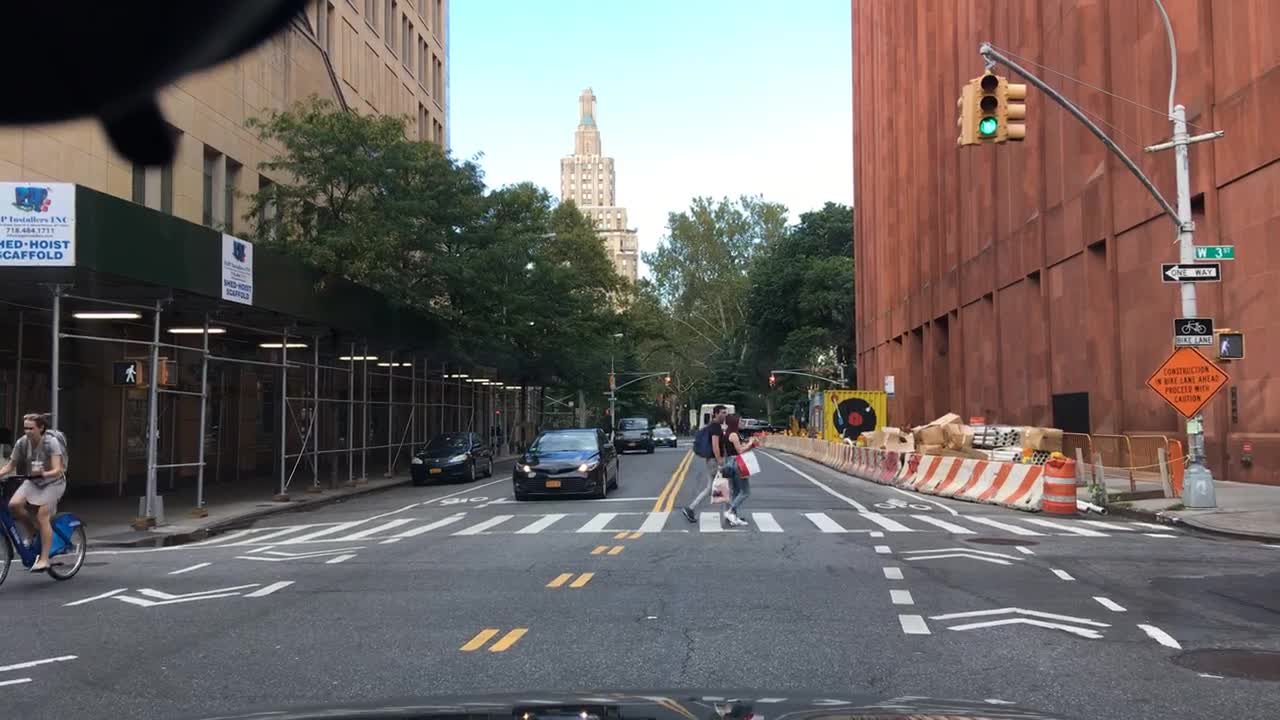}
    \end{subfigure}%
    \begin{subfigure}{0.33333\textwidth}
        \centering
        \caption*{A-BDD - Rain}
        \includegraphics[width=\linewidth]{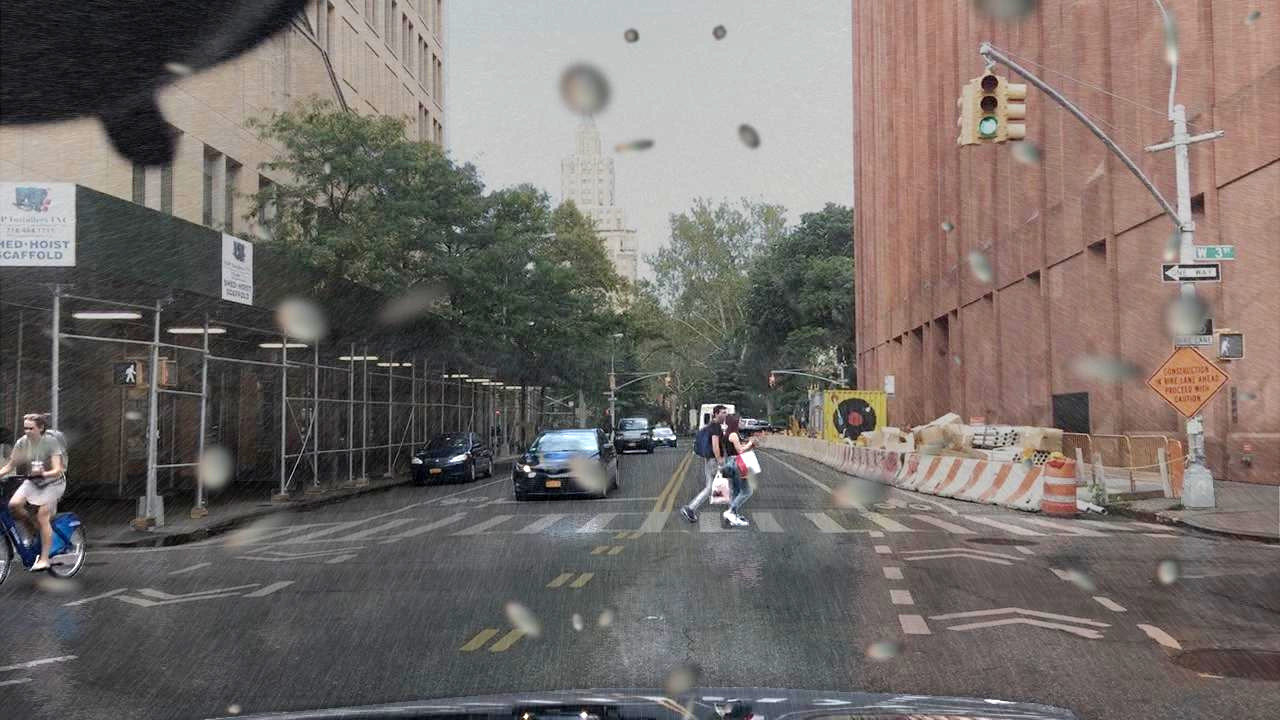}
    \end{subfigure}%
    \begin{subfigure}{0.33333\textwidth}
        \centering
        \caption*{BDD100k - Rain}
        \includegraphics[width=\linewidth]{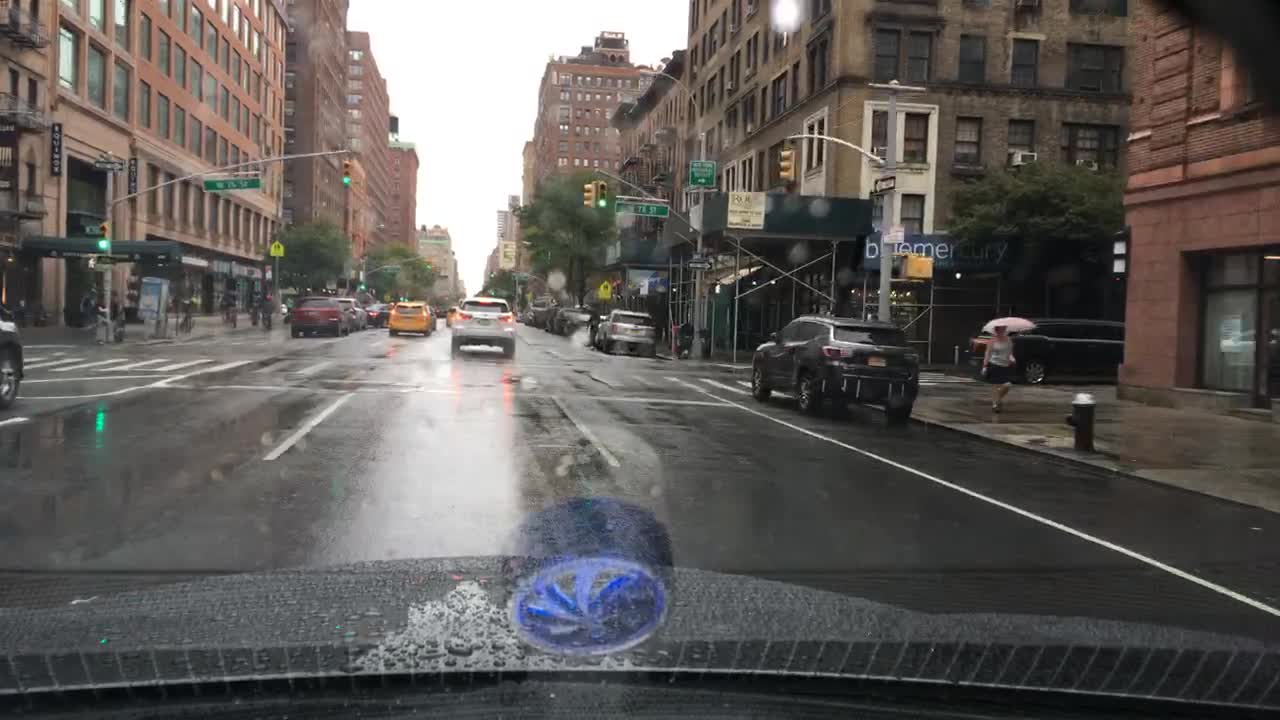}
    \end{subfigure}%

    \begin{subfigure}{0.33333\textwidth}
        \centering
        \includegraphics[width=\linewidth]{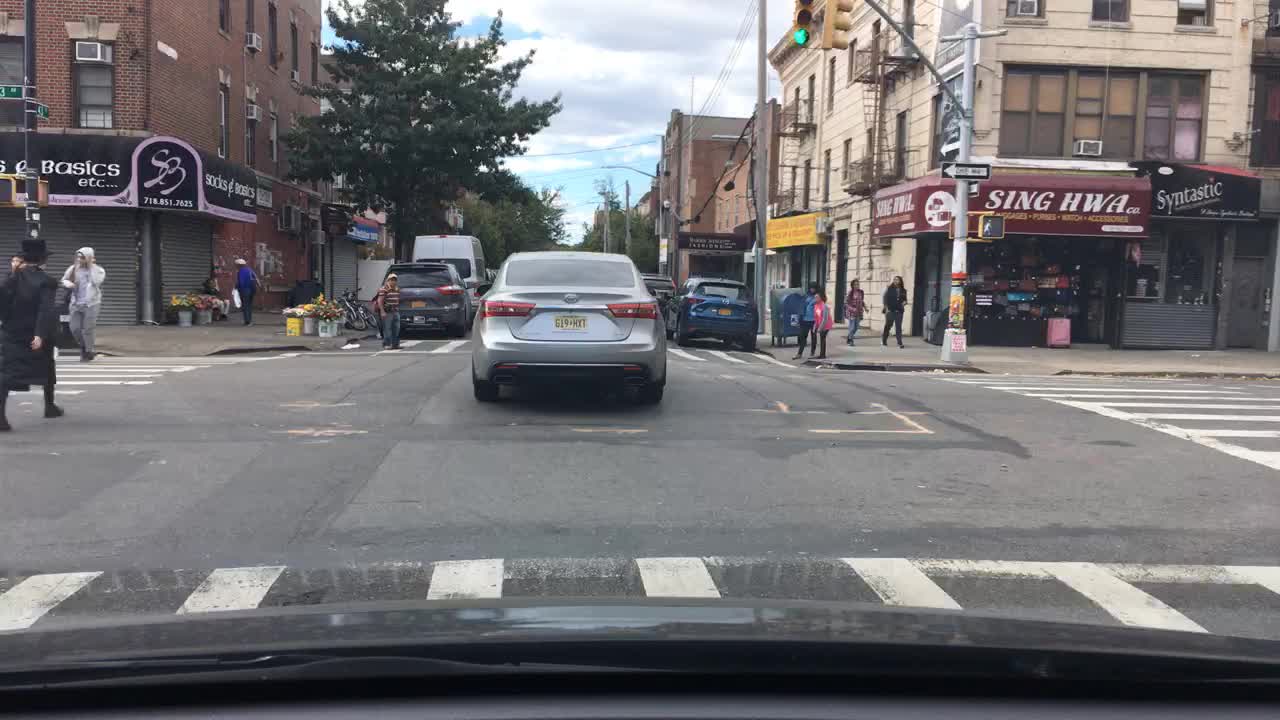}
        \caption*{BDD100k - Clear}
    \end{subfigure}%
    \begin{subfigure}{0.33333\textwidth}
        \centering
        \includegraphics[width=\linewidth]{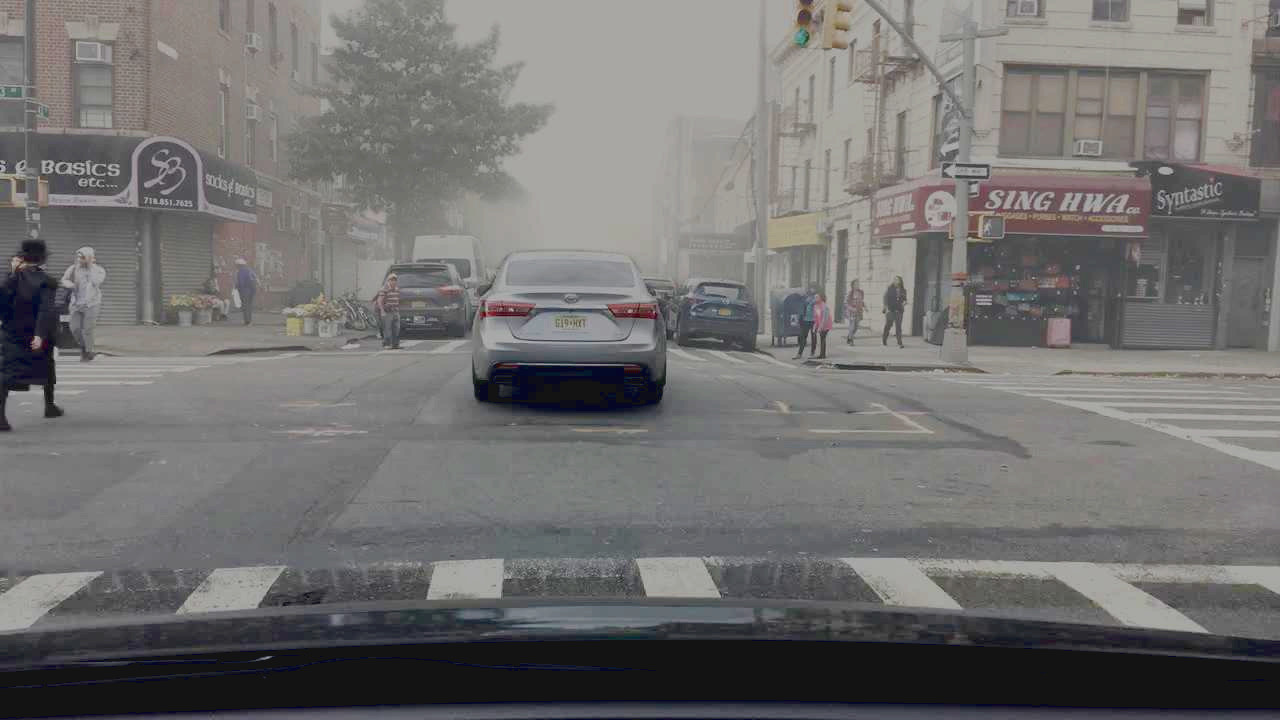}
        \caption*{A-BDD - Fog}
    \end{subfigure}%
    \begin{subfigure}{0.33333\textwidth}
        \centering
        \includegraphics[width=\linewidth]{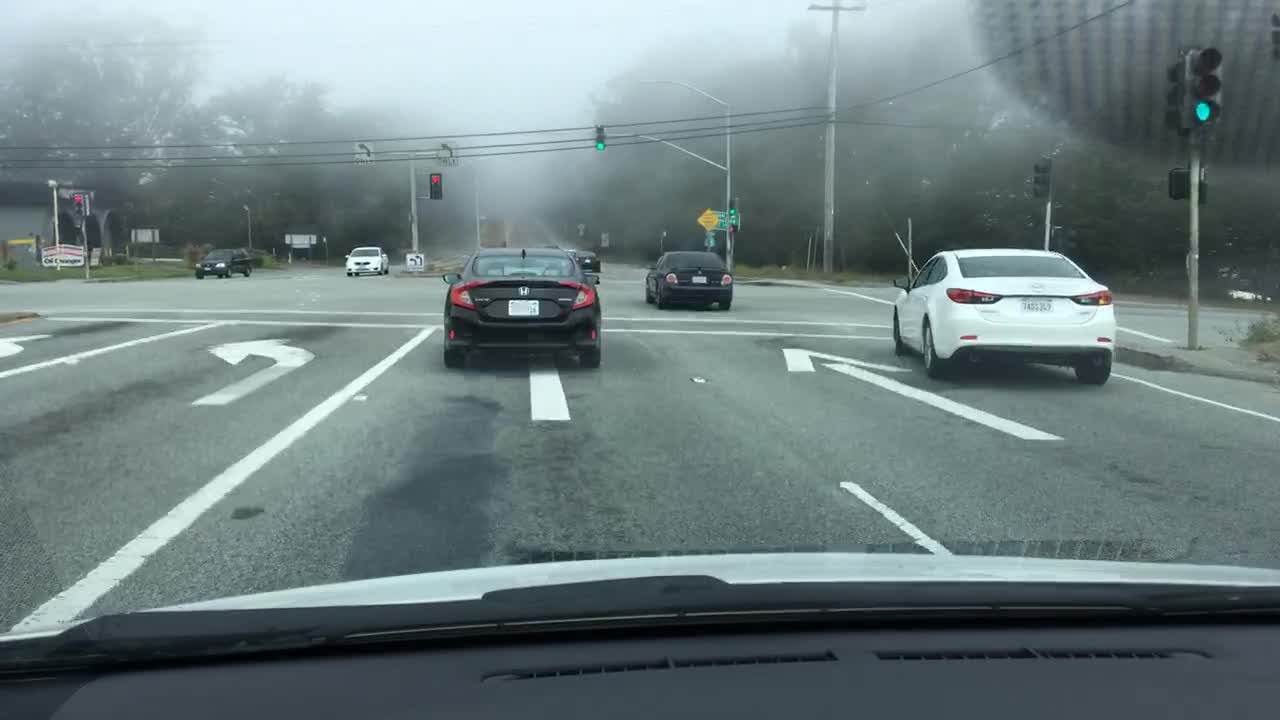}
        \caption*{BDD100k - Fog}
    \end{subfigure}

    \vspace{0.2cm}
    \caption{Example comparison between real-world data from \textsc{BDD100K} and augmented data from \textsc{A-BDD}. The first column presents reference fair weather images from \textsc{BDD100K}, while the second column shows corresponding augmented images from \textsc{A-BDD}. To emphasize the visual similarity to real-world rain and fog images from \textsc{BDD100K}, sample trigger data is included in the third column.}
    \label{fig:reference_images}
\end{figure*}

Between these two extremes, real-world and fully-synthetic data, there lives a third data type, one which has not yet been used to its full potential. Data augmentations are techniques used for expanding the size and diversity of datasets by applying image transformations to real-world data points. Simple data augmentations, like adding Gaussian noise, rotations and cropping, are ubiquitous in ML model training. These simple augmentations function as regularization mechanisms, which help models generalize better to unseen data \cite{krizhevsky2012imagenet, cubuk2018autoaugment}. \newline Partially related to these simple augmentations is the research area of adversarial robustness \cite{biggio2013evasion, szegedy2014intriguing, klemenc2023selecting}. Here, researchers develop adversarial attacks, which are optimization-based augmentation techniques, to exploit the brittleness of state-of-the-art neural networks.
Overall, adversarial attacks have received only limited attention within the automotive industry, mostly due to the belief that adversarial examples do not stem from realistic threat models.
Current developments in standardization and regulation, like the EU AI Act \cite{EUAIAct, EUAIActProposal} and the ISO PAS 8800 \cite{ISOPAS8800}, foreshadow that this particular area of ML robustness will see a new, well-deserved surge of interest across various industry verticals. \newline
More complex image augmentations \cite{Musat_2021_ICCV, Sakaridis_2018}, which mimic challenging safety-related ODD scenarios, like adverse weather and lighting conditions, do not yet see broad application in perception development. We think that this is due to a variety of reasons: Firstly, the research on these optical model-based, or more and more often also generative AI-based \cite{li2021weather}, image augmentations is still in an early phase, which implies that ML engineers only have limited access to these kind of techniques and related augmented datasets.  
Secondly, similar to the situation of simulation engines, existing augmentations often lack solid evidence that they can effectively close performance gaps on real-world scenarios, as well as that they can be part of safety argumentation for ML-based components within ADAS/AD systems. \newline

To tackle these limitations, we present \textsc{A-BDD}, an enriched version of the \textsc{BDD100K} \cite{Yu2018BDD100K} dataset (see Figure \ref{fig:reference_images}). \textsc{BDD100K} is one of the largest and most diverse available multitask learning datasets for image recognition. However, when focusing on the crucial task of semantic segmentation, \textsc{BDD100K} lacks the necessary variation of weather and lightning conditions. For example, we find less than $100$ rainy images and less than $20$ foggy images among the semantic segmentation data. \newline
Most of the augmentation techniques used for generating \textsc{A-BDD} do not significantly alter or obscure objects, allowing us to retain the original annotations from the unaltered dataset. This is the first large-scale, publicly available augmented dataset, which offers data across various weather and lightning effects, as well as different intensity levels for each trigger condition. \newline
To underline the potential of augmentations in the context of perception development and to give guidance on how to argue for the validity of synthetic data, we further provide experimental results conducted on \textsc{A-BDD}. Apart from a high degree of visual realism, \textsc{A-BDD} is able to fool a weather classifier, which detects similar weather characteristics in both the augmented and real-world adverse weather data. Furthermore, we enhance semantic segmentation model performance by including \textsc{A-BDD} into training processes.

One must bear in mind that the usefulness of a specific augmented dataset for ML training and testing will always depend on the given computer vision use case and data distribution. Merely incorporating available augmented datasets without conducting thorough analysis is unlikely to yield satisfactory results.
Thus, we utilize and extend existing image quality metrics, like FID \cite{heusel2017gans} and CMMD \cite{jayasumana2020rethinking}, and showcase correlations that help identify promising subsets of \textsc{A-BDD}. In our experimental study, we observe that these feature-based image quality metrics, typically used for GAN evaluation, can give valuable guidance in training/test dataset design. \newline

In summary, our main contributions are the following:
\begin{itemize}
\item  We release \textsc{A-BDD}, a dataset consisting of $35$ augmented versions of $1{,}820$ \textsc{BDD100K} \cite{Yu2018BDD100K} images (i.e., a total of $63{,}700$ images) with semantic segmentation and object detection labels. This augmentation dataset features a diverse collection of adverse weather and lighting conditions, including rain, fog, overcast, and sunglare/shadow.
\item We calculate the FID and CMMD distances between \textsc{A-BDD} and real-world weather data from \textsc{BDD100K} and \textsc{ACDC} \cite{ACDC2021}. These distances are then used to benchmark \textsc{A-BDD} against the Albumentations toolbox \cite{Buslaev2020Albumentations} by comparing the scores obtained from the augmented data.
\item We train a multi-weather classifier on real-world adverse weather images scraped from the internet. We then fool the classifier by evaluating it on \textsc{A-BDD}, as well as on data from Albumentations. 
\item We introduce `contrastive' variants of FID and CMMD, referred to as C-FID and C-CMMD, capable of predicting the likelihood of fooling the multi-weather classifier with subsets of the \textsc{A-BDD} dataset.
\item We fine-tune \textsc{BDD100K} pretrained segmentation models with \textsc{A-BDD} and demonstrate performance improvements on real-world rain data from \textsc{ACDC}. 
\item We observe a significant negative correlation between FID/CMMD scores and fine-tuning performance gains. This insight allows the selection of (re-)training data based on the analysis of feature-based image quality metrics.
\end{itemize}

We would like to stress that the purpose of this publication is not to present new image transformation algorithms, but to rather provide access to a large number of augmented images. These augmented images can be used to benchmark and improve semantic segmentation and object detection models under demanding weather and lighting conditions. Moreover, the experiments conducted in this paper should inspire future methodological approaches aimed at extracting value from real or synthetic data in perception development.

\section{Related Work} \label{sec: related_work}
In this section, we first review datasets for driving scene understanding, followed by a brief overview of research related to image augmentations and image quality metrics.

\subsection{Image Datasets}
The progress of AD research is highly dependent on the availability and the quality of large image datasets. Available datasets differ significantly with respect to size, environmental conditions, annotations and sensor modalities.
Important milestones in this domain are KITTI \cite{Geiger2012CVPR} and Cityscapes \cite{Cordts2016Cityscapes}. KITTI is one of the first open source datasets that contained LiDAR point clouds alongside stereo camera and GPS localization data. However, it does not contain semantic segmentation annotations, which partially explains the success of the later published Cityscapes dataset. Cityscapes provides driving data from $50$ different German cities with pixel-level, instance-level and panoptic semantic annotations.
These two popular datasets predominantly contain images taken under fair weather conditions. \newline
To overcome this lack of environmental diversity, research groups started publishing datasets with weather-affected images. With $100K$ videos and a variety of supported computer vision tasks, BDD100K \cite{Yu2018BDD100K} is one of the largest and most diverse datasets for driving scene understanding. In particular, the impressive collection of images  featuring rain and snow weather conditions with bounding box annotations allow extensive benchmarking of object detection models. 
However, regarding pixel-level semantic annotations, the BDD100K dataset exhibits notable limitations, comprising merely $10K$ images, of which approximately $1K$ depict challenging environmental conditions such as night scenes, snow, fog, and rain. \newline
Much like the BDD100K dataset, the ZOD dataset \cite{ZOD2021} pushes the state-of-the-art for multimodal perception development. ZOD encompasses driving scenes captured across $14$ European countries, while providing image data reflecting various weather conditions and lighting scenarios. However, ZOD only comes with pixel-level annotations for lane markings, road paintings, and the ego road, which limits its potential in the context of segmentation model training. \newline
One of the most recent dataset publications focusing on adverse weather conditions for semantic segmentation is ACDC \cite{ACDC2021}. This dataset consists of roughly $4K$ images which are equally distributed between $4$ different trigger conditions (fog, night, rain and snow). Every adverse weather image comes with a reference image of the same scene under fair weather conditions. The reference image itself is not annotated, which makes training models and working with augmentation techniques on ACDC challenging. Yet, the ACDC dataset remains the most effective way to evaluate if weather and lighting effects degrade semantic segmentation performance.  

\subsection{Image Augmentations}

Computer vision models struggle with real-world distribution shifts \cite{Recht2019Do, Hendrycks2021Many}. Data augmentations can help improve out-of-distribution generalization, and have thus become a standard part of model training pipelines \cite{yang2023imagedataaugmentationdeep, Shorten2019ASurvey}. Most ML frameworks offer simple image augmentations (or, corruptions), like rotation, flipping, and scaling. In addition, the research community has developed several useful augmentation libraries \cite{Buslaev2020Albumentations, Mueller2019ImgAug} and datasets \cite{hendrycks2019benchmarking, Sakaridis_2018}, which can help with the robustness benchmarking of vision models \cite{Hendricks2018Augmentations}. Existing tools, like Albumentations \cite{Buslaev2020Albumentations} and imaug \cite{Mueller2019ImgAug}, extend ML frameworks by offering more diverse common corruptions (e.g., Gaussian noise, blur, low-lighting noise, compression).  \newline

There exists a close relationship between corruption and adversarial robustness \cite{ford2019adversarial}, i.e., improvements in one notion of robustness can transfer to the other. Over the last years, there has been research directed towards the exploration of adversarial attacks based on various adversary threat models. Gradient-based attacks, like FGSM and PGD, have proven capable of altering classification outputs in any desired manner \cite{Goodfellow2015Explaining, Madry2018Towards}. \newline
The inclusion of corrupted and adversarial data in model training, is still viewed as one of the most promising approaches to mitigate these robustness vulnerabilities \cite{Hendrycks2021AugMix}. However, there are no known augmentation methods that consistently improve robustness across different data distributions and out-of-distribution phenomena \cite{Hendrycks2021AugMix}. \newline

As a consequence, we observe the development of more complex data transformation methods linked to specific real-world scenarios (e.g., rain, snow, and fog). These methods are either based on style transfer, or make use of physical/optical models. Due to the recent advances in GAN and CycleGAN performance \cite{Musat_2021_ICCV}, style transfer can achieve a high degree of perceptual realism, but lacks traceability and controllability \cite{Zhu2017Unpaired}. For example, varying the intensity levels of weather and lighting conditions is crucial for a deeper understanding of existing ML performance gaps, however this level of adaptability is not attainable for common style transfer approaches. Furthermore, the black-box nature of GANs requires thorough quality assurance procedures for the generated synthetic data. Depending on the style transfer method, it is important to ensure that existing objects persist and maintain local consistency after the transformation of a fair-weather image. \newline 
Physical-based augmentations come with traceability and controllability benefits compared to style transfer, at the cost of rather complex augmentation pipelines. In \cite{Tremblay_2020}, the authors present a rain rendering pipeline, which makes use of a particles simulator and a raindrop appearance database. Every rain streak is projected individually onto the image, which introduces a significant computational overhead. Although the paper presents a sophisticated multi-step approach to tackle different effects of rain, there are still elements, like overcast sky, droplets on the lens, wet surface and puddles, which are not yet included in the augmentation method. \newline
Similarly, Sakaridis et al. \cite{Sakaridis_2018} apply fog augmentations to clear-weather images based on a well-understood optical model, one which has already seen application in image dehazing. The presented approach was used to create Foggy Cityscapes, an augmented dataset, which consists of $550$ foggy images with semantic annotations. The presented fog augmentation pipeline generates visually appealing results, but still misses minor details like adding an overcast effect, adaptive blur and halo effects around light sources.

\begin{figure*}[htbp]
    \centering
    \captionsetup{skip=0.5\abovecaptionskip, font=footnotesize} 
    \begin{subfigure}{0.5\textwidth}
        \centering
        \caption*{Original Image}
        \includegraphics[width=\linewidth]{augmentations/0a56c2e8-e46ca9b7/augmentation_src.jpg}
    \end{subfigure}%
    \hspace{0pt}%
    \vspace{0pt}%
    \begin{subfigure}{0.5\textwidth}
        \centering
        \caption*{Rain Composition}
        \includegraphics[width=\linewidth]{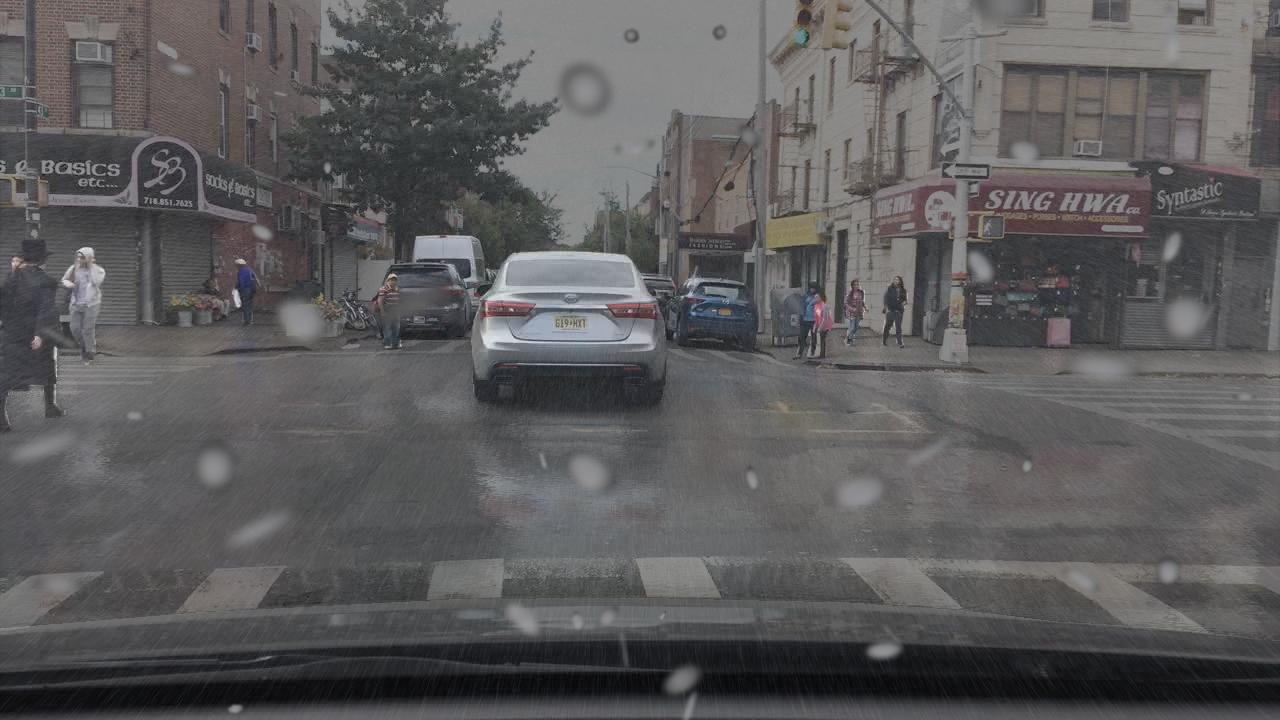}
    \end{subfigure}
    \hspace{0pt}%
    \vspace{0pt}%
    \begin{subfigure}{0.5\textwidth}
        \centering
        \includegraphics[width=\linewidth]{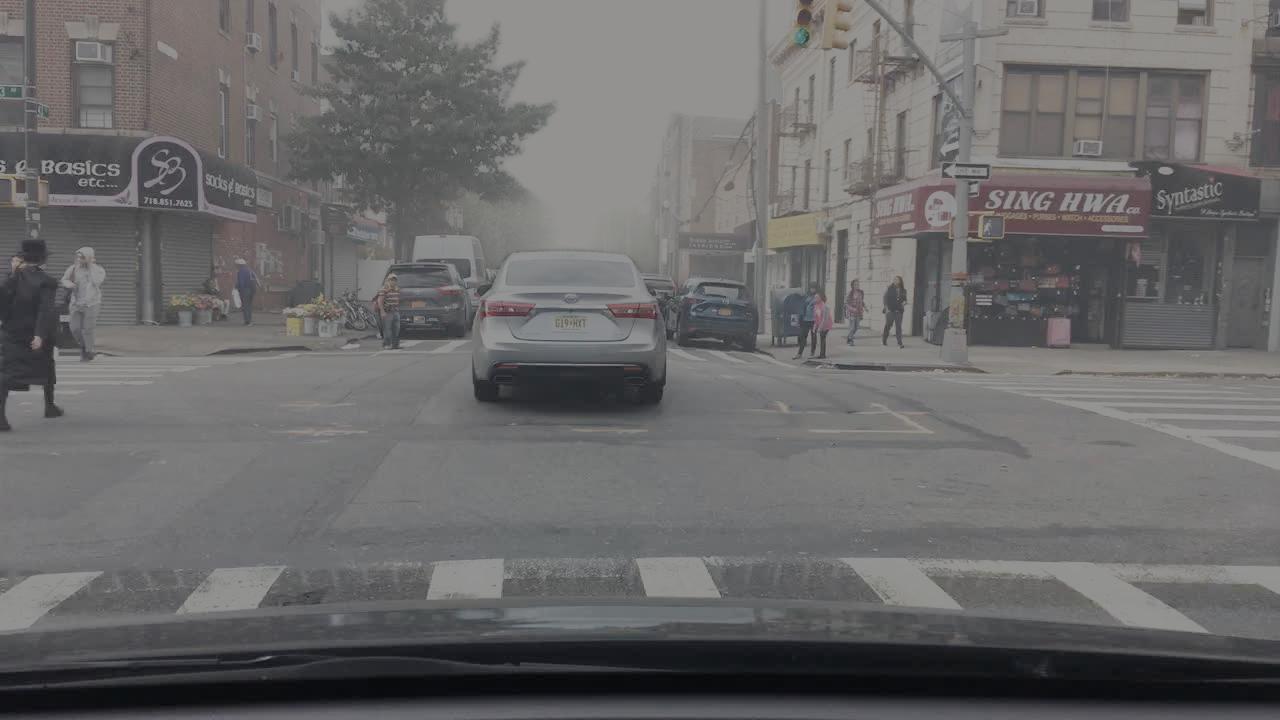}
        \caption*{Dense Fog}
    \end{subfigure}%
    \hspace{0pt}%
    \begin{subfigure}{0.5\textwidth}
        \centering
        \includegraphics[width=\linewidth]{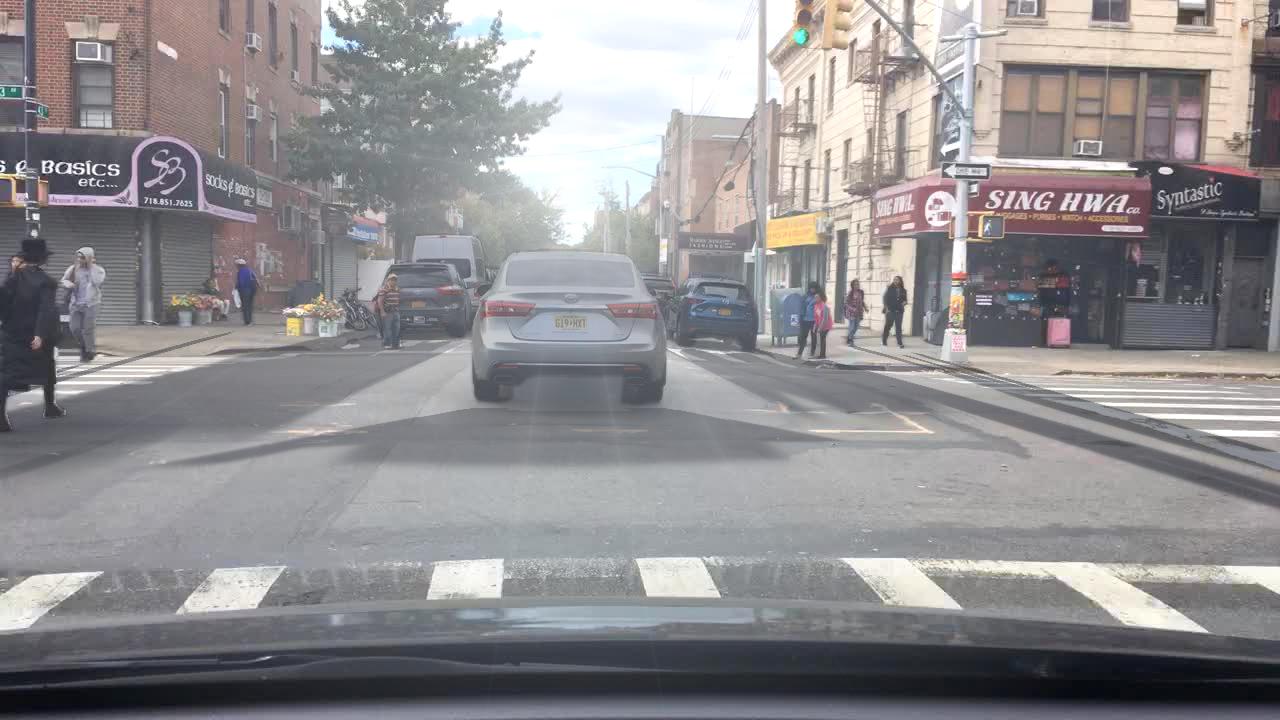}
        \caption*{Shadow \& Sunglare}
    \end{subfigure}
    \hspace{0pt}%
    \caption{Comparison of unaltered image from \textsc{BDD100K} and augmented images from \textsc{A-BDD}.}
    \label{fig:augmented_images}
\end{figure*}

\subsection{Image Quality Metrics} \label{sec: image_quality_metric}
The need to evaluate image quality of generated images has increased significantly with the success of generative models. 
In their seminal paper on Generative Adversarial Networks (GANs), Goodfellow et al.\@ \cite{goodfellow2014generativeadversarialnetworks} utilize Parzen window-based log-likelihood estimates as a means to assess the quality of generated data. However, they highlight the metric's susceptibility to high variance and dimensionality issues, advocating for the exploration of alternative approaches to address these limitations. \newline
In recent years, the Inception Score (IS) \cite{salimans2016improved} and the Fréchet Inception Distance (FID) \cite{heusel2017gans} have become the de-facto standards for image quality evaluation. Both metrics are based on a pretrained Inceptionv3 model, but differ in their algorithmic approach. \newline
The IS is calculated solely on the generated or augmented data and evaluates the image dataset with respect to quality and diversity. The metric is based on an expected KL divergence between marginal and conditional distributions, which are calculated with the help of the output of the underlying Inceptionv3 model. Thus, the IS metric is influenced significantly by the $1{,}000$ output classes of the Inceptionv3 model. A generated image is considered to be of `high quality' if the related output distribution under the Inceptionv3 model is clearly centered around one particular ImageNet class. This property, plus the fact that IS does not take real-world data into account, makes this score less applicable to our intended experiments.\newline
The feature-based FID score, on the other hand, places a multivariate normal distribution assumption on the penultimate layer of the Inceptionv3 model, and calculates the squared Fréchet distance of the real-world and the synthetic dataset with respect to these Inception feature embeddings. \newline
Recent research has pointed out that the FID score suffers from sample inefficiency, as well as its underlying normality assumption. Jayasumana et al. \cite{jayasumana2020rethinking} also show that FID may disagree with human judgment of image quality. Thus, the paper introduces CMMD, a sample efficient and distribution-free image quality metric, which estimates the Maximum Mean Discrepancy of the CLIP \cite{OpenAI2021CLIP} embeddings of two datasets.\newline

In this work, we make use of FID and CMMD to analyze \textsc{A-BDD}. Both metrics generate similar results across most augmentation sets, but differ with respect to required dataset sizes. Our focus lies in the usage of image quality metrics for selecting augmented data that aids in addressing performance deficiencies within perception algorithms.
For this, the capability of the metric to measure visual realism of augmented data is only a secondary objective. It is rather important that the given image quality metrics are able to discern whether the augmented data creates comparable activation patterns as the related real-world adverse weather condition. 
Hence, the disagreement of FID with human raters is not particularly relevant to our experiments.
\section{Augmented-BDD (A-BDD)} \label{sec: A-BDD}

The dataset is composed of $35$ subsets that replicate multiple adverse weather conditions with varying intensity levels. We provide augmented data for rainy, overcast and foggy weather, as well as sun-glared conditions with additional shadow effects. 
In the case of rain, the dataset includes multiple subsets, each representing different aspects of rainy weather, such as road reflections, water droplets on the camera lens, rain streaks, and combinations of these phenomena.\newline

The augmented data is derived from the images of \textsc{BDD100K}, which are annotated with semantic segmentation and bounding box labels. The segmentation training and validation data of \textsc{BDD100K} consists of $8K$ images. However, we ensure that we only augment data depicting daytime, fair weather conditions (i.e., with `clear' / `overcast' and `daytime' attributes), and exclude instances with unusable camera focus or the presence of reflections on the windshield. The daytime constraint mainly stems from limitations of our augmentation techniques, which have not been optimized for nighttime settings and would potentially generate unsatisfactory visual results in this context. \newline
Since only the \textsc{BDD100K} detection data comes with attribute tags, we intersect the bounding box and segmentation annotations, and filter for daytime, fair weather images. Subsequently, we conduct additional visual quality assurance to eliminate incorrectly tagged and unusable images, resulting in $1{,}820$ images forming the foundation of \textsc{A-BDD}. The same image basis was used for all augmentations, allowing for the comparison of distinct driving scenes under varying weather conditions (see Figure \ref{fig:augmented_images}). \newline
Our augmentation methods are custom implementations that build upon and extend the current state of the art. For some augmentation techniques, depth maps are also required. We use the DepthAnything \cite{yang2024depthanythingunleashingpower} model to generate depth maps for each of the $1{,}820$ images.  \newline

The augmentations are grouped into the following categories, each combining different techniques to simulate specific adverse weather and lightning conditions. The numbers in the names represent different intensity levels of the respective augmentation. These $7$ categories, each with $5$ intensity levels, ultimately result in the $35$ subsets of \textsc{A-BDD}:
\\[0.1em]
\begin{itemize}
    \item \textbf{Overcast} (\texttt{overcast\_<1-5>}): Adds a desaturation effect on the image and manipulates the sky to appear gray using the segmentation map.
    \item \textbf{Dense Fog} (\texttt{dense\_fog\_<1-5>}): Combines the overcast effect with a fog effect that uses the depth map of the scene to adapt the opacity and blur of the fog considering the distance to each object.
    \item \textbf{Shadow \& Sunglare} (\texttt{shadow\_sunglare\_<1-5>}): Uses the segmentation map to place a sun in the sky, adds saturation to the scene, places shadows on the road and shading to each object. The shadows are generated with the help of the segmentation of each object. The shape of the segmentation is warped using a homography based on the corners of the bounding box and their projection on the street (calculated with the sun's and object's position provided by the depth map).
    \item \textbf{Rain Streaks} (\texttt{rain\_streaks\_<1-5>}): Uses a particle system to generate rain streaks along with the overcast effect.
    \item \textbf{Wet Street \& Lens Droplets} (\texttt{wet\_street\_lens\_droplets\_<1-5>}): Combines the overcast effect with street reflections using the depth map to calculate reflection points. The intensity of the reflection and the roughness/reflectivity of the ground are parameterized. In these augmented subsets, the last two intensities also include lens droplets.
    \item \textbf{Puddles} (\texttt{puddles\_<1-5>}): Applies overcast and depth reflection effects to simulate puddles. The shape of the puddles is generated using Perlin Noise and is projected on the street with the help of the segmentation and the depth map of the street.
    \item \textbf{Rain Composition} (\texttt{rain\_composition\_<1-5>}): Combines overcast, rain streaks, and wet street effects with lens droplets and fog. Lens droplets are added with a certain transparency on the lens or windshield, inspired by visual inspection of \textsc{BDD100K}.
\end{itemize}

This dataset was created by iteratively selecting parameters for each augmentation building block, performing visual inspections on a small subset, and eventually applying the augmentations to all $1{,}820$ images. We did not optimize the augmentation parameters to achieve specific image quality metric or model fine-tuning results. The `intelligent' selection of augmentation parameters, beyond relying solely on visual inspection, is an area for future research. \newline
Apart from \textsc{A-BDD}, we also generated augmented versions of the $1{,}820$ images with the open-source tool Albumentations: \texttt{albu\_sun\_<1-6>}, \texttt{albu\_rain\_<1-3>}, and \texttt{albu\_fog\_<1-6>}. These alternative augmented sets are used for benchmarking in Section \ref{sec: Experimental Study}. \newline

In this work, we do not elaborate on the applied augmentation algorithms. The used methods predominantly consist of refined versions of optical model-based approaches (see Section \ref{sec: image_quality_metric}), supplemented with additional augmentation pipeline steps to incorporate weather-related artifacts such as overcast skies, droplets on the lens, street reflections, puddles, and shadows. We believe that the main complexity in this research field does not come from the development of new promising augmentation algorithms, but rather from the challenge of identifying and making use of the right - real or synthetic - data for a given perception use case.

\begin{table*}[htbp]
\centering
\captionsetup{skip=0.5\abovecaptionskip, font=footnotesize} 
\begin{minipage}[t]{0.48\linewidth}
  \centering
  \resizebox{\linewidth}{!}{
  \begin{tabular}{c|ccccc}
  \toprule
  BDD100K & Clear & Overcast & Fog & Rain & Snow\\
  \midrule
  Clear & 46.7 / 0.02 & 50.7 / 0.15 & - / 0.66 & 73.6 / 0.64 & 69.6 / 0.83 \\
  Overcast & 50.7 / 0.15 & 41.7 / 0.03 & - / 0.67 & 64.9 / 0.38 & 66.2 / 0.69 \\
  Fog & - / 0.66 & - / 0.67 & - / 0.29 & - / 0.6 & - / 1.21 \\
  Rain & 73.6 / 0.64 & 64.9 / 0.38 & - / 0.6 & 60.5 / 0.02 & 64.4 / 0.63 \\
  Snow & 69.6 / 0.83 & 66.2 / 0.69 & - / 1.21 & 64.4 / 0.63 & 62.9 / 0.02 \\
  \bottomrule
  \end{tabular}
  }
    \vspace{0.1cm}
  \caption{Cross product of FID/CMMD distances on \textsc{BDD100K} trigger data. The distances were calculated on 629 clear, 563 overcast, 43 fog, 500 rain, and 507 snow images. We do not report FID scores for fog because \textsc{BDD100K} lacks a sufficient number of foggy images for the FID metric to converge.}
  \label{tab:bdd_distance}
\end{minipage}\hfill
\begin{minipage}[t]{0.48\linewidth}
  \centering
  \resizebox{0.9\linewidth}{!}{
  \begin{tabular}{c|cccc}
  \toprule
  ACDC & Clear & Fog & Rain & Snow \\
  \midrule
  Clear & 28.2 / 0.03 & 92.7 / 2.26 & 83.8 / 0.86 & 88.2 / 1.49 \\
  Fog & 92.7 / 2.26 & 58.8 / 0.3 & 117.8 / 1.93 & 94.5 / 1.81 \\
  Rain & 83.8 / 0.86 & 117.8 / 1.93 & 68.2 / 0.31 & 86.6 / 1.18 \\
  Snow & 88.2 / 1.49 & 94.5 / 1.81 & 86.6 / 1.18 & 48.7 / 0.09 \\
  \bottomrule
  \end{tabular}
  }
      \vspace{0.1cm}
  \caption{Cross product of FID/CMMD distances on \textsc{ACDC} trigger data. The distances were calculated on 1802 clear, 1800 fog, 1800 rain, and 1520 snow images.}
  \label{tab:acdc_distance}
\end{minipage}
\end{table*}
\section{Experimental Study} \label{sec: Experimental Study}

In our experimental study, we illustrate the value of \textsc{A-BDD}, as well as inspire novel strategies leveraging feature-based image quality metrics for the evaluation and selection of augmented data for ML training and testing. \newline
We will give insights into the differences between feature embeddings coming from real-world adverse weather and weather-augmented images. We expect that a high degree of similarity in these representations transfers to a high value add of augmentations for perception model training and testing. \newline

Our experimental study can also serve as a source of inspiration on how to develop evidence for the suitability of synthetic data within safety argumentation. We believe that a strong argument for synthetic data can be established by addressing three major pillars:

\begin{enumerate}
    \item \textbf{Visual Appearance:} Does the synthetic data closely resemble the corresponding real-world trigger data from a human perspective?
    \item \textbf{Algorithmic Similarity:} Does the synthetic data closely resemble the corresponding real-world trigger data from the perspective of ML algorithms?
    \item \textbf{Performance Boost:} Can the performance of ML algorithms on real-world trigger data be improved with the help of the synthetic data?
\end{enumerate}

Conducting user studies on the `visual appearance' of augmented data is relatively straightforward. Therefore, we limit our work on this pillar to presenting illustrative examples, as shown in Figure \ref{fig:augmented_images} and Section \ref{sec: augmentation_samples} of the Appendix. \newline
To address `algorithmic similarity' we conduct (1) an analysis of FID and CMMD scores in Section \ref{sec: FID_CMMD_Analysis}, and (2) an evaluation of augmented data on a multi-weather classifier in Section \ref{sec: Weather_Classifier}. \newline
Finally, to give evidence for `performance boost', we fine-tune semantic segmentation models with \textsc{A-BDD} to enhance performance on real-world rain data from \textsc{ACDC} in Section \ref{sec: performance_boost}.

\subsection{FID \& CMMD Analysis} \label{sec: FID_CMMD_Analysis}

Before incorporating synthetic adverse weather data into training and testing processes, it is helpful to determine whether existing real-world weather and lighting conditions represent substantial distributional shifts from the perspective of perception algorithms. If adverse weather conditions lead to significant distributional shifts in feature embeddings, future model candidates are likely to handle adverse weather data differently from fair weather data, potentially resulting in weaker performance in these areas of the ODD. \newline
We aim to analyze this question using the \textsc{BDD100K} and \textsc{ACDC} perception datasets.
Usually, engineers have access to at least some real-world data corresponding to challenging weather and lighting conditions, but not enough to actually utilize this data in model training. A similar situation holds true for \textsc{BDD100K} and \textsc{ACDC}. In both datasets, we find adverse weather data, but corresponding dataset sizes are quite limited. 

To assess the presence of challenging distributional shifts, we can leverage the feature-based image quality metrics FID and CMMD. While these scores are typically employed to compare real-world and synthetic data, there's no reason why they should not also aid in detecting distributional shifts between different real-world trigger conditions. Therefore, we calculate the cross product of these metrics across the different weather conditions given in the two datasets. The results are summarized in Table~\ref{tab:bdd_distance} and Table~\ref{tab:acdc_distance}. Unfortunately, we are unable to report FID scores for the foggy data in \textsc{BDD100K} due to insufficient real-world fog images for the metric to converge ($\approx50$ images). This limitation partly motivated the use of the second, more sample-efficient image quality metric, CMMD. \newline

\begin{figure}[H]
    \centering
    \captionsetup{skip=0.5\abovecaptionskip, font=footnotesize} 
    \includegraphics[width=\linewidth]{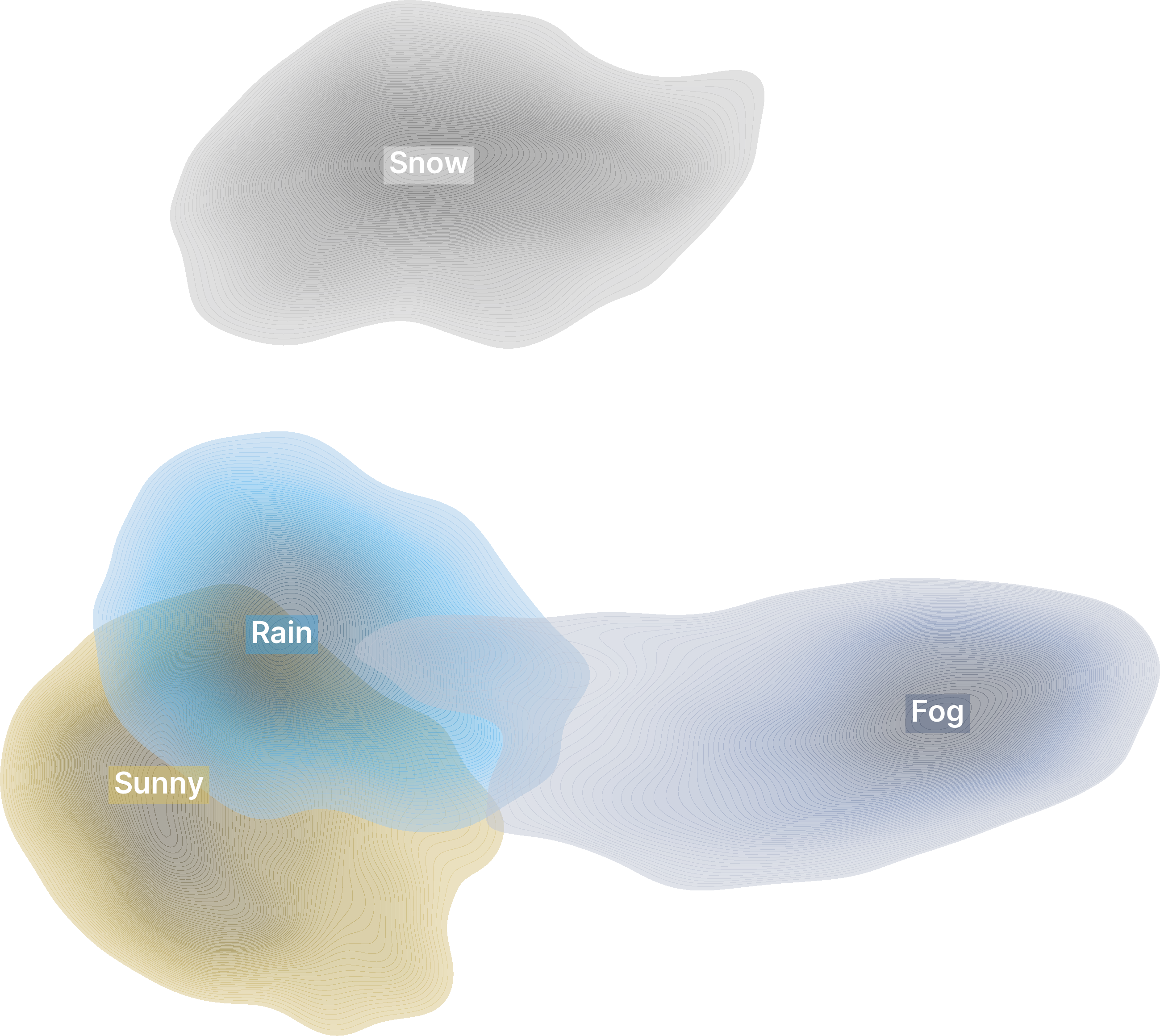}
      \vspace{0.1cm}
    \caption{Kernel Density Estimation (KDE) distributions of CLIP feature embeddings for \textsc{ACDC} trigger data projected with Principal Component Analysis (PCA). The CLIP feature embeddings are the basis for the CMMD calculation (see Section \ref{sec: image_quality_metric}).}
    \label{fig:embedding_distribution}
\end{figure}

There are noticeable distances between the different triggers, which suggests that the corresponding adverse weather conditions induce distinct activation patterns within state-of-the-art ML models. In Figure \ref{fig:embedding_distribution} we illustrate this intuition derived from the FID/CMMD scores by plotting projected versions of the underlying CLIP \cite{OpenAI2021CLIP} feature embeddings for \textsc{ACDC}. \newline
Overall, these activation differences appear to be more pronounced in \textsc{ACDC} as compared to \textsc{BDD100K}, which is consistent with our observations during data preparation and cleaning. 
The \textsc{BDD100K} adverse weather data is often hard to differentiate from clear/fair weather (e.g.\@, only small puddles on the street for rain data, or minor snow piles at the side of the road for snow data). 
Especially, the separation between the `overcast' and `clear' weather attributes of \textsc{BDD100K} appears somewhat arbitrary, and during visual inspection it was challenging to clearly assign images to only one of these two categories. With a FID score of $50.7$, and a CMMD score of $0.15$, these two triggers are extremely close to each other. Thus, the slightly darker, more cloudy sky of overcast images does not seem to have a significant impact on the activation patterns within ML models. \newline
Across both datasets, the trigger conditions `rain', `snow' and `fog' have a rather high distance to fair weather data (with CMMD scores between $0.64$ and $2.26$). Simultaneously, these adverse trigger conditions exhibit high distance scores from one another, which suggests that they create different activation patterns and represent different distributional shifts from the perspective of an ML model. \newline

\begin{figure*}[t]
    \centering
    \captionsetup{skip=0.5\abovecaptionskip, font=footnotesize} 
    \begin{minipage}[t]{0.45\linewidth}
        \centering
        \includegraphics[width=\linewidth]{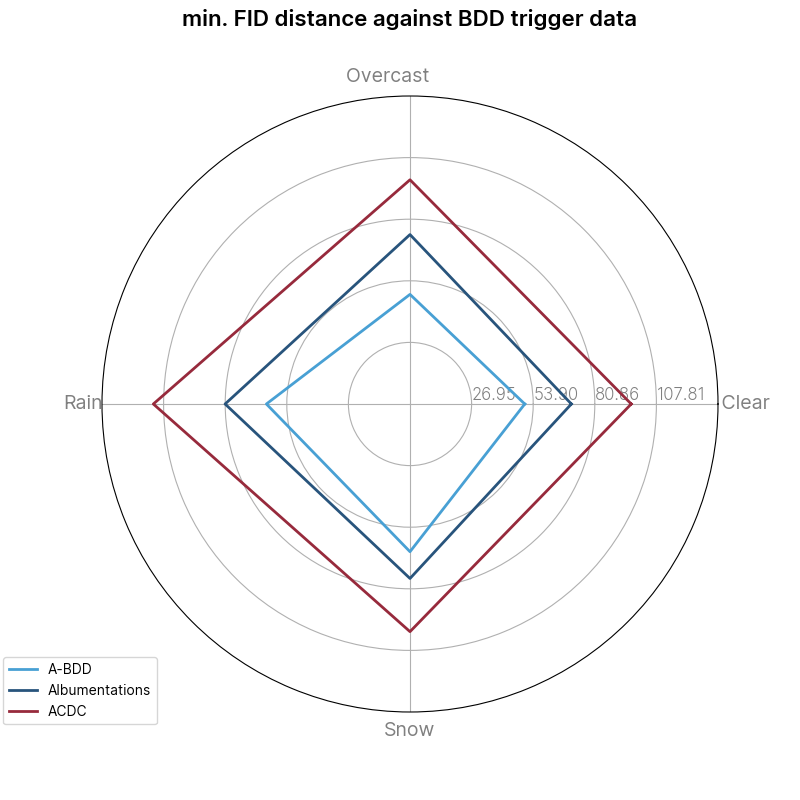}
    \end{minipage}\hfill
    \begin{minipage}[t]{0.45\linewidth}
        \centering
        \includegraphics[width=\linewidth]{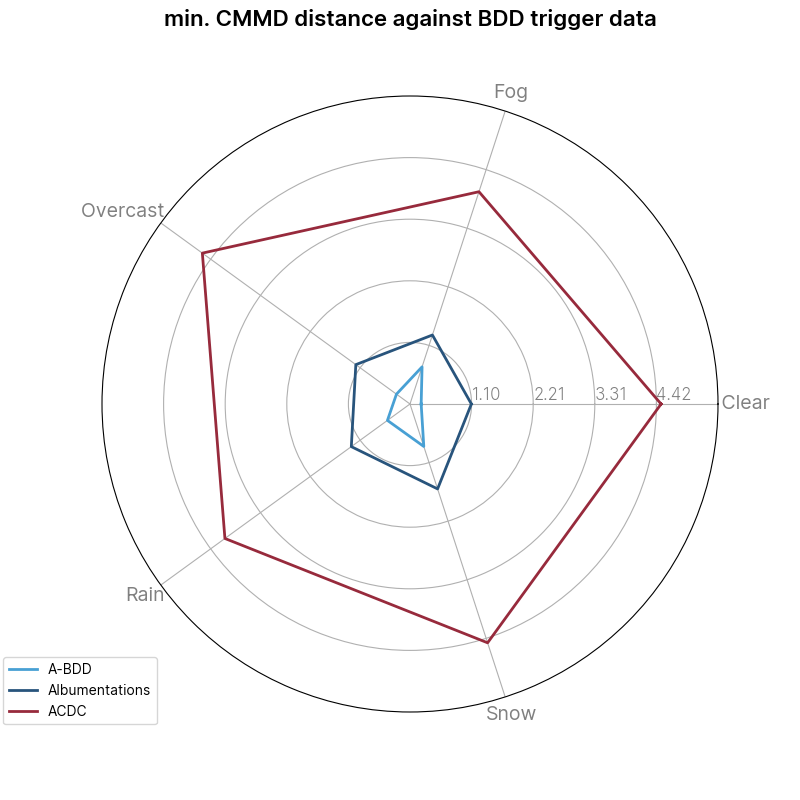}
    \end{minipage}
    \caption{Minimal FID/CMMD distance between augmentation subset of \textsc{A-BDD}, augmentation subset of Albumentations, and \textsc{ACDC} trigger data to \textsc{BDD100K} trigger data. The augmentation sets of \textsc{A-BDD} are significantly closer to the weather conditions of \textsc{BDD100K} compared to the other two datasets. In particular, we observe a notable distributional shift between the real-world trigger data from \textsc{BDD100K} and \textsc{ACDC}.}
    \label{fig:spiderplot}
\end{figure*}
Having identified meaningful distances between real-world weather data, we now calculate FID and CMMD scores comparing the augmented data of \textsc{A-BDD}, as well as augmented data generated with the open-source tool Albumentations, to the unaugmented data of \textsc{BDD100K} and \textsc{ACDC}. In Table \ref{tab:fid_scores} and Table \ref{tab:cmmd_scores} of the Appendix we list all calculated distances. Additionally, we summarize these - rather hard to comprehend - results with spider charts, see Figure~\ref{fig:spiderplot}. \newline
The Albumentations augmentations are based on the same $1{,}820$ unaltered images as \textsc{A-BDD} (see Section \ref{sec: A-BDD}). We generate $15$ different versions of these $1{,}820$ images with Albumentations, based on various parameter configurations and trigger conditions (i.a., sun, rain, and fog). \newline
The results depicted in Figure~\ref{fig:spiderplot} indicate that \textsc{A-BDD} has a close distance to \textsc{BDD100K}. The augmented sets of \textsc{A-BDD} are able to get closer to adverse weather conditions, compared to unaltered clear and overcast data. For example, we see that the augmentation \texttt{wet\_street\_lens\_droplets\_4}, which adds overcast to the sky and reflections to the streets, obtains an FID distance of $62.02$ to real \textsc{BDD100K} rain data, which is significantly lower than $73.6$ and $64.9$ for clear and overcast \textsc{BDD100K} data, respectively. \newline
At the same time, the augmented data of Albumentations is not able to obtain a similar level of proximity. Taking the adverse weather condition `rain' as an example, we do not find any augmentation and parameter configuration of Albumentations that results in an FID distance to real \textsc{BDD100K} rain data below $80.9$, which is comparably high when looking at \textsc{A-BDD}. Twenty-seven of the provided $35$ augmented sets of \textsc{A-BDD} are below this value, i.e., have a lower distance to real-world rain data. \newline

The two metrics, FID and CMMD, do not always align in their relative distance estimation. Thus, one can find augmented datasets where the FID score suggests close, minimal proximity to one real-world trigger, while the CMMD score indicates otherwise. This seems to be particularly often the case for the `puddles', `shadow' and `overcast' augmentations of \textsc{A-BDD}, where the FID score is minimal with respect to overcast data, while CMMD is minimal with respect to clear data of \textsc{BDD100K}. However, as previously mentioned, there is an extremely high visual similarity between these two image attributes of \textsc{BDD100K}. Therefore, it is not surprising that image quality metrics may differ in their assessments of these two attributes. \newline

In Table \ref{tab:fid_scores} and Table \ref{tab:cmmd_scores} we also list the distances of \textsc{A-BDD} to \textsc{ACDC}. We observe that the FID and CMMD scores are significantly higher in this context and are not able to match those of the cross product between real-world data shown in Table \ref{tab:acdc_distance}. The average FID distance to rainy data, calculated across all $35$ augmented sets of \textsc{A-BDD}, is $129.32$ in the case of \textsc{ACDC}, whereas it is only $74.27$ for \textsc{BDD100K}. This indicates that there is a substantial distributional gap between these two datasets. This gap also is visible in the spider charts of Figure~\ref{fig:spiderplot}, where we report minimal distances between the two unaltered real-world datasets. \newline
One question that we are interested in is whether the augmented data from \textsc{BDD100K} can still be helpful in perception tasks related to \textsc{ACDC}. In the upcoming sections, we will analyze this in more detail. \newline

To sum up, looking at image quality metric scores we observe significant differences in feature embeddings between real-world weather data, which increases the need to take protective measures during perception training. Due to its rather low metric values, compared to augmentations from Albumentations, we have obtained first indications of the potential value of \textsc{A-BDD}. The augmented data of \textsc{A-BDD} is able to come close to real-world adverse weather data, based on the given image quality metrics. In the following sections, we provide further evidence that these calculated scores can be leveraged for data selection in perception training and testing. 

\subsection{Adverse Weather Classifier} \label{sec: Weather_Classifier}

A multi-weather classifier can be used to learn features corresponding to different weather conditions, for example rain, fog, snow and sun/sunglare. It is uncertain whether a weather classifier perceives similar features when facing augmented/synthetic adverse weather data. \newline
To investigate this, we first train a weather classifier on real-world data. Afterwards we then use the trained classifier to predict the weather on various augmented datasets. A high classification certainty would give us a further indication that the augmented data effectively mimics real-world adverse weather data. \newline

\begin{table}
\centering
\captionsetup{skip=0.5\abovecaptionskip, font=footnotesize} 
\begin{tabular}{l|c|c|c}
\toprule
Weather Classifier & Precision & Recall & F1 \\
\midrule
Fog & 0.94 & 0.68 & 0.79  \\
Rain & 0.71 & 0.87 & 0.78 \\
Snow & 0.64 & 0.88 & 0.74  \\
Sun/Sunglare & 0.92 & 0.68 & 0.78 \\
\bottomrule
\end{tabular}
\vspace{0.2cm}
\caption{Evaluation results of fine-tuned weather classifier on \textsc{ACDC} validation data. The \textsc{ACDC} validation data consists of $801$ images for fog, $800$ for rain, $660$ for snow, and $801$ for sun/sunglare. The slight imbalance with respect to snowy data explains the comparably low precision score for this trigger condition.}
\label{tab:weather_classifier_performance}
\end{table}
We start with an ImageNet pretrained VGG16 model. We only train the parameters of the last fully connected layer, ensuring that the output classes match the number of desired weather conditions (i.e., rain, fog, snow and sun). The model is fine-tuned using open-source data comprising $4{,}310$ images sourced from the internet (not ADAS/AD focused) and evenly distributed across the four weather conditions. \newline
For validation, we focus on data from \textsc{ACDC} due to the larger visual differences and, the previously discussed, larger image quality metric distances between the different weather conditions (see Section \ref{sec: FID_CMMD_Analysis}). These characteristics are expected to facilitate strong performance of the multi-weather classifier on \textsc{ACDC}.
We take images from the \textsc{ACDC} training and validation dataset, but remove images that can not easily be mapped to only one of the four weather conditions.
\begin{table*}[t]
\centering
\captionsetup{skip=0.5\abovecaptionskip, font=footnotesize}
\begin{tabular}{lcc}
\toprule
\textbf{Correlation:}  & C-CMMD \& Predictions & C-FID \& Predictions  \\
\midrule
Fog & $0.96 / 7.4e$-$20$ & $0.46 / 5.7e$-$3$ \\
Rain & $0.90 / 2.6e$-$13$ & $0.02 / 9.0e$-$1$ \\
Sunglare & $0.80 / 5.9e$-$9$ & $0.64 / 3.8e$-$5$  \\
\bottomrule
\end{tabular}
\vspace{0.1cm}
\caption{Correlation results (Pearson correlation / p-value) between image quality metric scores of the subsets of \textsc{A-BDD} and the corresponding class predictions of the weather classifier with respect to a given trigger condition. All used FID/CMMD, and consequently also C-FID/C-CMMD, scores refer to \textsc{ACDC}. We did not include snow in this correlation analysis, as we did not incorporate any snow augmentations into \textsc{A-BDD}.}
\label{tab:weather_classifier_correlation}
\end{table*} 
\newline We fine-tune the VGG16 model by training $5$ epochs with a batch size of $16$ and an Adam optimizer. This results in a training accuracy of $81 \%$ on the scraped training data, and a validation accuracy of $78.5 \%$ on the cleaned \textsc{ACDC} dataset. Further details on the performance of the weather classifier on the validation set can be found in Table \ref{tab:weather_classifier_performance}. 
For the trigger conditions relevant to our dataset - `fog', `rain', and `sunglare' - the model achieves an accuracy of at least $76 \%$ on the corresponding \textsc{ACDC} trigger data. \newline

After having trained a weather classifier, we then test the model on augmented data. We run inference on data from all $35$ subsets of \textsc{A-BDD}, as well as on data from all $15$ generated Albumentations sets. From every subset we sample $800$ out of the $1{,}820$ augmented images for this evaluation (see Table \ref{tab:contrastive_scores} of Appendix). \newline
Since \textsc{A-BDD} often combines weather characteristics of fog and rain (e.g., reflections, puddles, and overcast), there is not always a single clear classification target for every augmented set. In other words, it is often hard to tell whether an image should be classified as fog or rain by the multi-weather classifier. \newline
Nevertheless, the weather classifier is quite confident about $12$ of the $35$ augmented sets of \textsc{A-BDD}. This suggests that it perceives similar features to those found in real-world adverse weather data. On these $12$ sets, it assigns over $70 \%$ of the augmented images to one particular weather class. The results are particularly convincing for fog. Here, the higher intensity levels (i.e., $\geq 3$) of the augmentations \texttt{rain\_composition} and \texttt{dense\_fog} are classified as fog with over $91 \%$ accuracy. \newline
In general, for the three weather and lighting conditions also covered by \textsc{A-BDD} - rain, fog, and sunglare - we find an augmented subset of \textsc{A-BDD} where over $52 \%$ of the images are assigned to the respective condition. \newline
Comparing the inference results of \textsc{A-BDD} with those of Albumentations, we do not see major differences for the weather conditions rain and sunglare. Both augmented datasets contain subsets that achieve similar classification results. However, for fog the results deviate significantly. Here, none of the Albumentations sets gets more than $43 \%$ of the images classified as fog, whereas the best set of \textsc{A-BDD} has a classification rate of $99 \%$. \newline


Lastly, we also want to link the inference results of the weather classifier to the calculated FID/CMMD scores. Here, it becomes apparent that a lower distance to a trigger condition does not necessarily mean that the weather classifier will assign the augmented set to that particular weather phenomenon. For example, for the lowest intensity level of \texttt{rain\_composition}, the weather classifier assigns $46 \%$ of the images to fog and only roughly $10 \%$ to rain, even though the CMMD score suggests that the data is closer to rain than to fog (CMMD: $4.03$ for rain vs. $4.09$ for fog). This brings us to the hypothesis that for fooling a weather classifier, it is not only important that the augmented data is close to the desired trigger, but, at the same time, maintains a certain distance from the other weather classes. \newline
This rather qualitative intuition motivates the definition of Contrastive-FID (C-FID) and Contrastive-CMMD (C-CMMD). Let $t_1, ..., t_n$ denote the $n$ different weather and lighting triggers of a real-world dataset (i.e.\@, $n=4$ for \textsc{ACDC}). We define

\begin{align*}
\begin{split}
&C\text{-}FID(\hat{X}, t_i) := \frac{\sum_{t_j \neq t_i } FID(\hat{X}, X_{t_j}) - FID(\hat{X}, X_{t_i})}{ FID(\hat{X}, X_{t_i})} \\
&\phantom{C\text{-}FID(\hat{X}, t_i) :}= \sum_{t_j \neq t_i } \frac{FID(\hat{X}, X_{t_j})}{FID(\hat{X}, X_{t_i})} - (n-1) \\
&C\text{-}CMMD(\hat{X}, t_i) := \sum_{t_j \neq t_i }\frac{CMMD(\hat{X}, X_{t_j})}{CMMD(\hat{X}, X_{t_i})} - (n-1), \
\end{split}
\end{align*}

where $\hat{X}$ denotes a real-world or synthetic dataset, which should be compared to the trigger condition of interest $t_i$, and its corresponding trigger data $X_{t_i}$. C-FID and C-CMMD are designed such that their values increase when $\hat{X}$ is close to $X_{t_i}$, while maintaining a greater distance from other triggers, $X_{t_j}$, where $j \neq i$. \newline

We calculate C-FID and C-CMMD for all subsets of \textsc{A-BDD} based on their respective FID and CMMD values for \textsc{ACDC} (see Appendix Table \ref{tab:contrastive_scores}). Subsequently, we calculate the Pearson correlation coefficients between the C-FID/C-CMMD scores and the number of images that were classified as a certain weather phenomenon. The correlation results are shown in Table \ref{tab:weather_classifier_correlation}. We did not include snow in this correlation analysis, as we did not incorporate any snow augmentations into \textsc{A-BDD}. \newline
As expected, there is a positive correlation for all weather triggers. In particular, C-CMMD leads to correlation coefficients above $0.8$ with p-values below $0.05$, highlighting a strong correlation between an increase in this contrastive score and the classification as the respective weather condition by the classifier. In other words, the C-CMMD for a weather condition can be a indicator of whether the weather classifier will assign the data to the respective weather class. \newline

Summing up, the synthetic data of \textsc{A-BDD} successfully fools the weather classifier across multiple weather conditions, leading to similar classification results as real-world weather and lighting data from \textsc{ACDC}. In addition to the visual appearance and promising image quality scores (see Section \ref{sec: FID_CMMD_Analysis}), this further supports the usefulness of the provided dataset. By examining adapted versions of FID and CMMD, we observed that in order to deceive a weather classifier into inferring a specific weather class it is crucial not only to minimize the distance to the target trigger but also to maintain a distinct distance from other weather phenomena.

\subsection{Semantic Segmentation Fine-Tuning} \label{sec: performance_boost}
Ultimately, for perception development teams, the relevance of an augmented dataset hinges on its ability to enhance model performance when included into \mbox{(re-)training} sets. Therefore, we use our augmented data to fine-tune semantic segmentation models with the aim of increasing their performance. \newline

\begin{figure*}[t]
\captionsetup{skip=0.5\abovecaptionskip, font=footnotesize}
    \centering
    \begin{minipage}[t]{0.5\linewidth}
        \centering
        \includegraphics[width=\linewidth]      {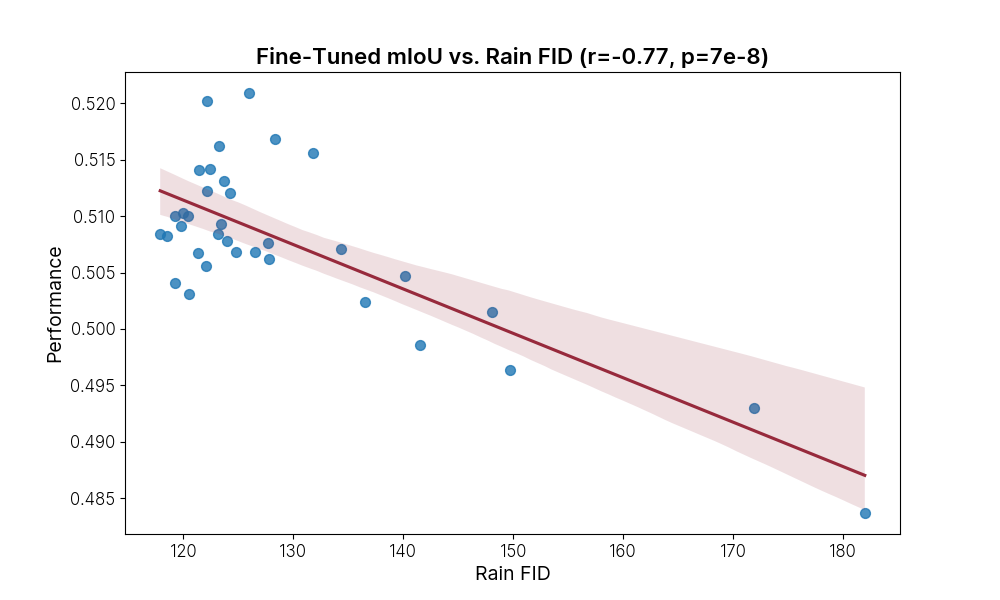}     
    \end{minipage}\hfill
    \begin{minipage}[t]{0.5\linewidth}
        \centering
        \includegraphics[width=\linewidth]{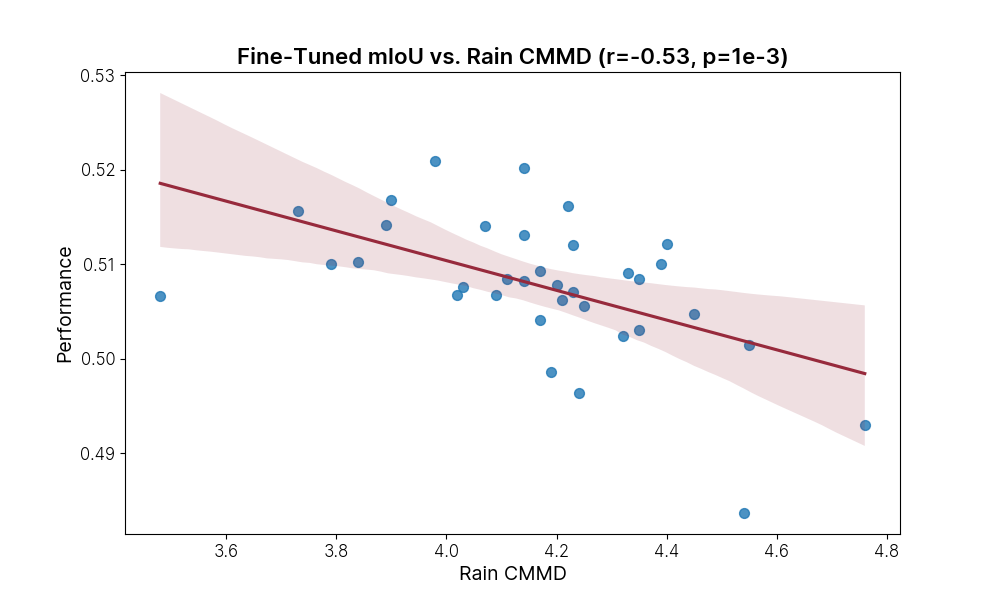}
       
    \end{minipage}
    \caption{The plots show FID/CMMD distances to \textsc{ACDC} rain (x-axis) and corresponding mIoU results on \textsc{ACDC} rain after model fine-tuning (y-axis) of all $35$ augmentation sets of \textsc{A-BDD}. A clear negative correlation is observed between FID/CMMD distances and performance gains, highlighting the importance of feature embedding similarity for the success of model training with augmentations.}
    \label{fig:combined_correlation_plots}
\end{figure*}

We start with a DeepLabv3 model with an R-50-D8 backbone that has been trained on \textsc{BDD100K}. Without further fine-tuning, this model obtains an mIoU of $61.43$ on the validation set of \textsc{BDD100K}. However, when facing the yet unseen \textsc{ACDC} rain training data, the performance of the model drops to $50.67$ mIoU, which underlines the previously identified distributional shift between these two datasets. \newline 
The open question is whether the augmented data from \textsc{BDD100K} can help reduce this performance gap. \newline
The outlined experimental setup closely mirrors the situation faced by perception development teams in the automotive sector. These teams typically have access to collected fleet data for model training, and they must ensure that the developed perception functions maintain a sufficient performance level across demanding adverse weather conditions of the end user's driving environment. However, the existing fleet data may not adequately capture the end user's driving environment, and this limitation could also extend to the augmented data derived from it. \newline

\begin{table}
\centering
\captionsetup{skip=0.5\abovecaptionskip, font=footnotesize} 
\begin{tabular}{l|c|c|c}
 & FID & CMMD & mIoU \\
\toprule
Baseline & 124.1 & 4.5 & 50.7\% \\
\midrule
+ rain\_streaks\_2 & 126.0 & 4.0 & 52.1\% \\
+ rain\_streaks\_4 & 122.2 & 4.1 & 52.0\%  \\
+ albu\_fog\_3 & 234.7 & 4.2 & 51.3\%  \\
+ albu\_sun\_4 & 119.9 & 4.4 & 51.1\% \\
\bottomrule
\end{tabular}
\vspace{0.2cm}
\caption{Excerpt of Table \ref{tab:fid_scores} and Table \ref{tab:cmmd_scores} showcasing mIoU results of the semantic segmentation models on \textsc{ACDC} rain after fine-tuning with the respective augmentations. In the first row we list the baseline \textsc{BDD100K} pretrained model, which serves as the starting point for the fine-tuned models. Each row also contains FID/CMMD distances of the underlying dataset to \textsc{ACDC} rain data. For the baseline model, these scores refer to the distance between the \textsc{BDD100K} training set and the \textsc{ACDC} rain data.}
\label{tab:excerpt_fine_tuning}
\end{table}
We fine-tune the DeepLabv3 base model by training for an additional $30$ epochs separately on each augmentation and intensity level from \textsc{A-BDD} and Albumentations. We use only the $1{,}820$ augmented images from every augmentation subset without adding any additional \textsc{BDD100K} or \textsc{ACDC} data to the fine-tuning dataset. For our analysis, we then select the best performing model of the $30$ epochs with respect to the \textsc{ACDC} rain training data. To ensure comparability between augmented datasets, we keep training configurations and hyperparameters fixed across all training runs (i.a., SGD optimizer with learning rate $0.005$ and momentum $0.9$). \newline
We end up with $50$ fine-tuned DeepLabv3 models, $35$ based on \textsc{A-BDD} and $15$ based on Albumentations data. \newline

Most of the augmented datasets seem to have a positive effect on the \textsc{ACDC} rain data performance, i.e., the model fine-tuning slightly reduces the existing performance gap. All fine-tuning results can be found in Table \ref{tab:fid_scores} and Table \ref{tab:cmmd_scores} in the Appendix. A simplified direct comparison of the two best-performing \textsc{A-BDD} and the two best-performing Albumentations models is provided in Table \ref{tab:excerpt_fine_tuning}. \newline
The best performing fine-tuned model displays a $2.8 \%$ higher mIoU than the DeepLabv3 base model (fine-tuned with \texttt{rain\_streaks\_2}). The averaging nature of the mIoU metric obscures more significant per-class improvements. For example, the $2.8 \%$ improvement entails improvements of over $10\%$ for $4$ of the $19$ object classes (classes: wall, traffic light, bus and motorcycle). \newline
Overall, $7$ out of the $35$ \textsc{A-BDD} fine-tuned models increase the performance by more than $1.4\%$ compared to the base line model. \newline
On the other hand, none of the Albumentations fine-tuned models improved upon the baseline by more than $1.28 \%$ (best performing Albumentations model: \texttt{albu\_fog\_3}). Hence, the augmentations of \textsc{A-BDD} are more successful in decreasing the performance gap, which might be a consequence of the comparably lower distance to the \textsc{ACDC} adverse weather data (see Section \ref{sec: FID_CMMD_Analysis}). \newline

Similarly to the experiment of Section \ref{sec: Weather_Classifier}, we want to give evidence for the hypothesized usefulness of feature-based image quality metrics for data selection. We calculate the Pearson correlation coefficient between the mIoU of the fine-tuned models and the corresponding FID/CMMD distances to the \textsc{ACDC} rain data (see Figure \ref{fig:combined_correlation_plots}). In this context, we do not make use of the contrastive versions of FID/CMMD, as these scores intuitively make more sense for a weather classification task, where the target model needs to distinguish between different weather conditions. \newline 
We obtain correlation scores of $-0.77$ and $-0.53$ for FID and CMMD (with p-values below $0.05$). Hence, there is a clear negative correlation between the distance to \textsc{ACDC} rain and the fine-tuning results of the respective augmentation. In other words, the fine-tuned model performance on real-world rainy data tends to improve when the augmented data closely matches the relevant rain feature distribution. This suggests that image quality metrics, particularly FID in this experimental setup, can serve as good predictors for the value of synthetic data in perception training.  \newline

This insight can even be leveraged beyond the evaluation of synthetic data. The observed correlations point to the possibility of a more structured data selection process, one which builds around the usage of feature-based image quality metrics. In ADAS/AD development, one often faces a vast amount of collected fleet data alongside relatively arbitrary data annotation and selection decisions. It is hard to determine which collected scenarios will contribute efficiently to improving model performance on critical areas of the ODD. The given image quality metrics could function as early indicators without requiring any expensive annotations. These metrics can help select subsets of collected fleet data, which can then be prioritized for annotation and subsequently used in ML training processes.

\section{Conclusion}
In this paper, we introduce \textsc{A-BDD}, the largest publicly available augmented dataset designed for semantic segmentation and object detection training and testing across a variety of adverse weather and lighting conditions. The dataset consists of $35$ versions of the same $1{,}820$ images from \textsc{BDD100K} related to different adverse trigger conditions and intensity levels (i.a., rain, fog, and sunglare). We showcase the potential of \textsc{A-BDD} by fooling a weather classifier, as well as by improving the performance of state-of-the-art semantic segmentation models on \textsc{ACDC} adverse weather data. \newline
We propose the usage of feature-based image quality metrics, like FID and CMMD, for the identification of promising synthetic data for a given image recognition use case. In particular, we observe strong correlation between image quality metric scores and success in model fine-tuning with augmented data. This correlation opens the door to more sophisticated data selection processes, and in the end to more efficient training processes resulting in model candidates with satisfactory performance results across the ODD. \newline
In general, we hope that more researchers pick up on the idea of leveraging image quality metrics, which utilize feature embeddings of neural networks, outside of GAN development. One drawback of FID and CMMD is their reliance on perception models that are unrelated to AD/ADAS tasks, specifically Inceptionv3 and CLIP. Future work should investigate how incorporating more automotive-relevant perception models influences the effectiveness of image quality metrics for data selection.

\newpage
\section*{Acknowledgments}

We want to thank the Research \& Engineering team at neurocat for making this publication possible. \newline
Additionally, we extend our gratitude to Madhusudhan Vasavakkala and Jesse Paul Lehrke for their efforts in driving business growth, which facilitated obtaining crucial industry feedback and preparing this publication for release.\newline
Most importantly, we want to thank the entire neurocat team for the incredible adventure we have experienced together over the past seven years. We wish you all the best for the exciting next chapters of your life. \newline
Furthermore, we are sincerely grateful to our investor, dSPACE, for being a reliable and constructive partner throughout these rather turbulent years.


\clearpage
\bibliographystyle{ieee}
\bibliography{egbib}

\appendix
\clearpage

\onecolumn              

\begin{landscape}

\section*{Appendix}

\addcontentsline{toc}{section}{Appendix}
In this Appendix we provide additional visual impressions of \textsc{A-BDD} data (see Section \ref{sec: augmentation_samples}) and present all calculated image quality metric results (see Section \ref{sec: big_image_quality_tables}).

\section{Augmentations Samples} \label{sec: augmentation_samples}
\subsection{Rain}
\begin{longtable}{cccc}
    \rotatebox[origin=c]{90}{\textbf{ }} & \textbf{0a56c2e8-e46ca9b7} & \textbf{00067cfb-e535423e} & \textbf{0492b183-725063b0} \\
    \rotatebox[origin=c]{90}{Original Image} &
    \raisebox{-.5\height}{\includegraphics[width=0.3\linewidth]{augmentations/0a56c2e8-e46ca9b7/augmentation_src.jpg}} &
    \raisebox{-.5\height}{\includegraphics[width=0.3\linewidth]{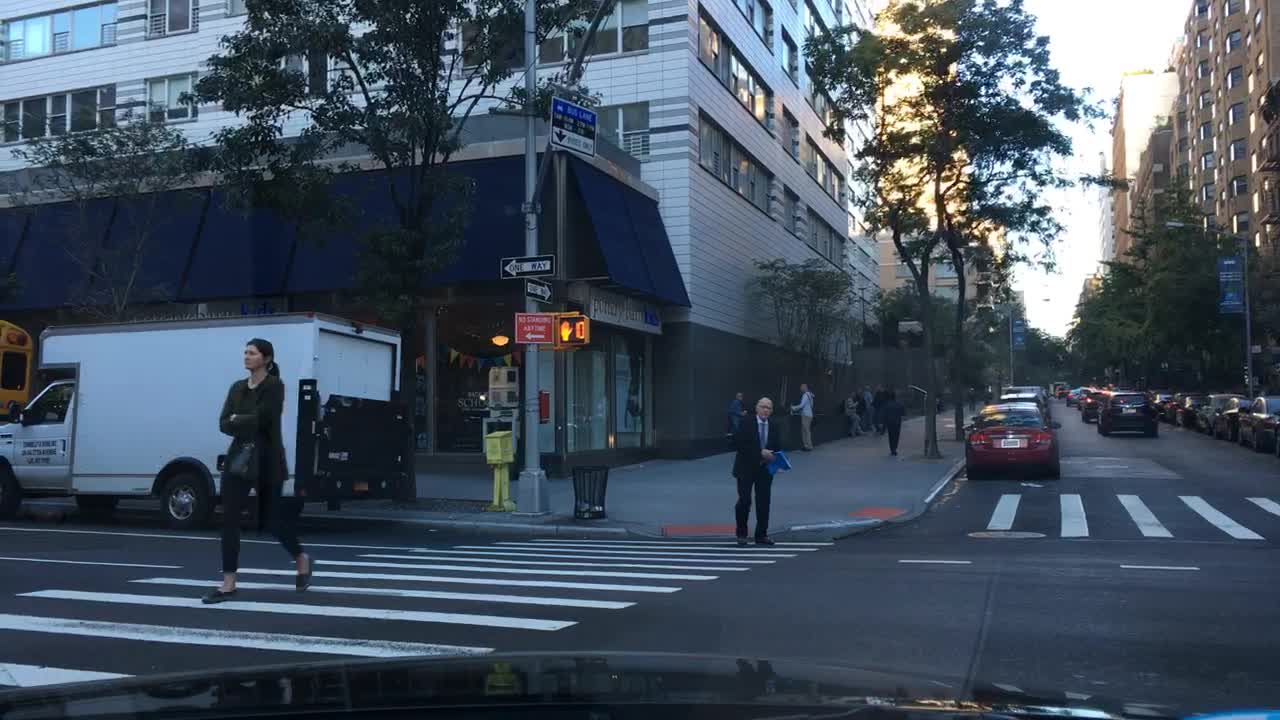}} &
    \raisebox{-.5\height}{\includegraphics[width=0.3\linewidth]{augmentations/0492b183-725063b0/augmentation_src.jpg}} \\
    
    \rotatebox[origin=c]{90}{A-BDD: Overcast 5} &
    \raisebox{-.5\height}{\includegraphics[width=0.3\linewidth]{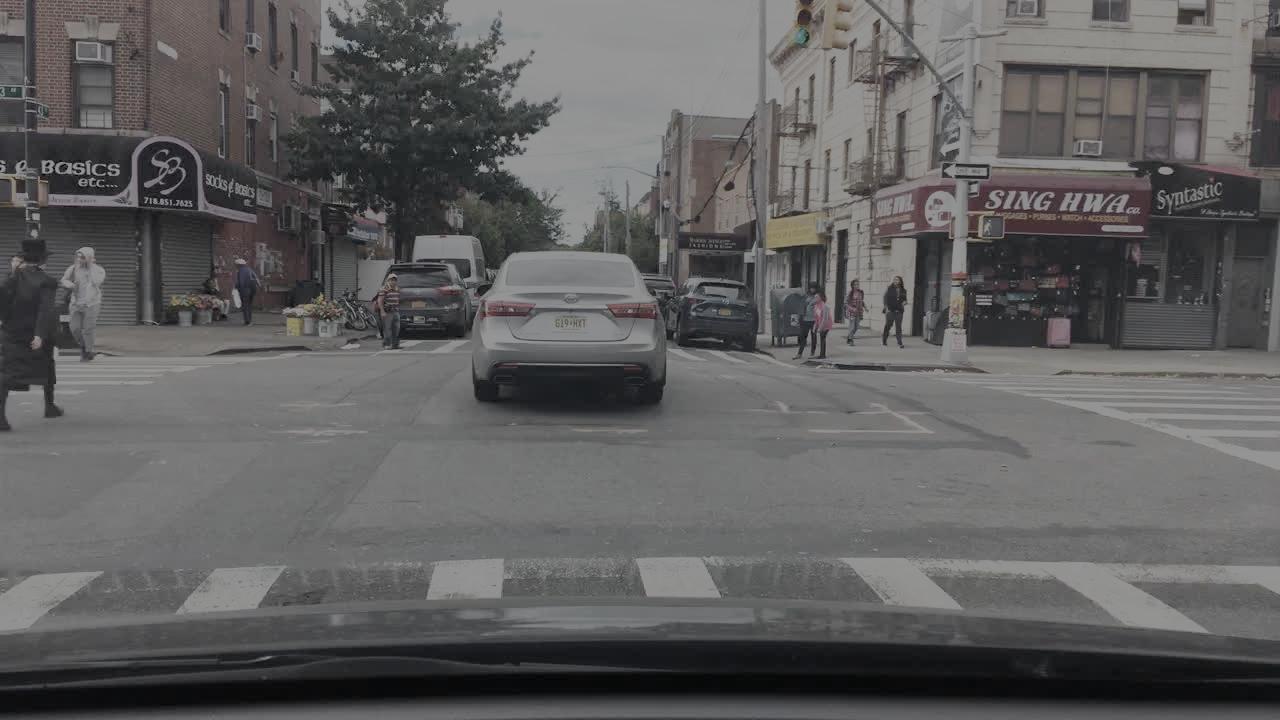}} &
    \raisebox{-.5\height}{\includegraphics[width=0.3\linewidth]{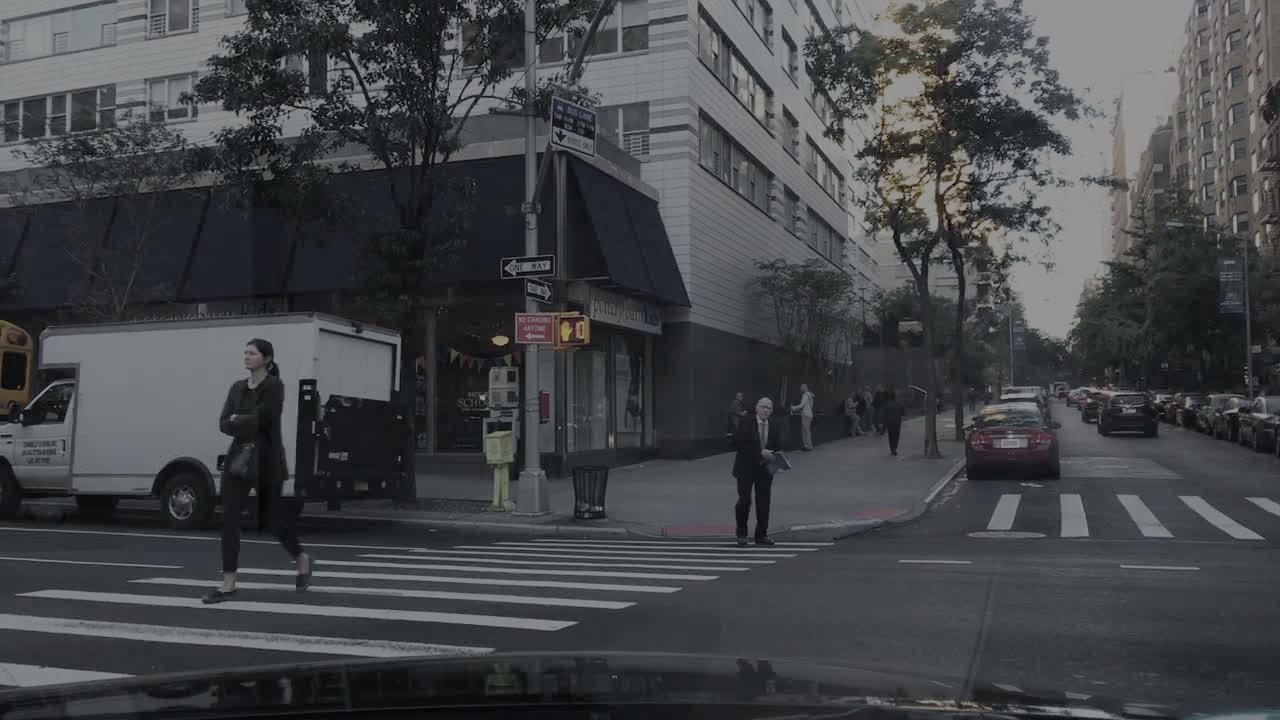}} &
    \raisebox{-.5\height}{\includegraphics[width=0.3\linewidth]{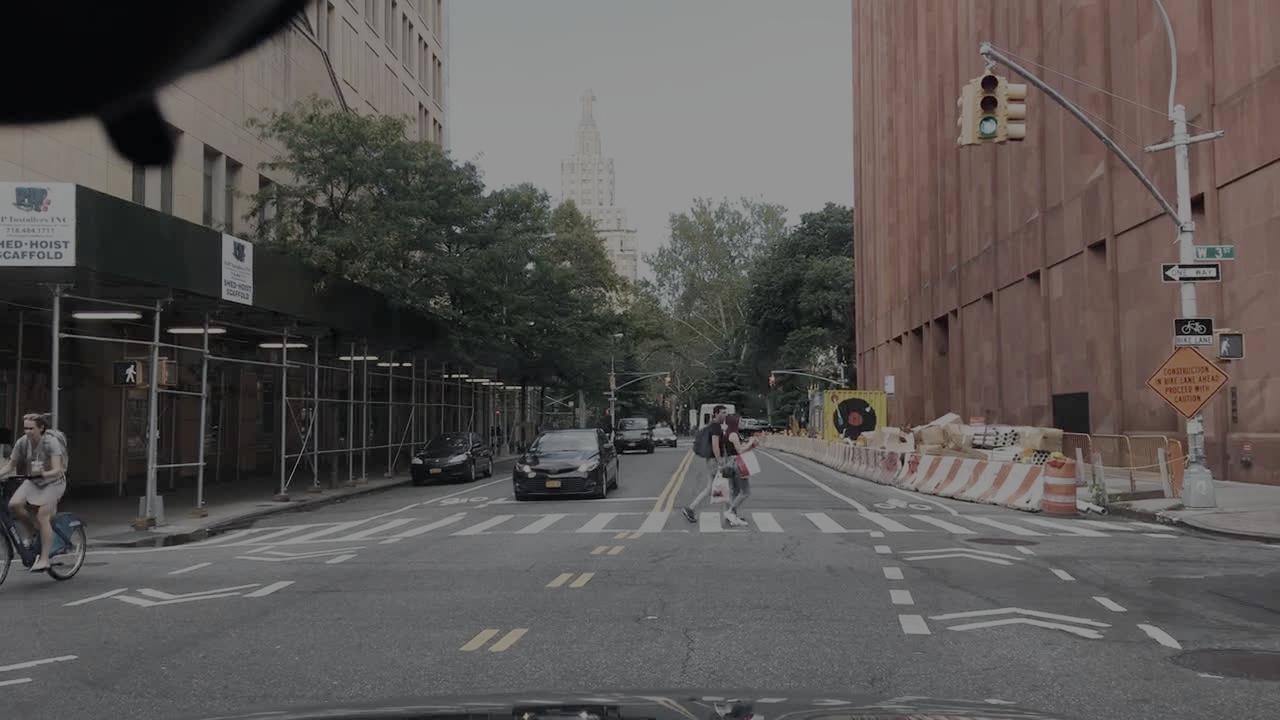}} \\
    
    \rotatebox[origin=c]{90}{A-BDD: Rain Streaks 5} &
    \raisebox{-.5\height}{\includegraphics[width=0.3\linewidth]{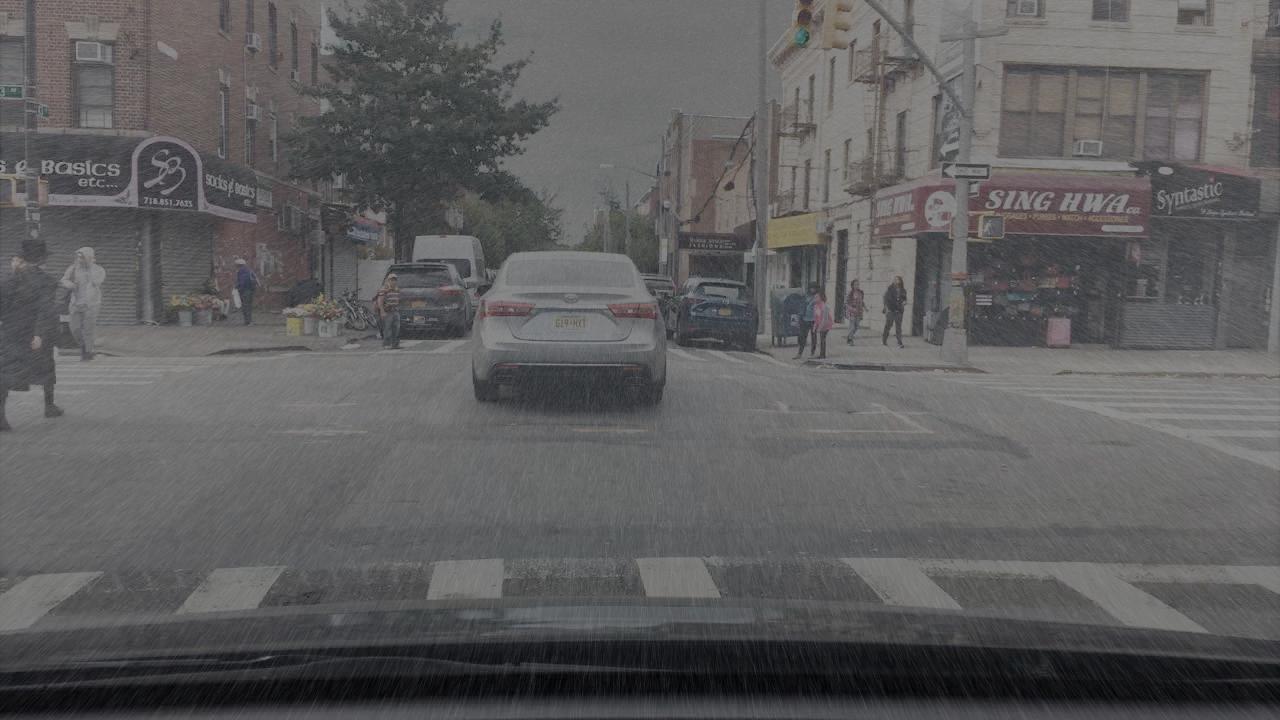}} &
    \raisebox{-.5\height}{\includegraphics[width=0.3\linewidth]{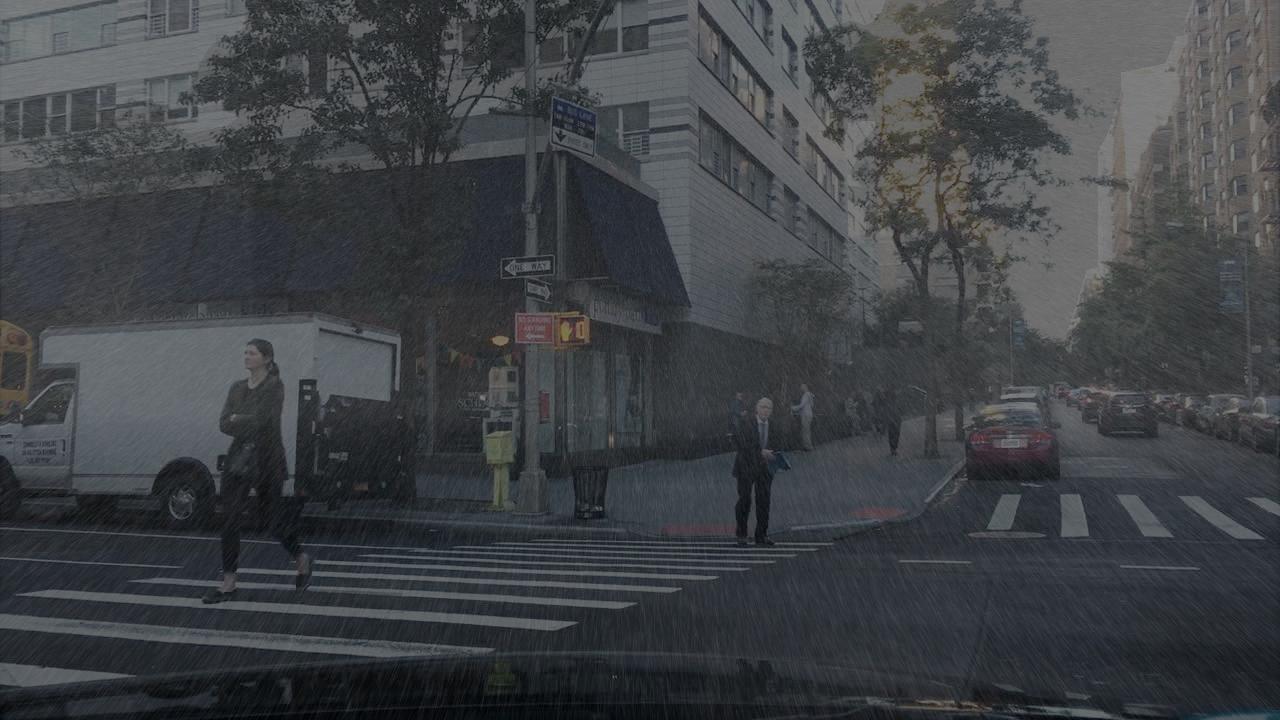}} &
    \raisebox{-.5\height}{\includegraphics[width=0.3\linewidth]{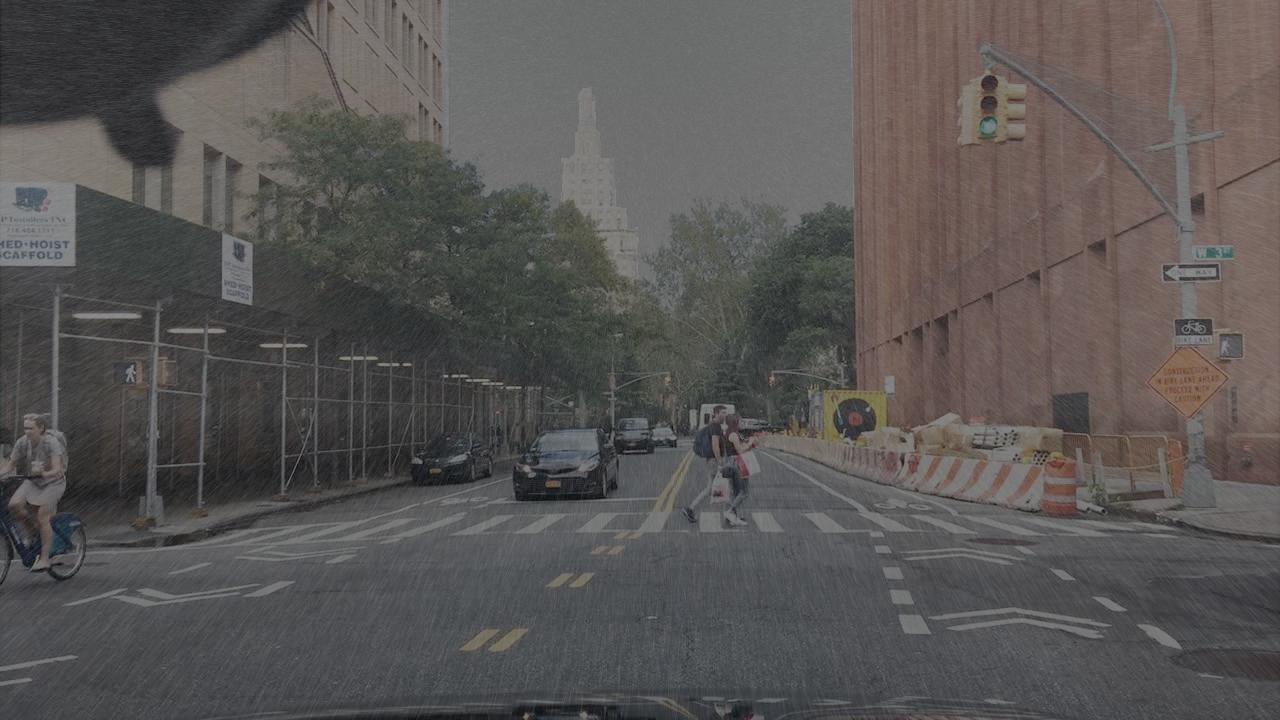}} \\
    
    \rotatebox[origin=c]{90}{A-BDD: Rain Composition 5} &
    \raisebox{-.5\height}{\includegraphics[width=0.3\linewidth]{augmentations/0a56c2e8-e46ca9b7/depth_reflection_8.jpg}} &
    \raisebox{-.5\height}{\includegraphics[width=0.3\linewidth]{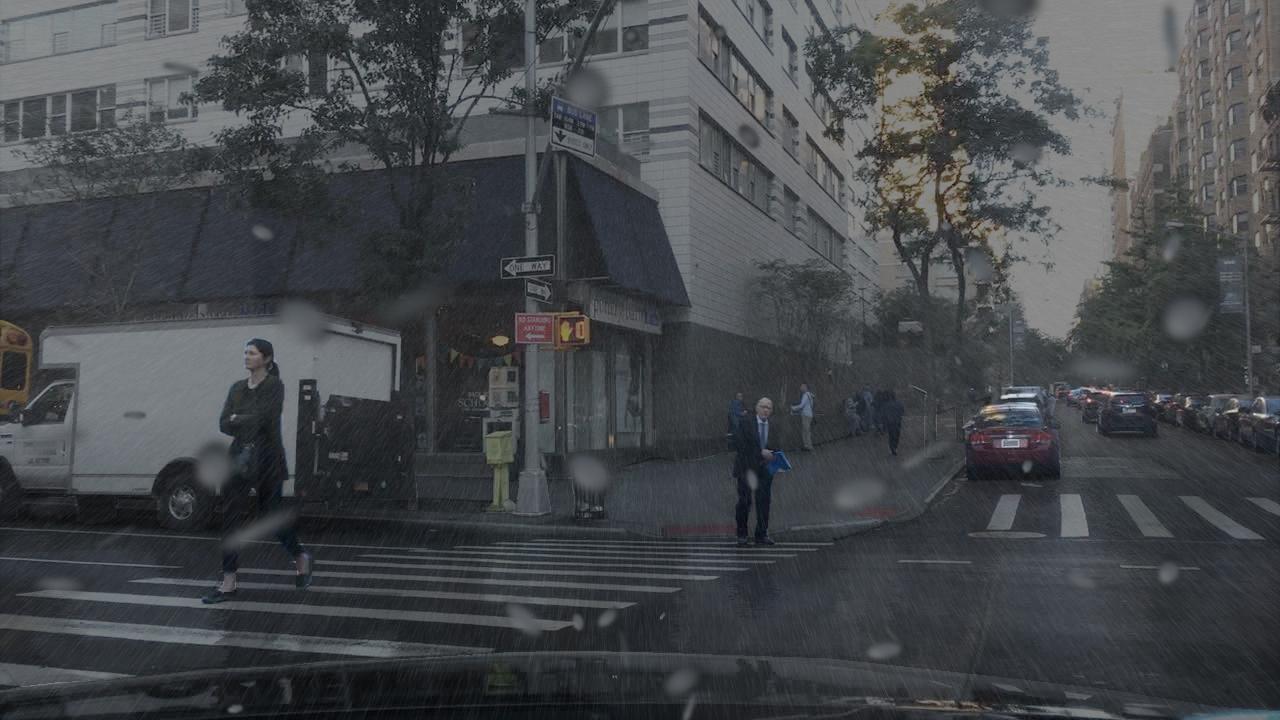}} &
    \raisebox{-.5\height}{\includegraphics[width=0.3\linewidth]{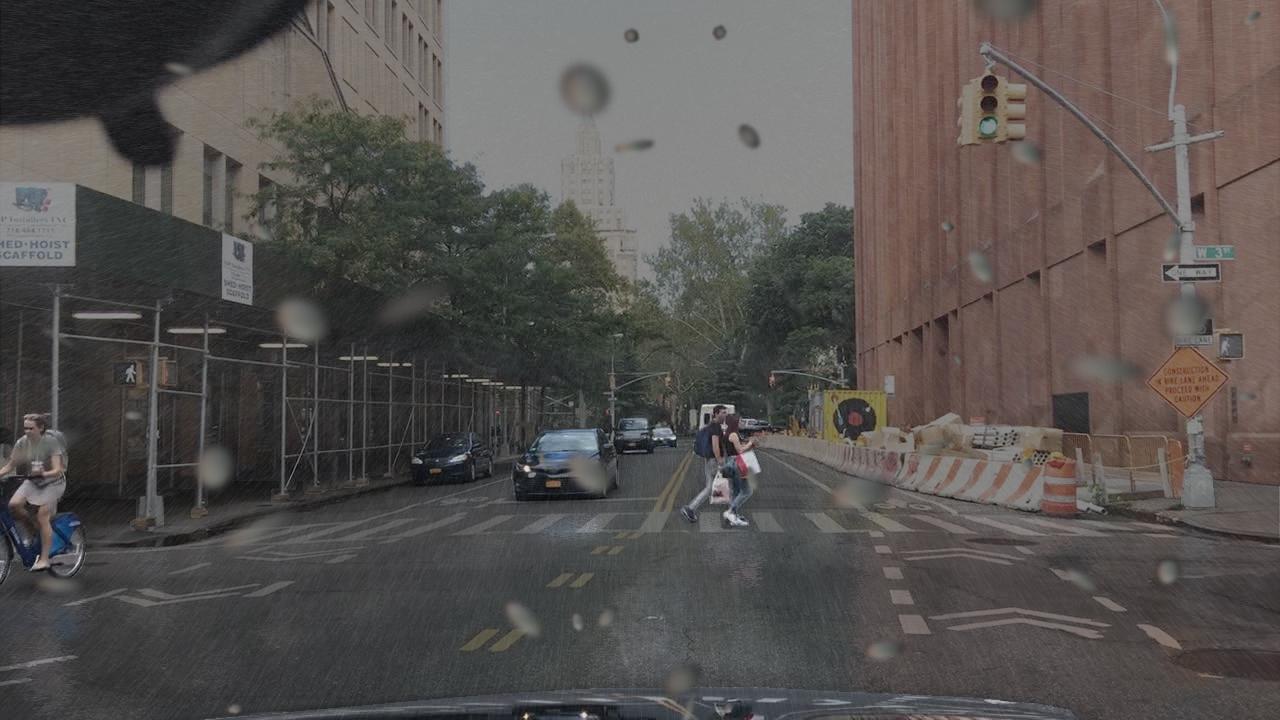}} \\
    
    \rotatebox[origin=c]{90}{A-BDD: Puddles 5} &
    \raisebox{-.5\height}{\includegraphics[width=0.3\linewidth]{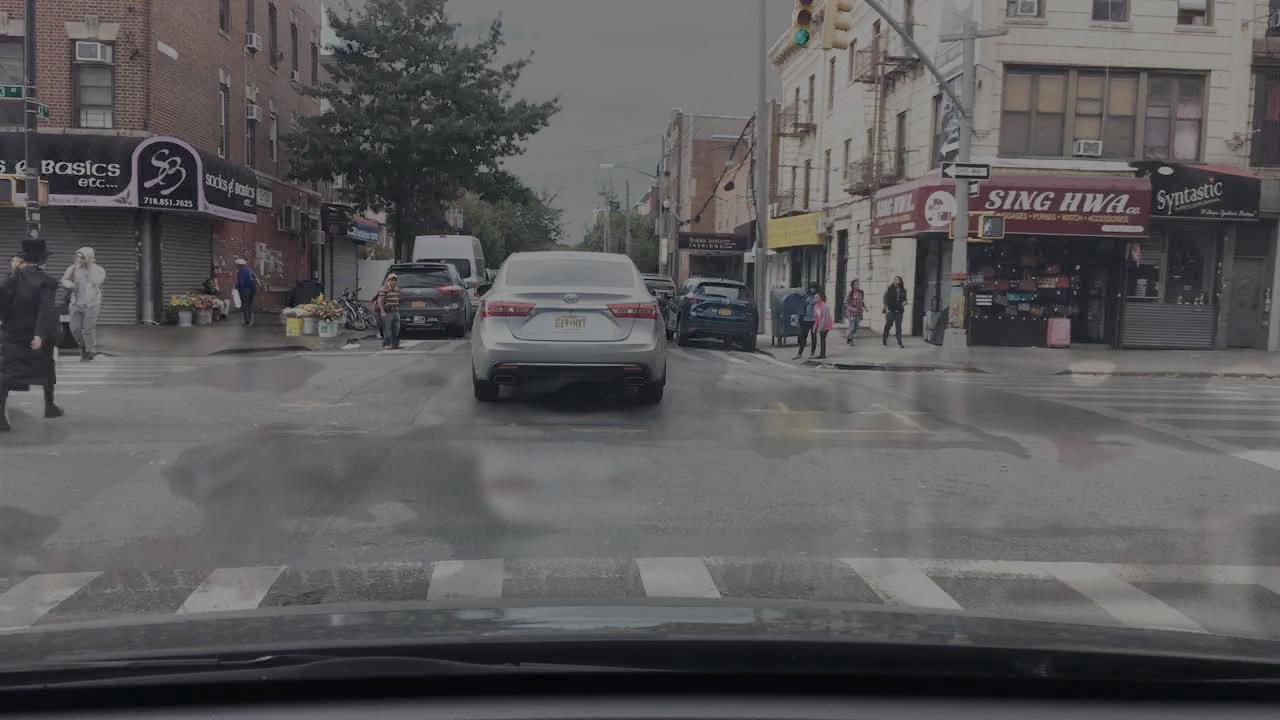}} &
    \raisebox{-.5\height}{\includegraphics[width=0.3\linewidth]{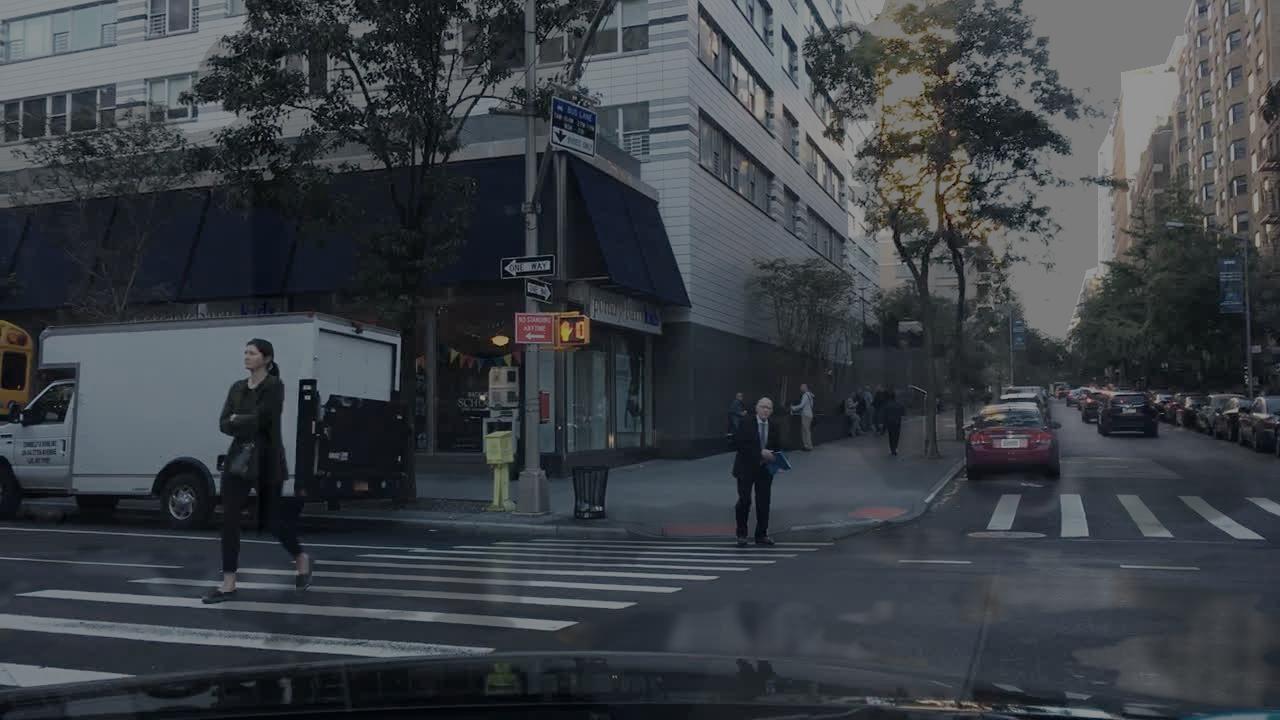}} &
    \raisebox{-.5\height}{\includegraphics[width=0.3\linewidth]{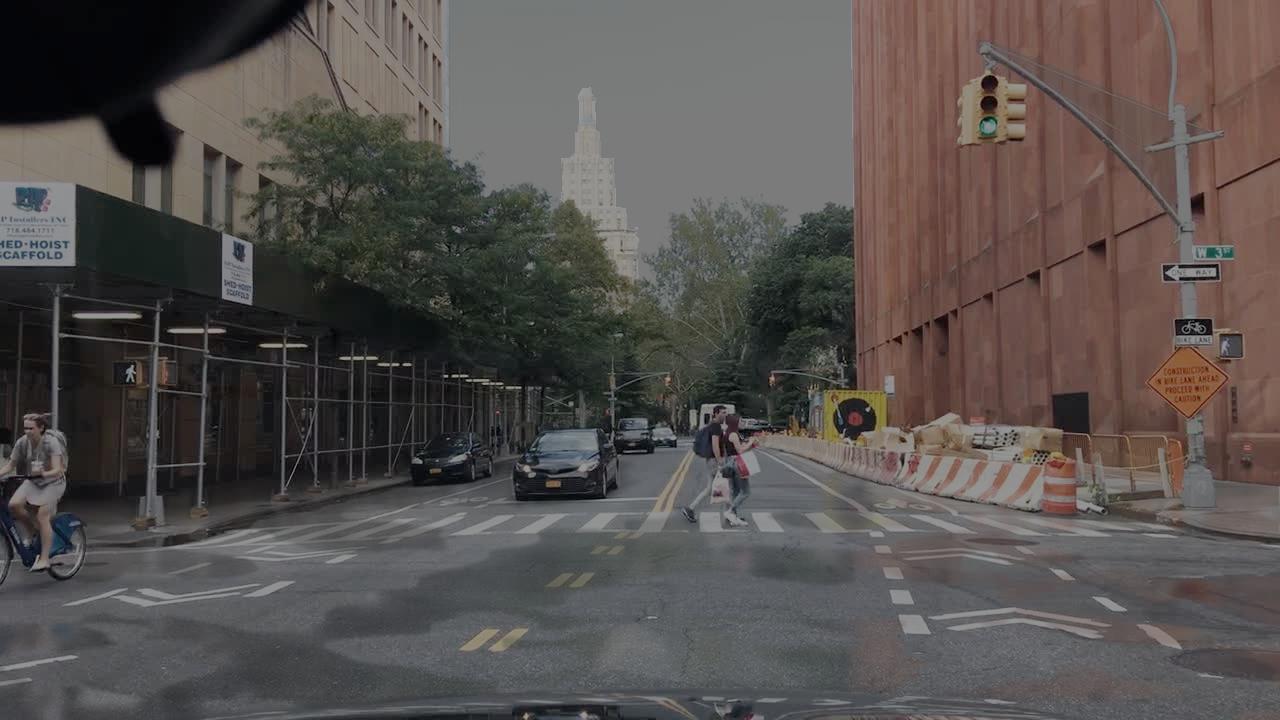}} \\
    
    \rotatebox[origin=c]{90}{Albumentations: Rain 2} &
    \raisebox{-.5\height}{\includegraphics[width=0.3\linewidth]{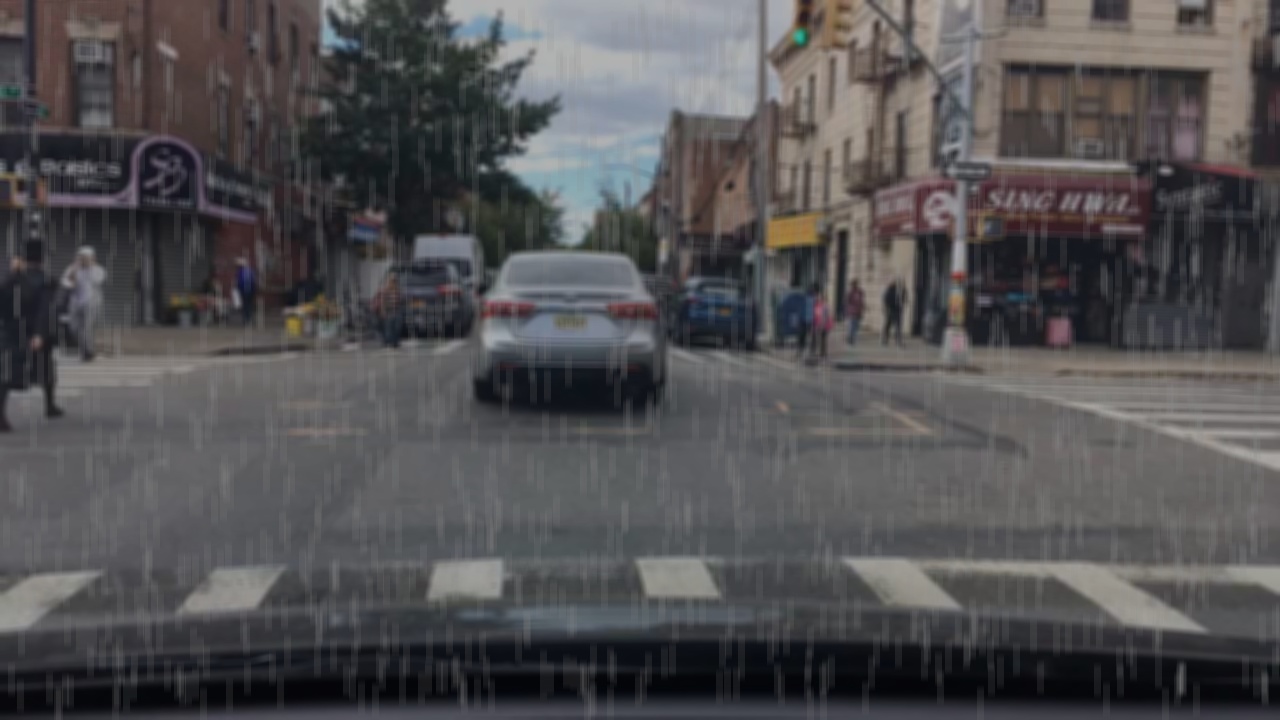}} &
    \raisebox{-.5\height}{\includegraphics[width=0.3\linewidth]{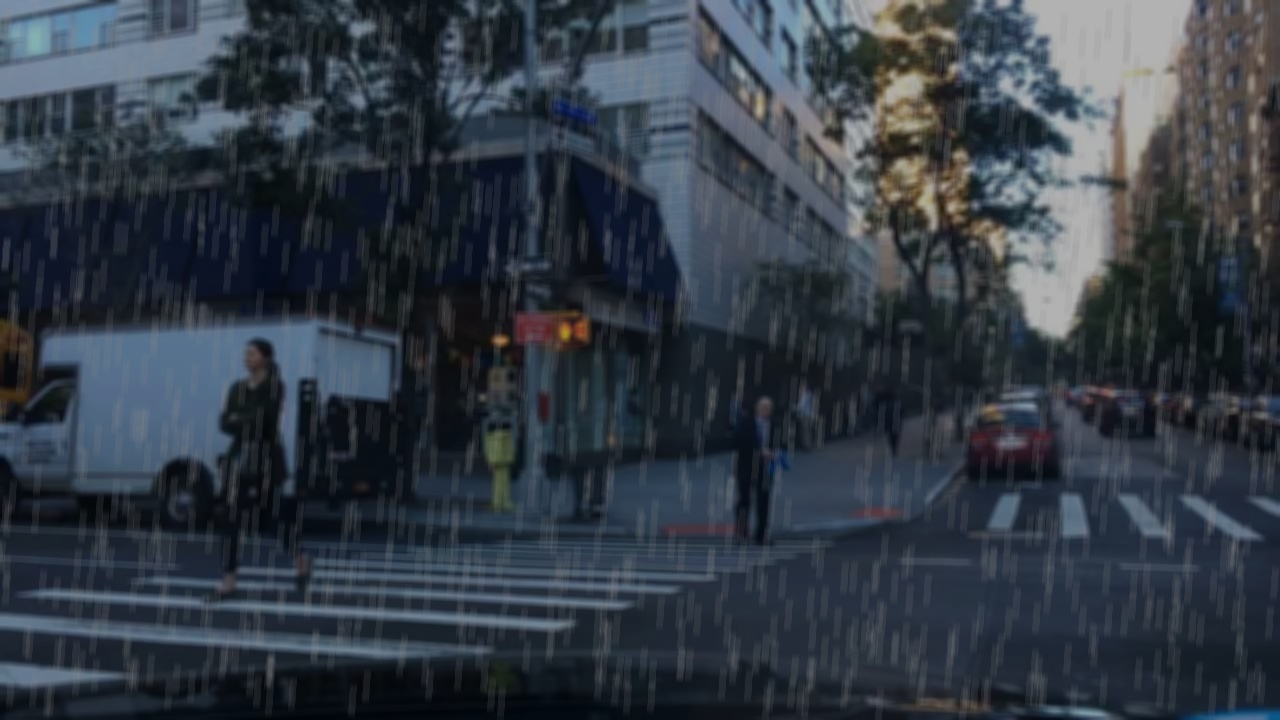}} &
    \raisebox{-.5\height}{\includegraphics[width=0.3\linewidth]{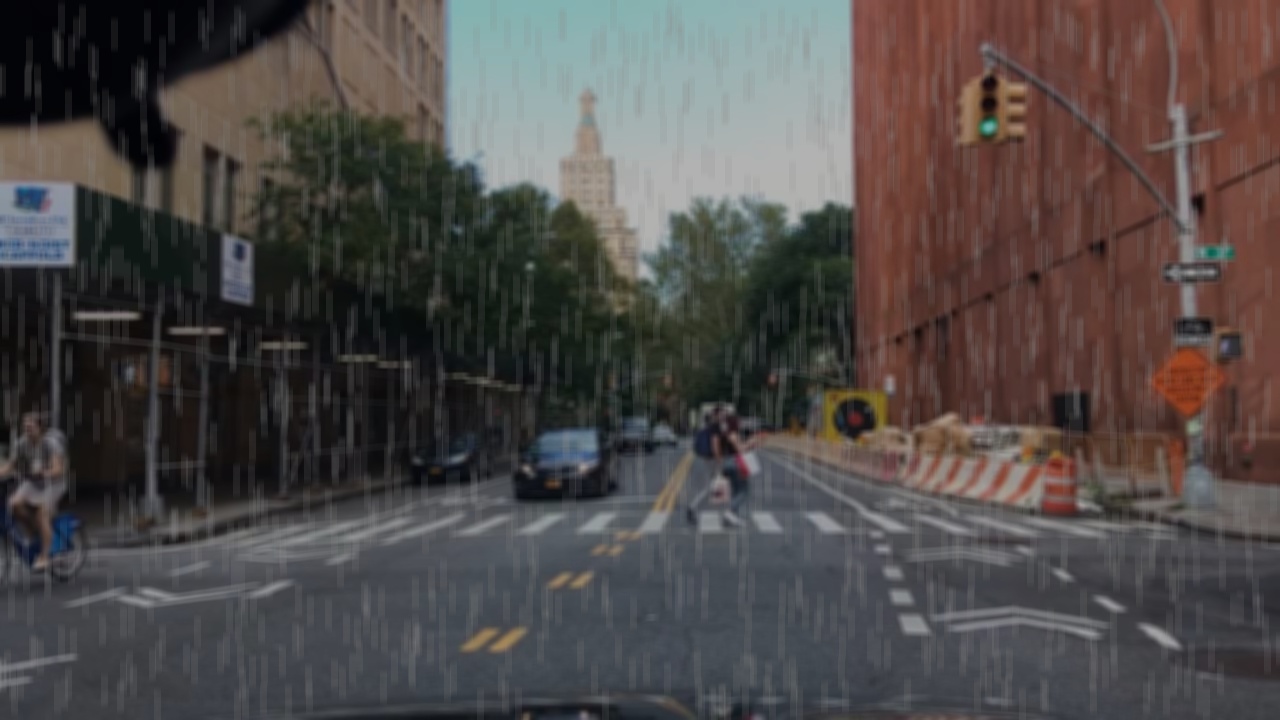}} \\
    
\end{longtable}
\clearpage

\subsection{Fog}
\begin{longtable}{cccc}
    
    \rotatebox[origin=c]{90}{\textbf{ }} & \textbf{0a56c2e8-e46ca9b7} & \textbf{00067cfb-e535423e} & \textbf{0492b183-725063b0} \\
    
    \rotatebox[origin=c]{90}{Original Image} &
    \raisebox{-.5\height}{\includegraphics[width=0.3\linewidth]{augmentations/0a56c2e8-e46ca9b7/augmentation_src.jpg}} &
    \raisebox{-.5\height}{\includegraphics[width=0.3\linewidth]{augmentations/00067cfb-e535423e/augmentation_src.jpg}} &
    \raisebox{-.5\height}{\includegraphics[width=0.3\linewidth]{augmentations/0492b183-725063b0/augmentation_src.jpg}} \\
    
    \rotatebox[origin=c]{90}{A-BDD: Dense Fog 1} &
    \raisebox{-.5\height}{\includegraphics[width=0.3\linewidth]{augmentations/0a56c2e8-e46ca9b7/dense_fog_1.jpg}} &
    \raisebox{-.5\height}{\includegraphics[width=0.3\linewidth]{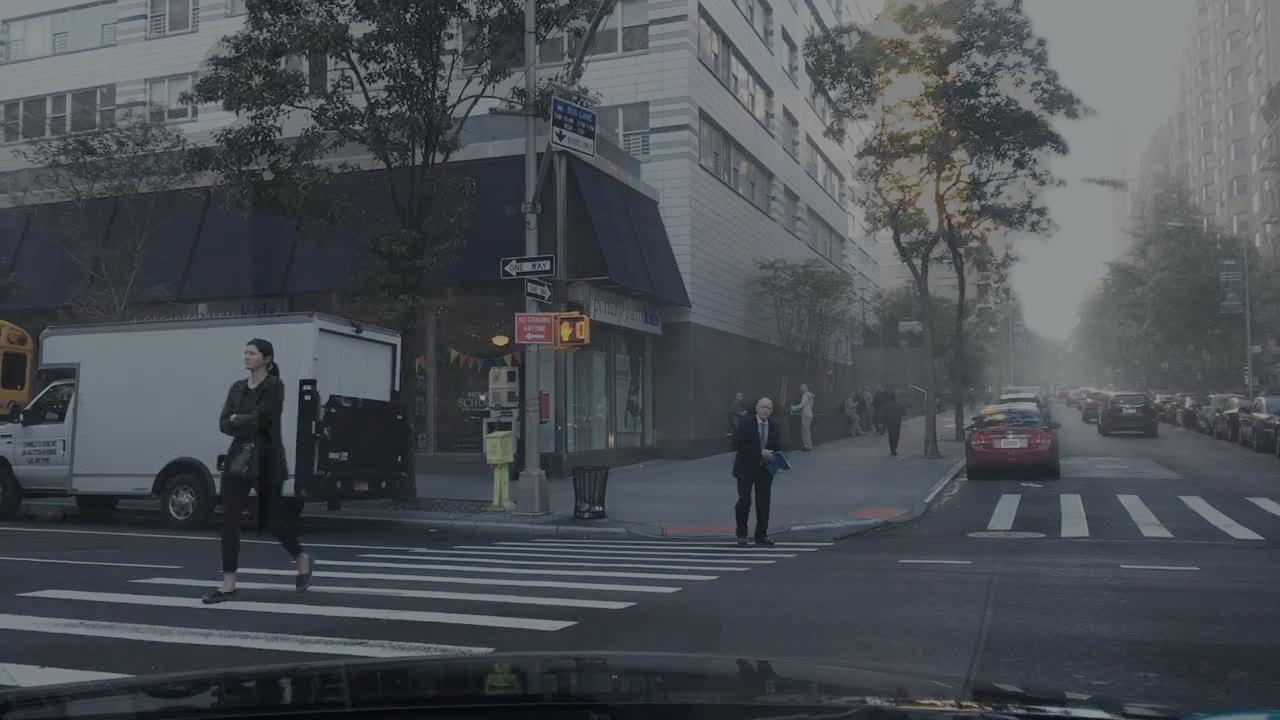}} &
    \raisebox{-.5\height}{\includegraphics[width=0.3\linewidth]{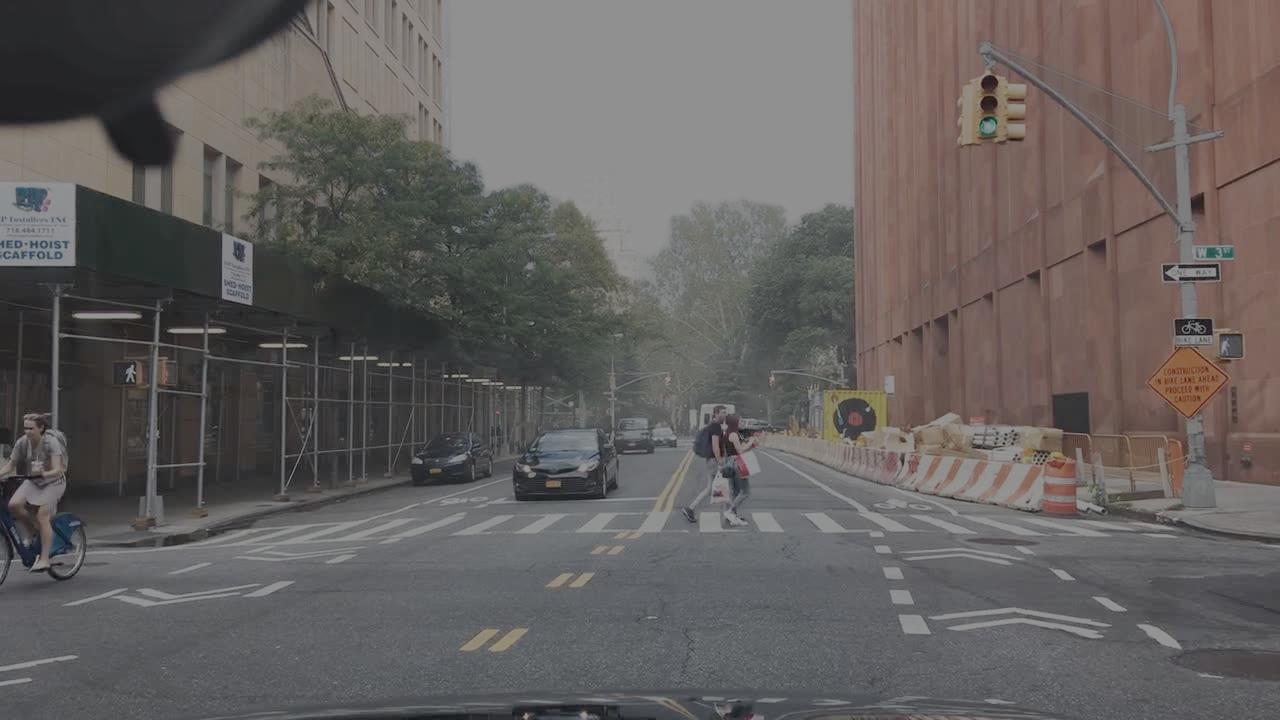}} \\
    
    \rotatebox[origin=c]{90}{A-BDD: Dense Fog 2} &
    \raisebox{-.5\height}{\includegraphics[width=0.3\linewidth]{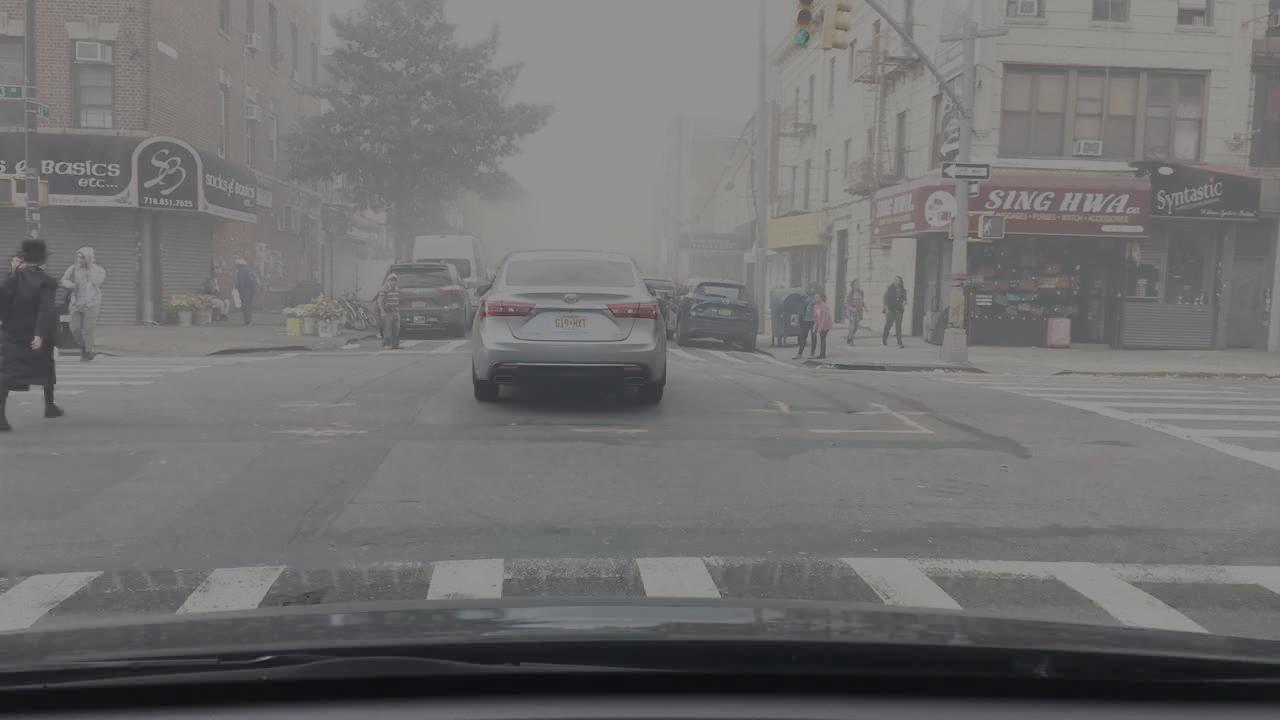}} &
    \raisebox{-.5\height}{\includegraphics[width=0.3\linewidth]{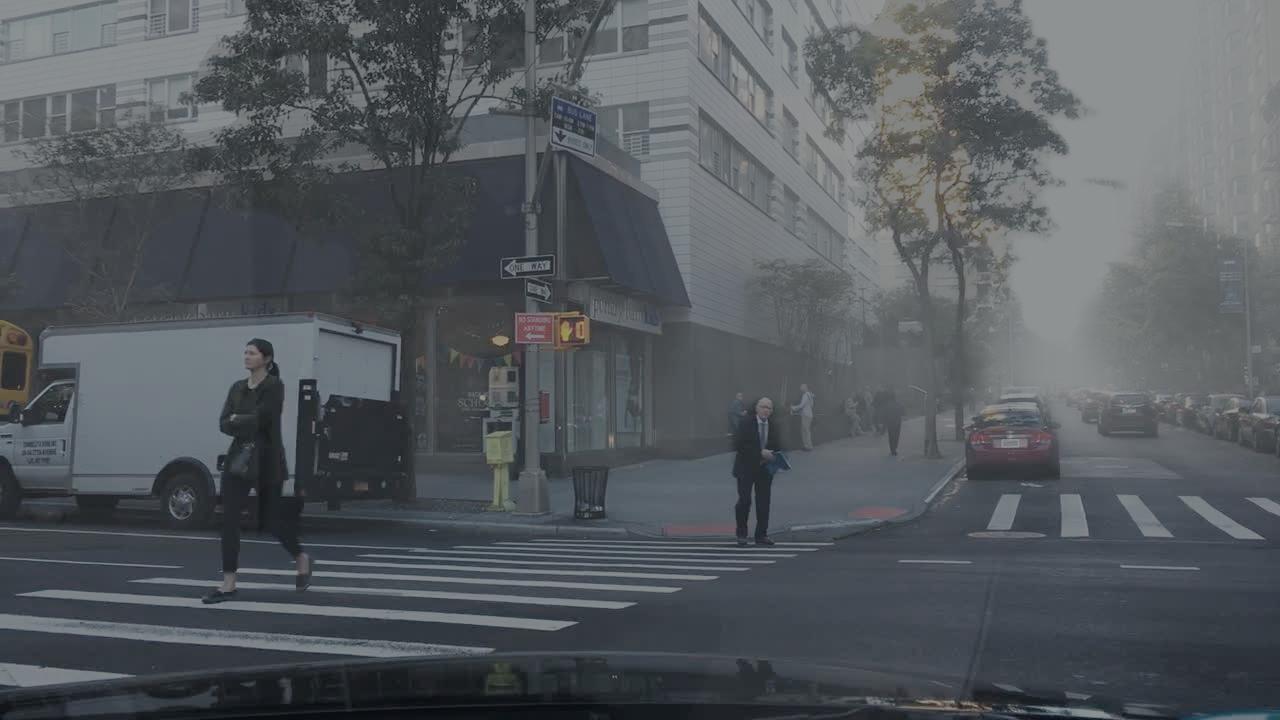}} &
    \raisebox{-.5\height}{\includegraphics[width=0.3\linewidth]{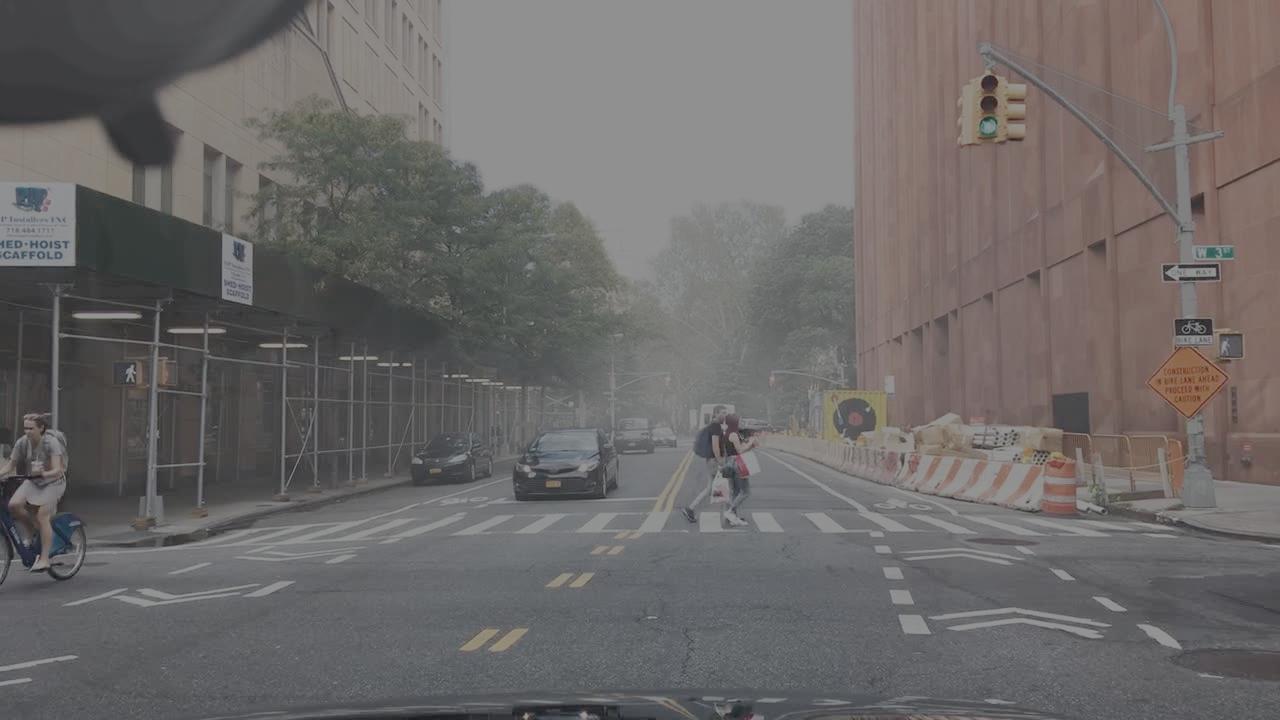}} \\
    
    \rotatebox[origin=c]{90}{A-BDD: Dense Fog 4} &
    \raisebox{-.5\height}{\includegraphics[width=0.3\linewidth]{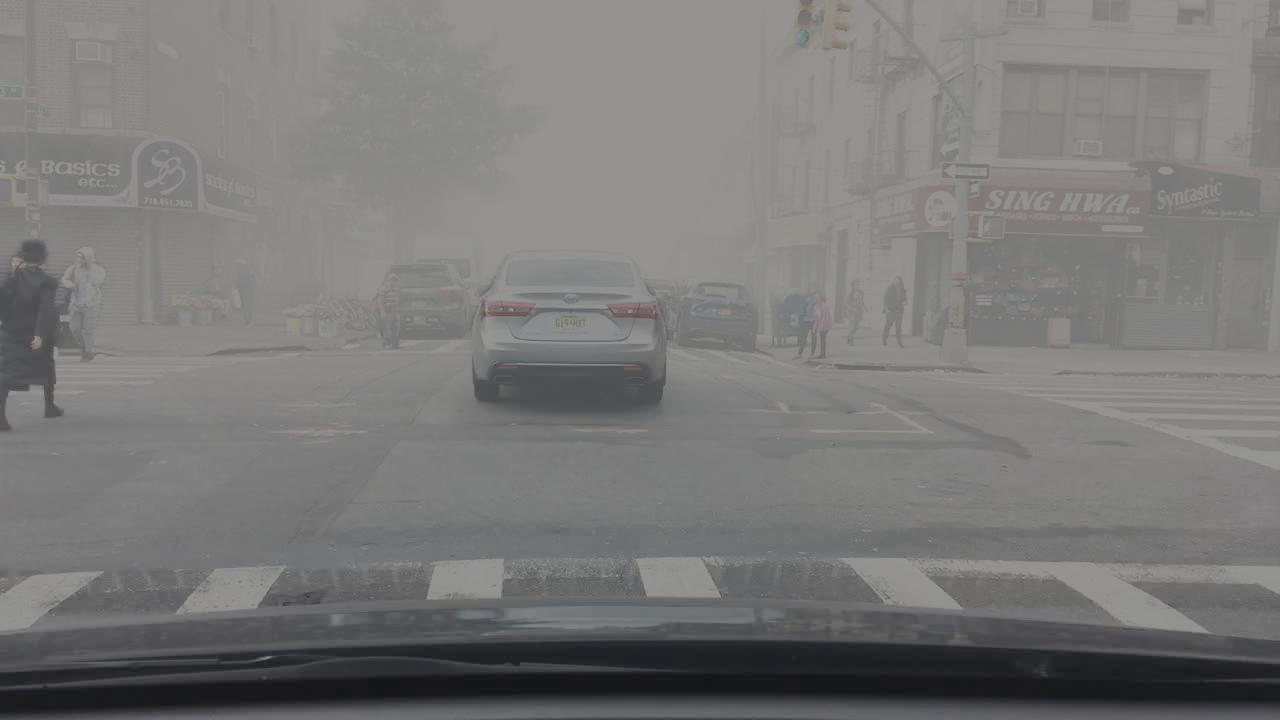}} &
    \raisebox{-.5\height}{\includegraphics[width=0.3\linewidth]{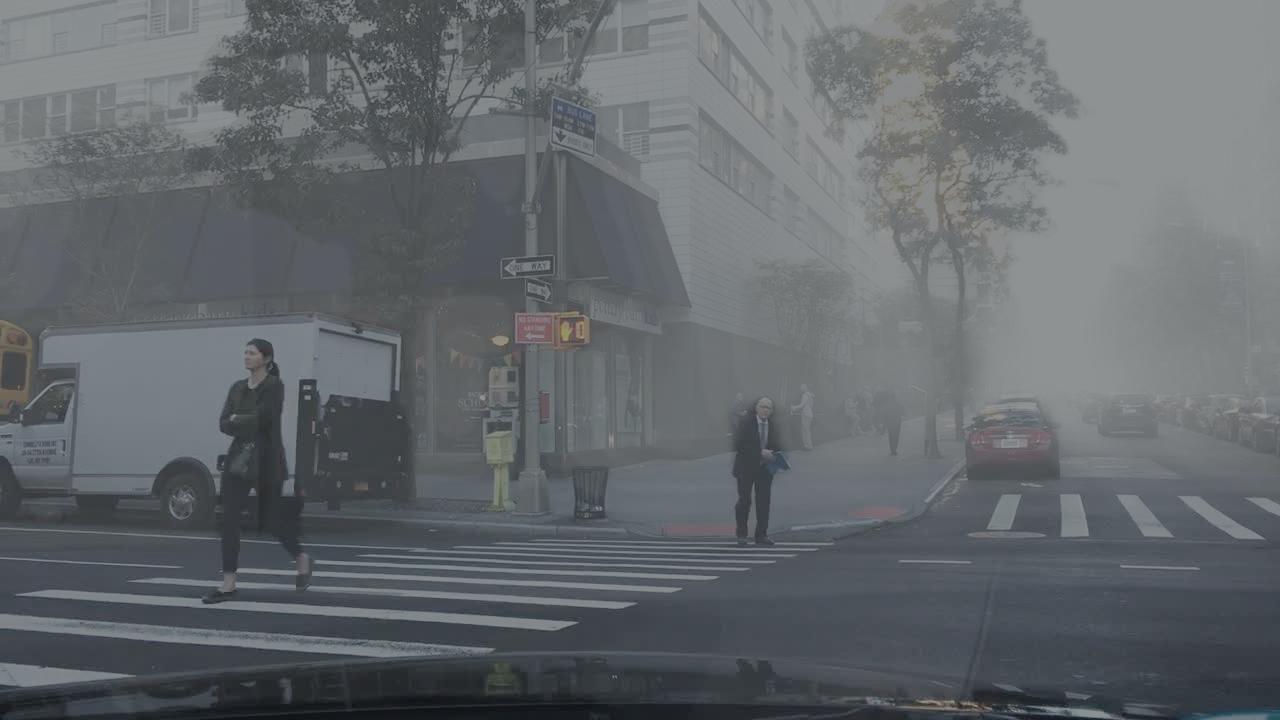}} &
    \raisebox{-.5\height}{\includegraphics[width=0.3\linewidth]{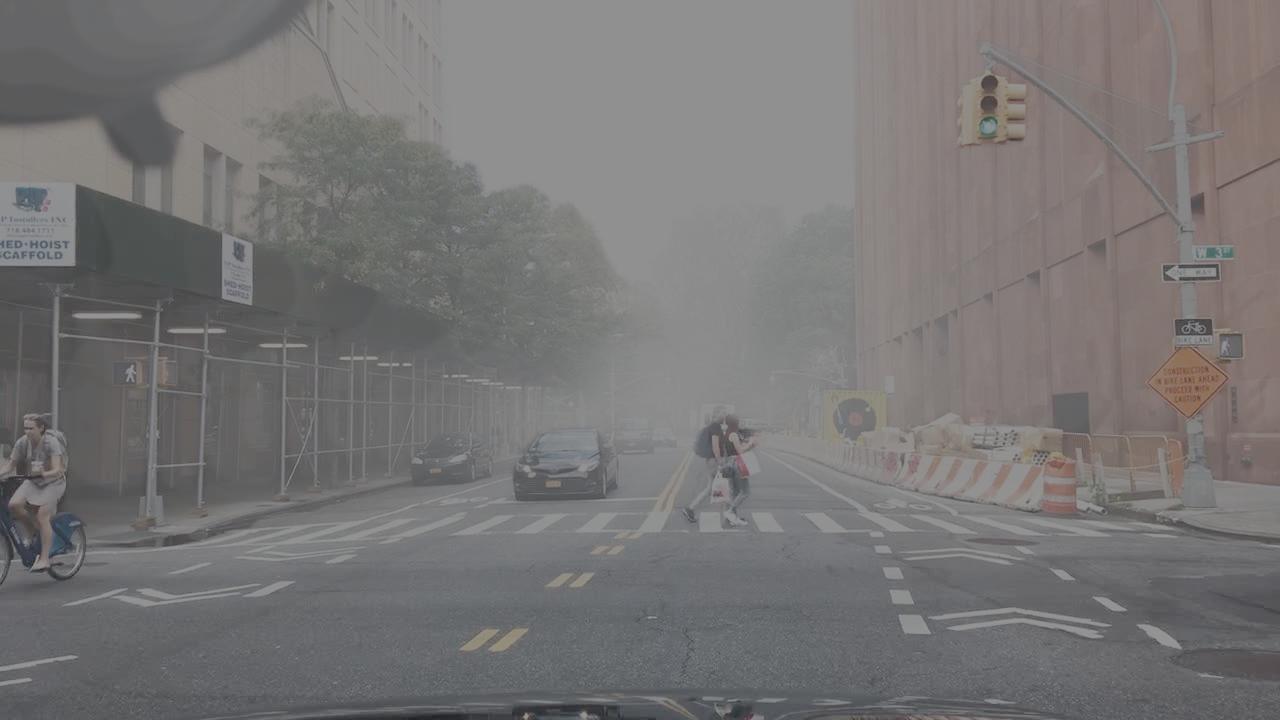}} \\
    
    \rotatebox[origin=c]{90}{Albumentations: Fog 1} &
    \raisebox{-.5\height}{\includegraphics[width=0.3\linewidth]{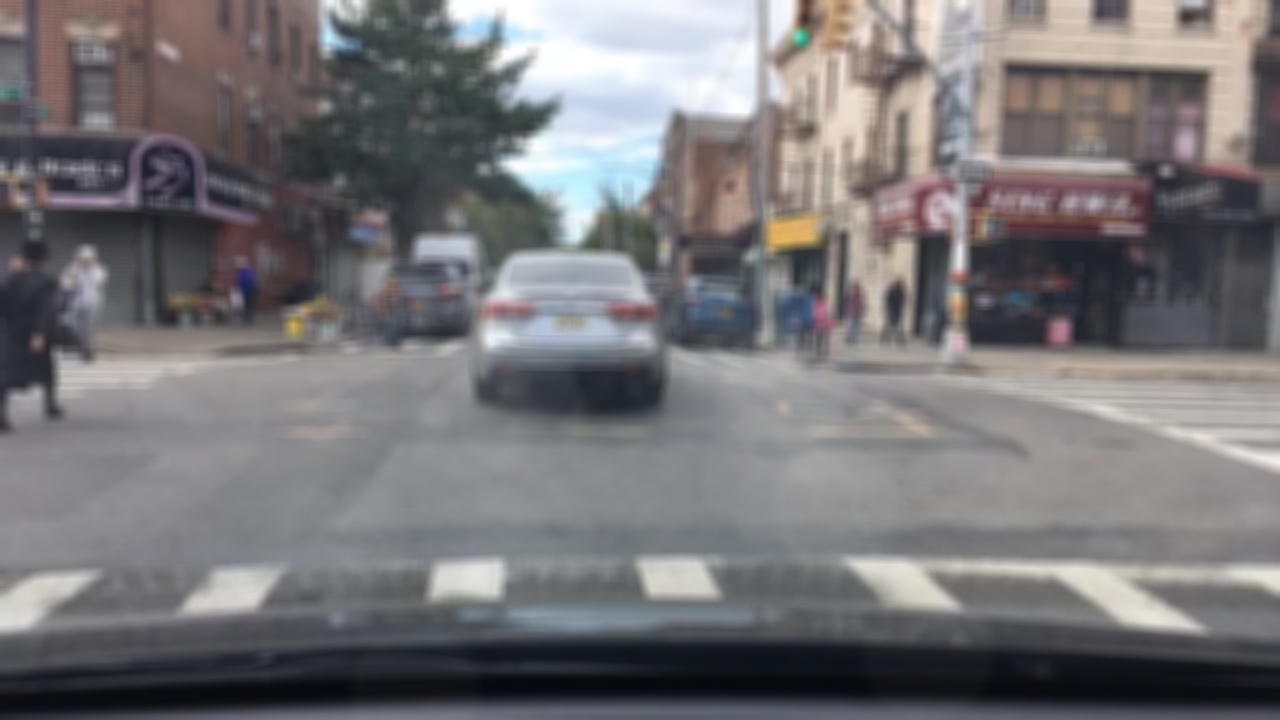}} &
    \raisebox{-.5\height}{\includegraphics[width=0.3\linewidth]{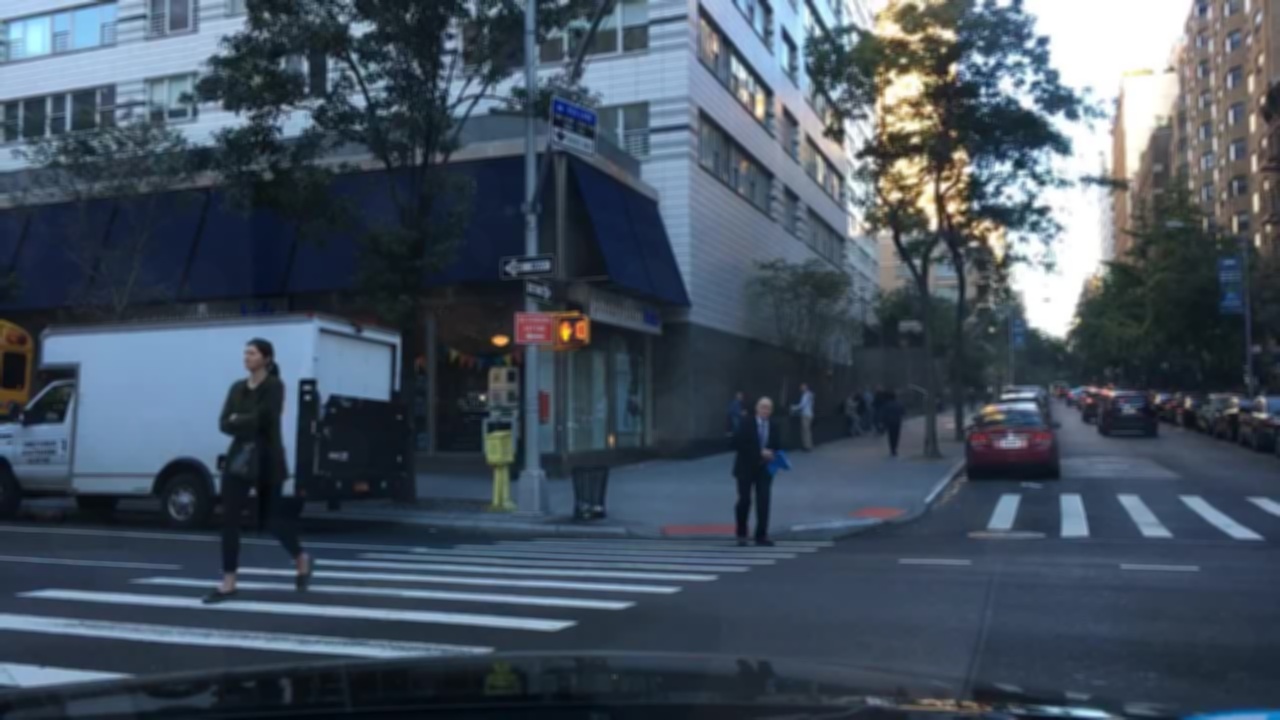}} &
    \raisebox{-.5\height}{\includegraphics[width=0.3\linewidth]{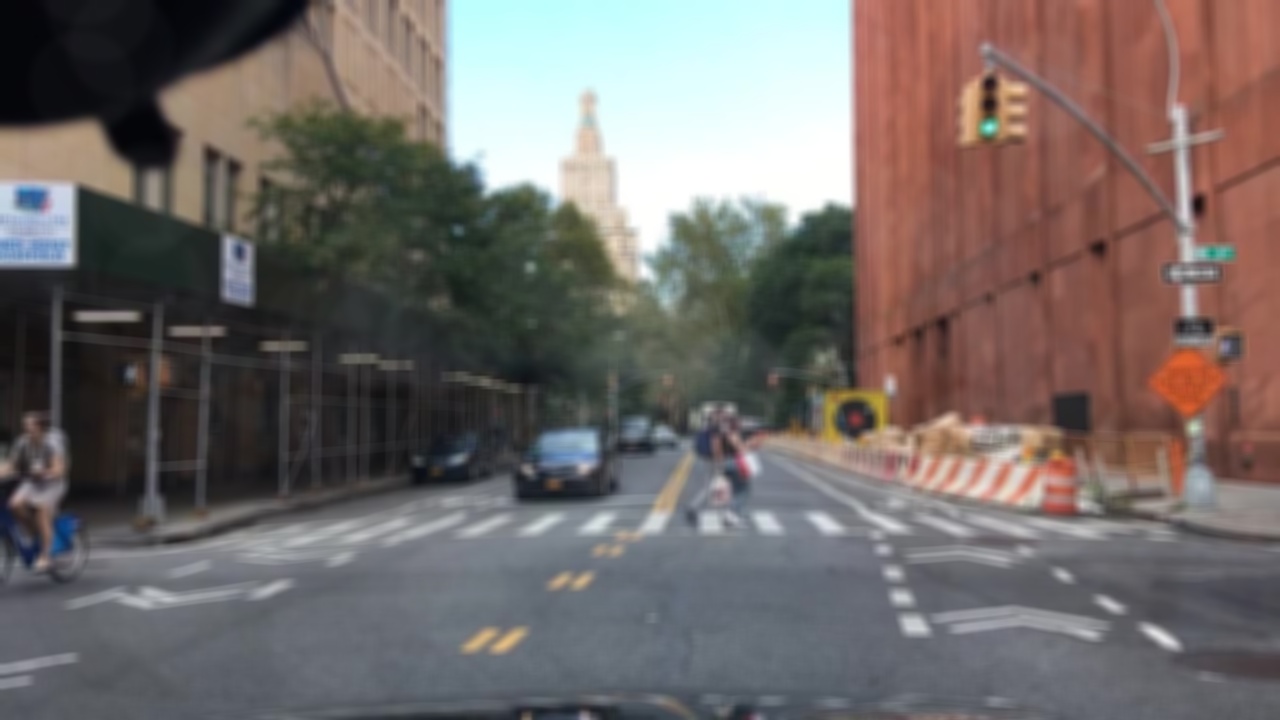}} \\
    
    \rotatebox[origin=c]{90}{Albumentations: Fog 2} &
    \raisebox{-.5\height}{\includegraphics[width=0.3\linewidth]{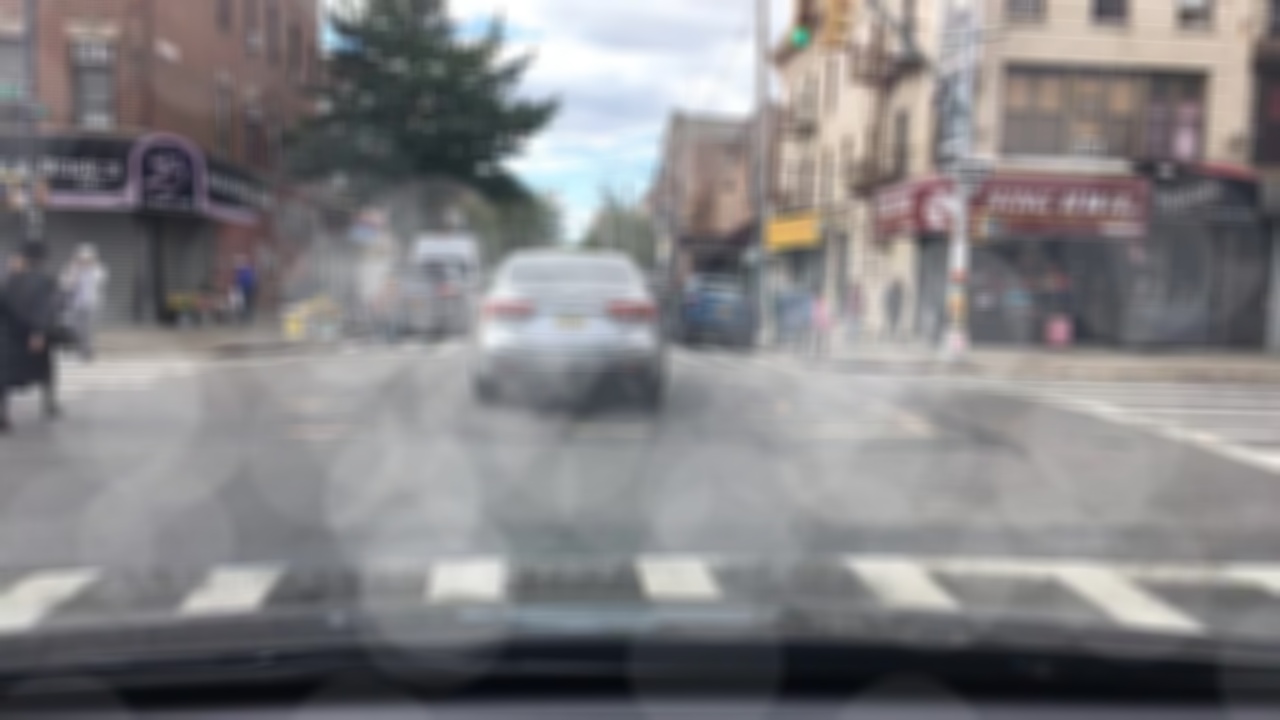}} &
    \raisebox{-.5\height}{\includegraphics[width=0.3\linewidth]{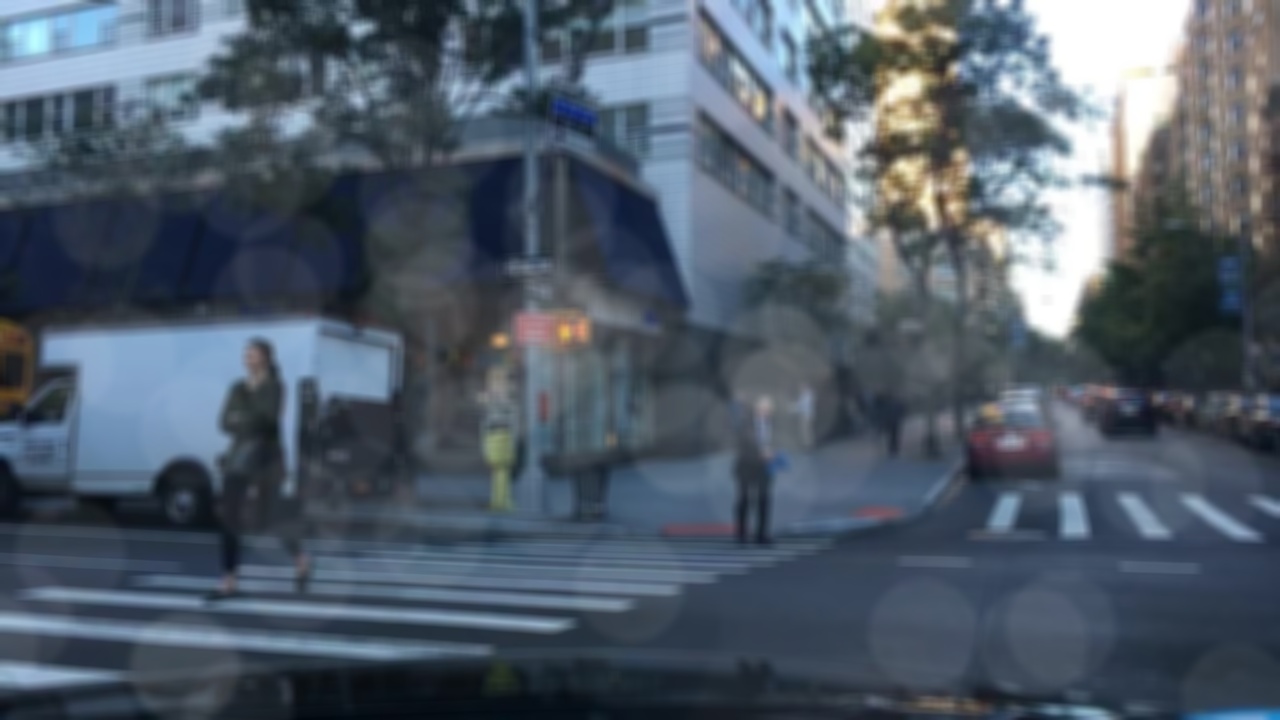}} &
    \raisebox{-.5\height}{\includegraphics[width=0.3\linewidth]{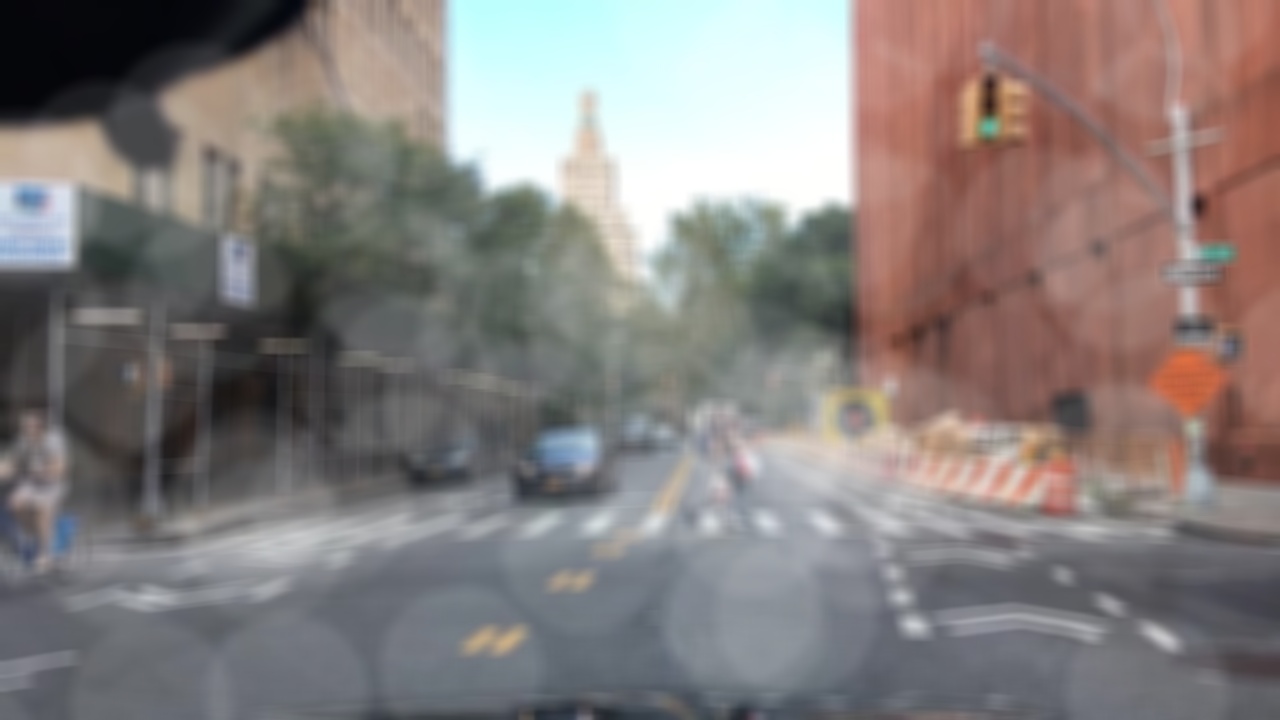}} \\
    
\end{longtable}

\subsection{Sunglare / Lighting}
\begin{longtable}{cccc}
    
    \rotatebox[origin=c]{90}{\textbf{ }} & \textbf{0a56c2e8-e46ca9b7} & \textbf{00067cfb-e535423e} & \textbf{0492b183-725063b0} \\
    
    \rotatebox[origin=c]{90}{Original Image} &
    \raisebox{-.5\height}{\includegraphics[width=0.3\linewidth]{augmentations/0a56c2e8-e46ca9b7/augmentation_src.jpg}} &
    \raisebox{-.5\height}{\includegraphics[width=0.3\linewidth]{augmentations/00067cfb-e535423e/augmentation_src.jpg}} &
    \raisebox{-.5\height}{\includegraphics[width=0.3\linewidth]{augmentations/0492b183-725063b0/augmentation_src.jpg}} \\
    
    \rotatebox[origin=c]{90}{A-BDD: Shadow/Sunglare 1} &
    \raisebox{-.5\height}{\includegraphics[width=0.3\linewidth]{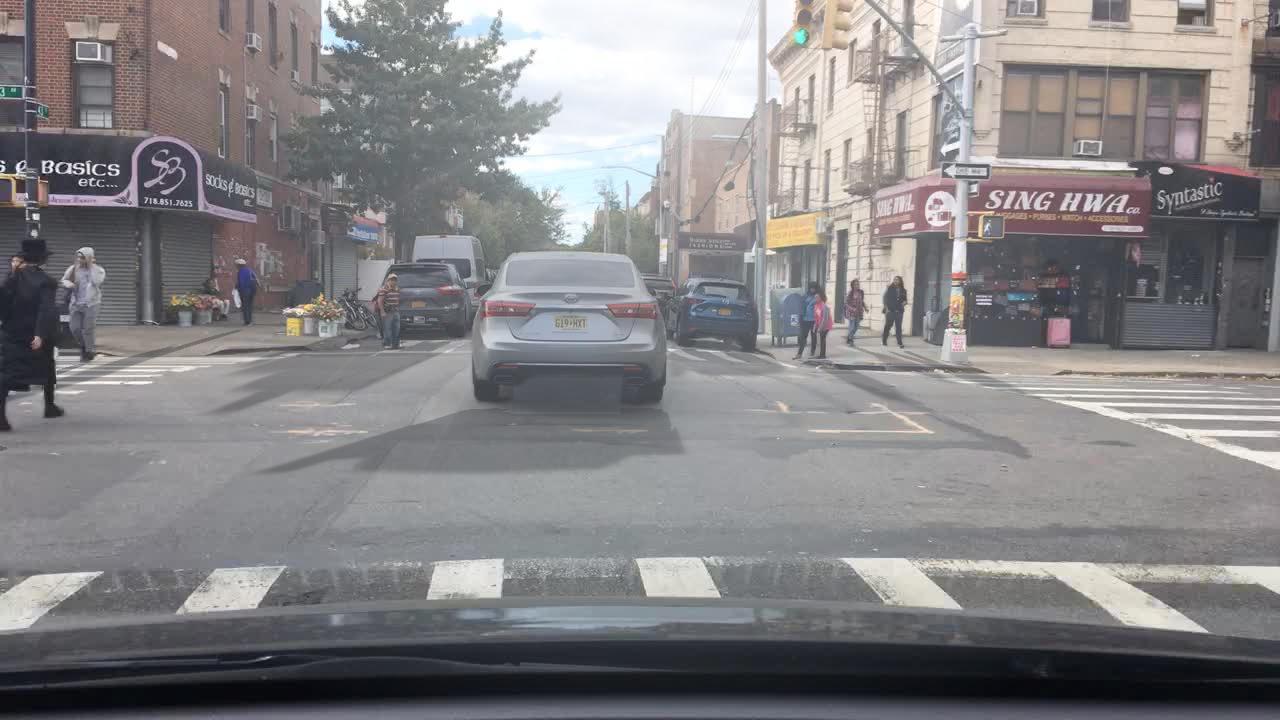}} &
    \raisebox{-.5\height}{\includegraphics[width=0.3\linewidth]{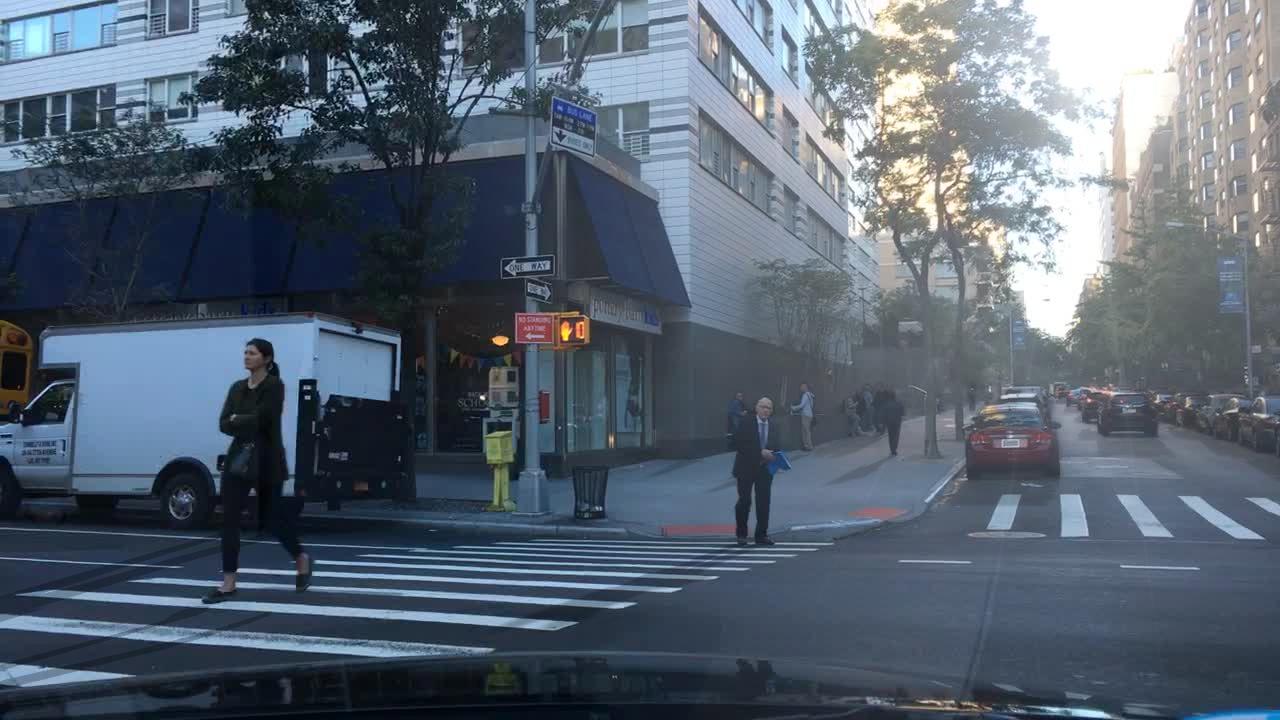}} &
    \raisebox{-.5\height}{\includegraphics[width=0.3\linewidth]{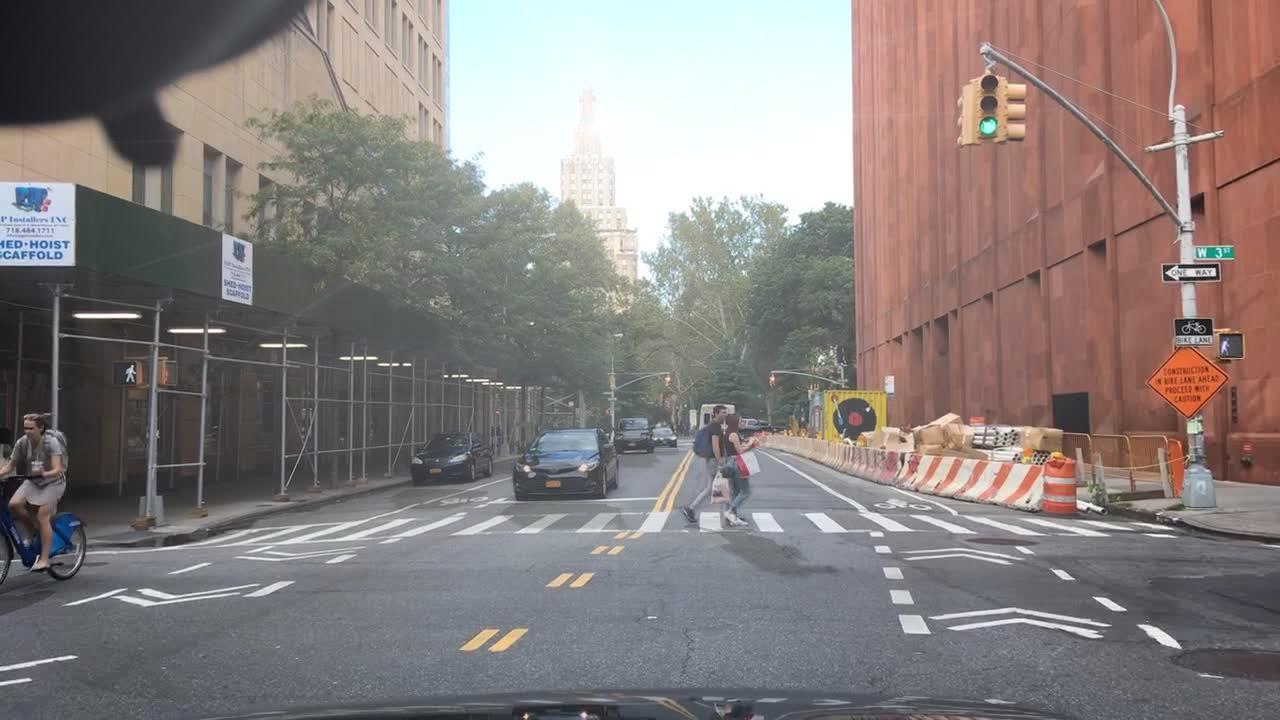}} \\
    
    \rotatebox[origin=c]{90}{A-BDD: Shadow/Sunglare 3} &
    \raisebox{-.5\height}{\includegraphics[width=0.3\linewidth]{augmentations/0a56c2e8-e46ca9b7/shadow_3.jpg}} &
    \raisebox{-.5\height}{\includegraphics[width=0.3\linewidth]{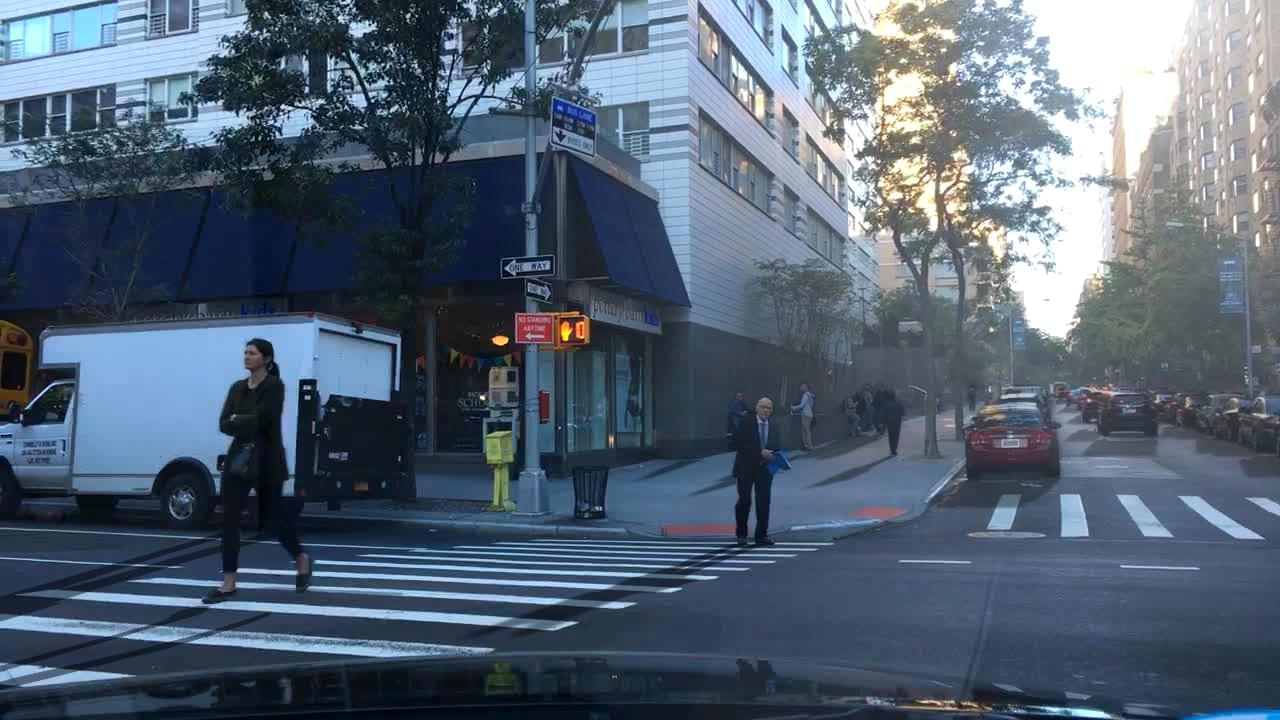}} &
    \raisebox{-.5\height}{\includegraphics[width=0.3\linewidth]{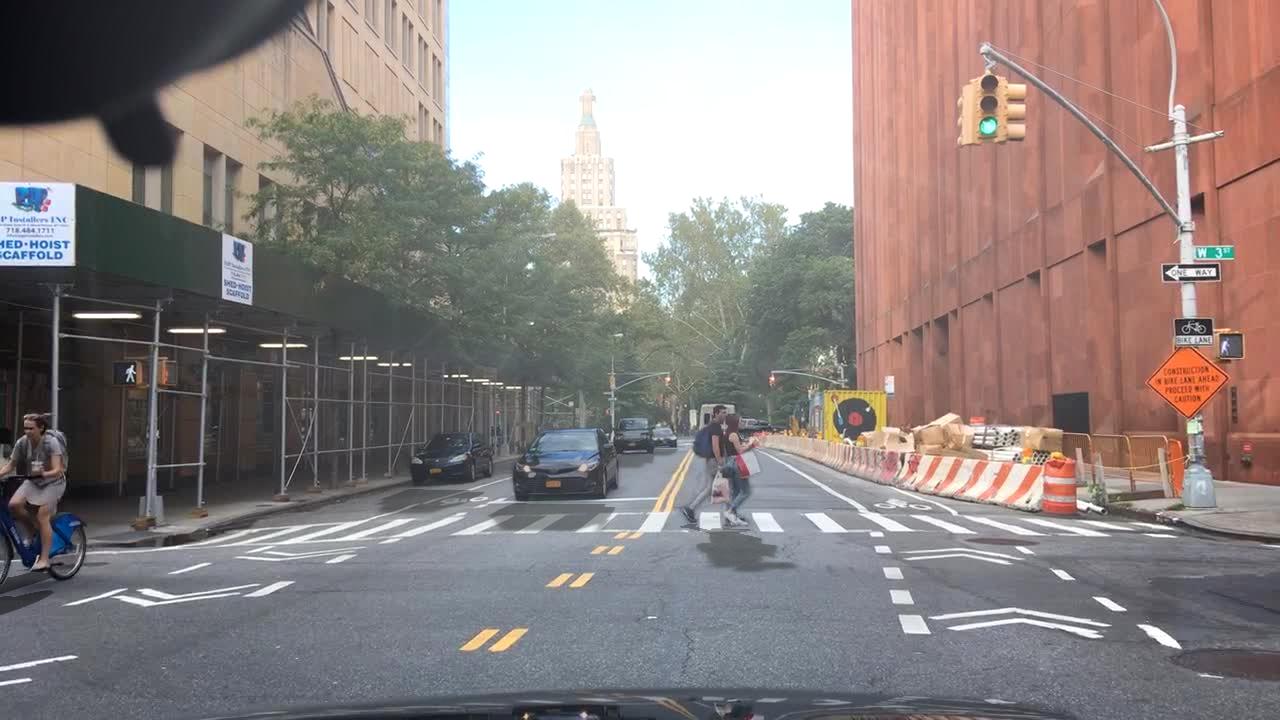}} \\
    
    \rotatebox[origin=c]{90}{A-BDD: Shadow/Sunglare 5} &
    \raisebox{-.5\height}{\includegraphics[width=0.3\linewidth]{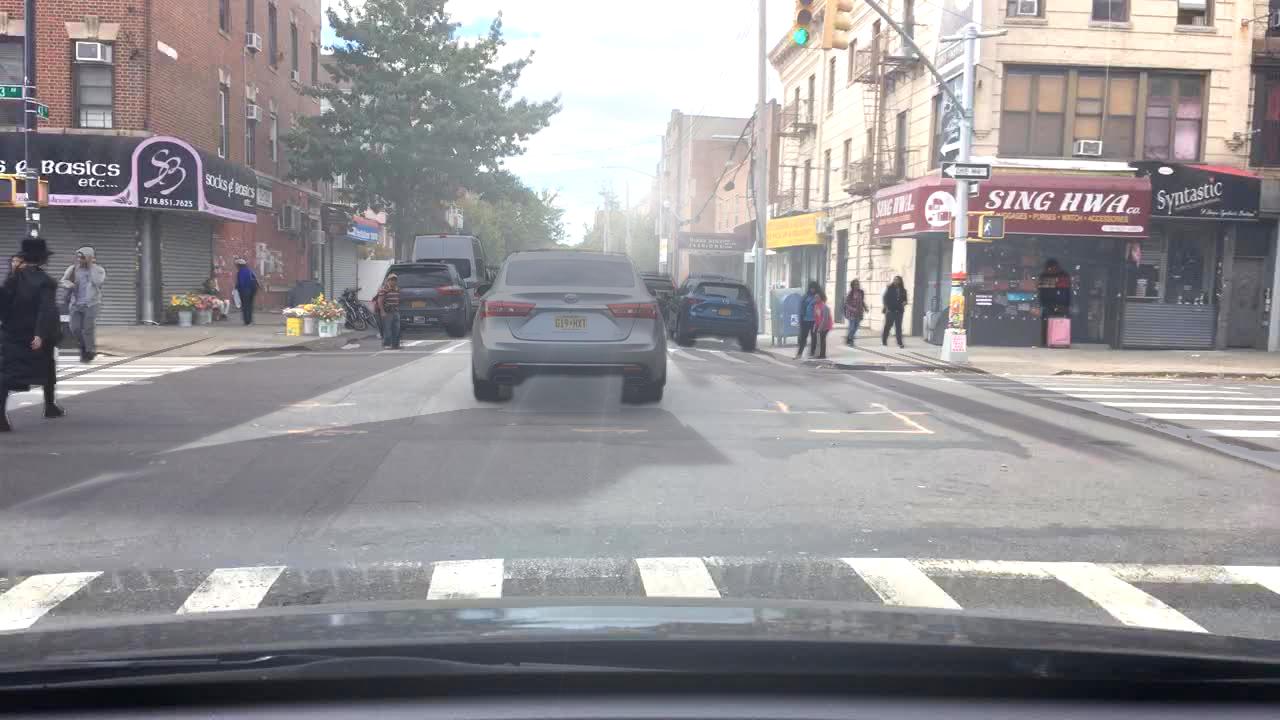}} &
    \raisebox{-.5\height}{\includegraphics[width=0.3\linewidth]{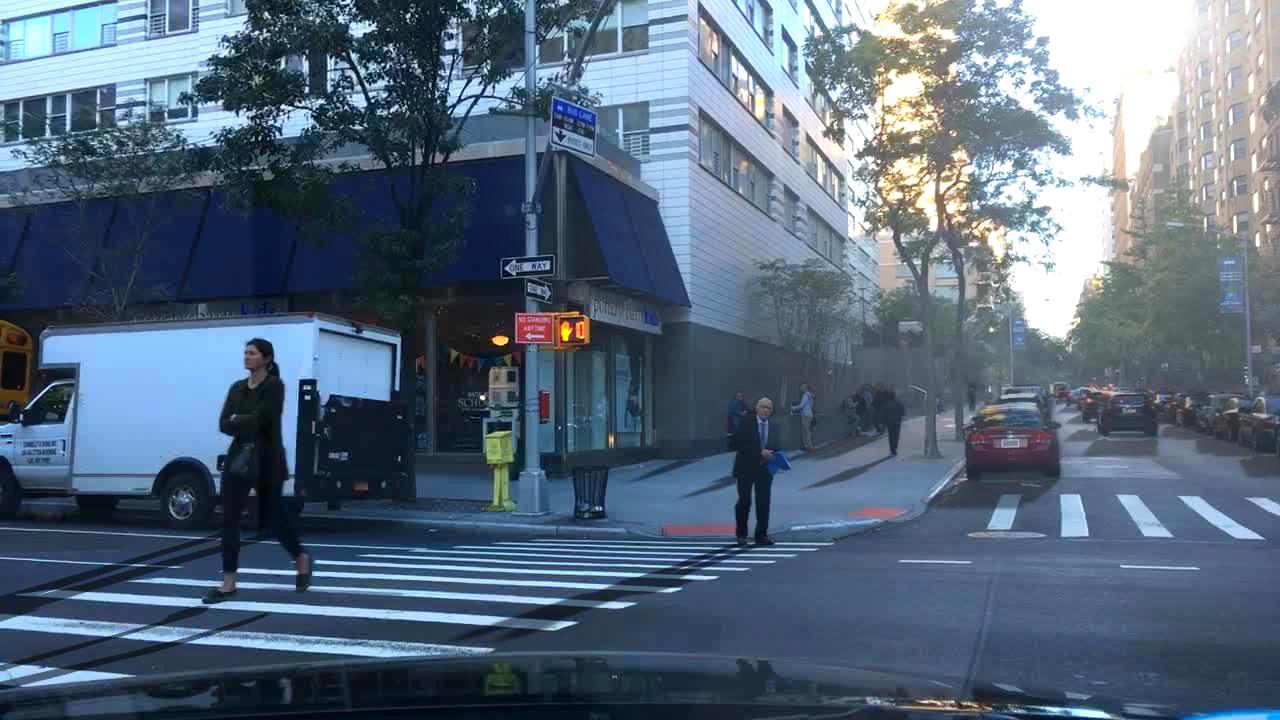}} &
    \raisebox{-.5\height}{\includegraphics[width=0.3\linewidth]{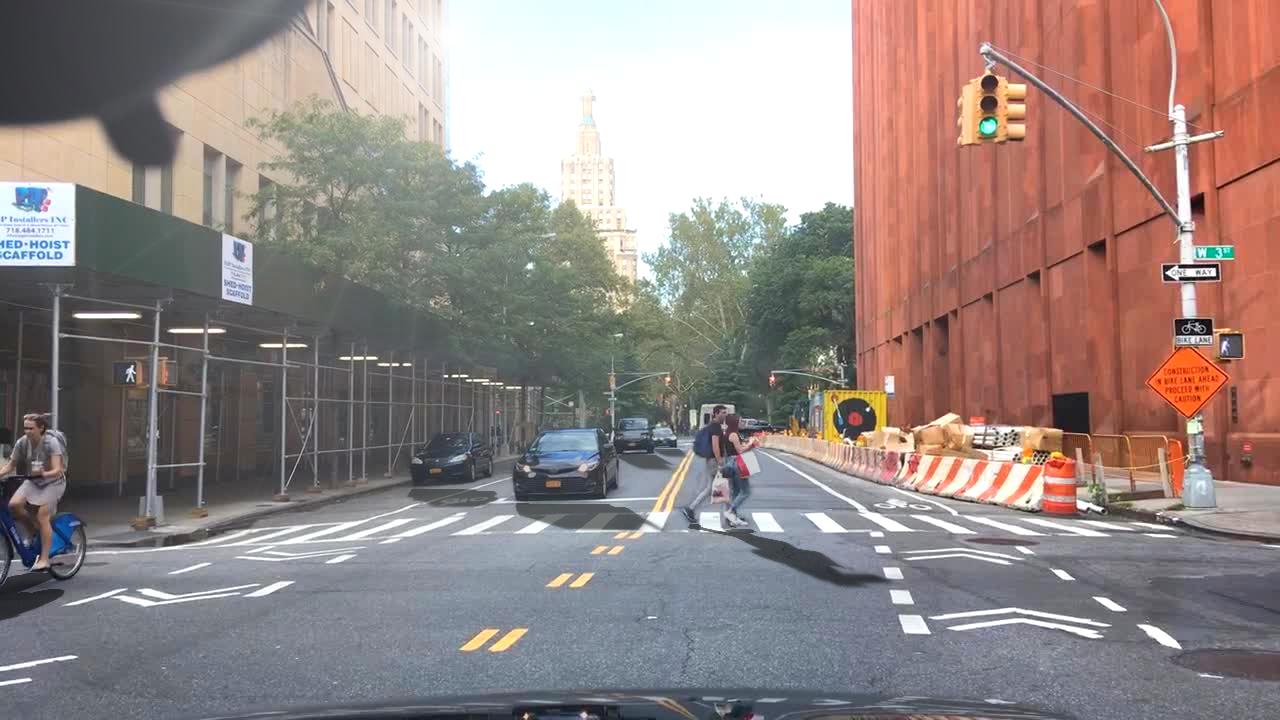}} \\
    
    \rotatebox[origin=c]{90}{Albumentations: Sun 1} &
    \raisebox{-.5\height}{\includegraphics[width=0.3\linewidth]{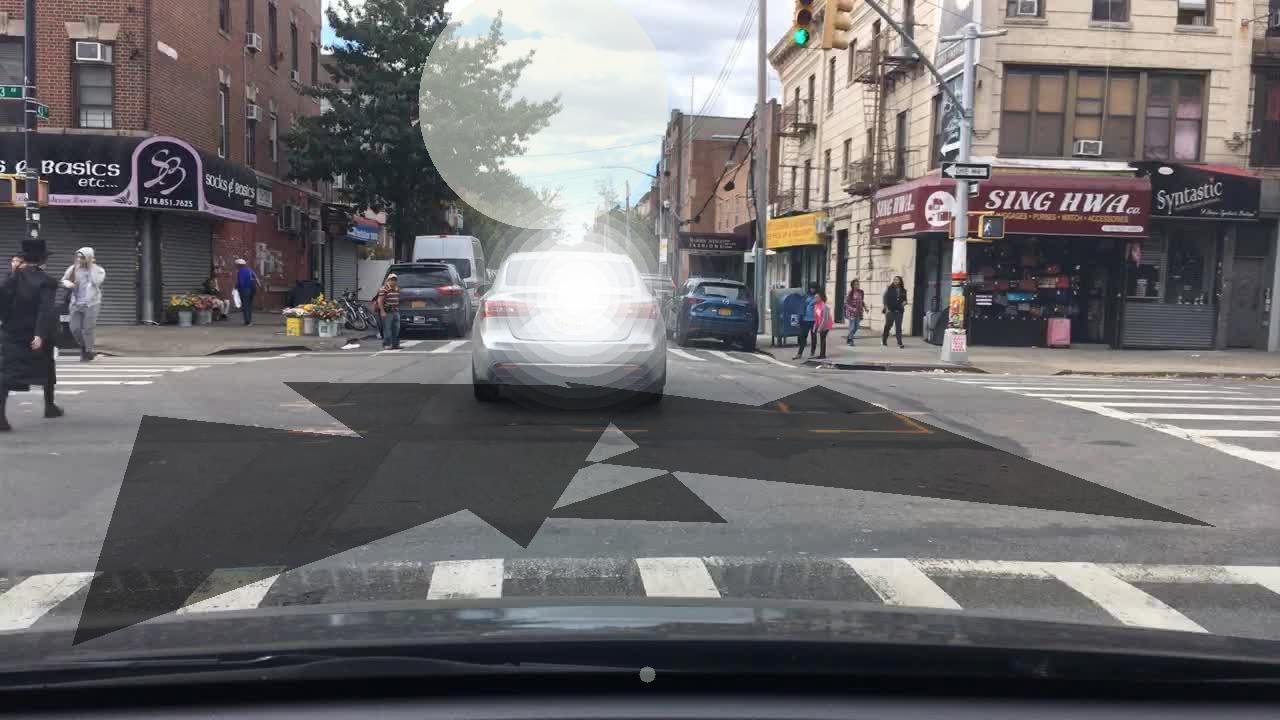}} &
    \raisebox{-.5\height}{\includegraphics[width=0.3\linewidth]{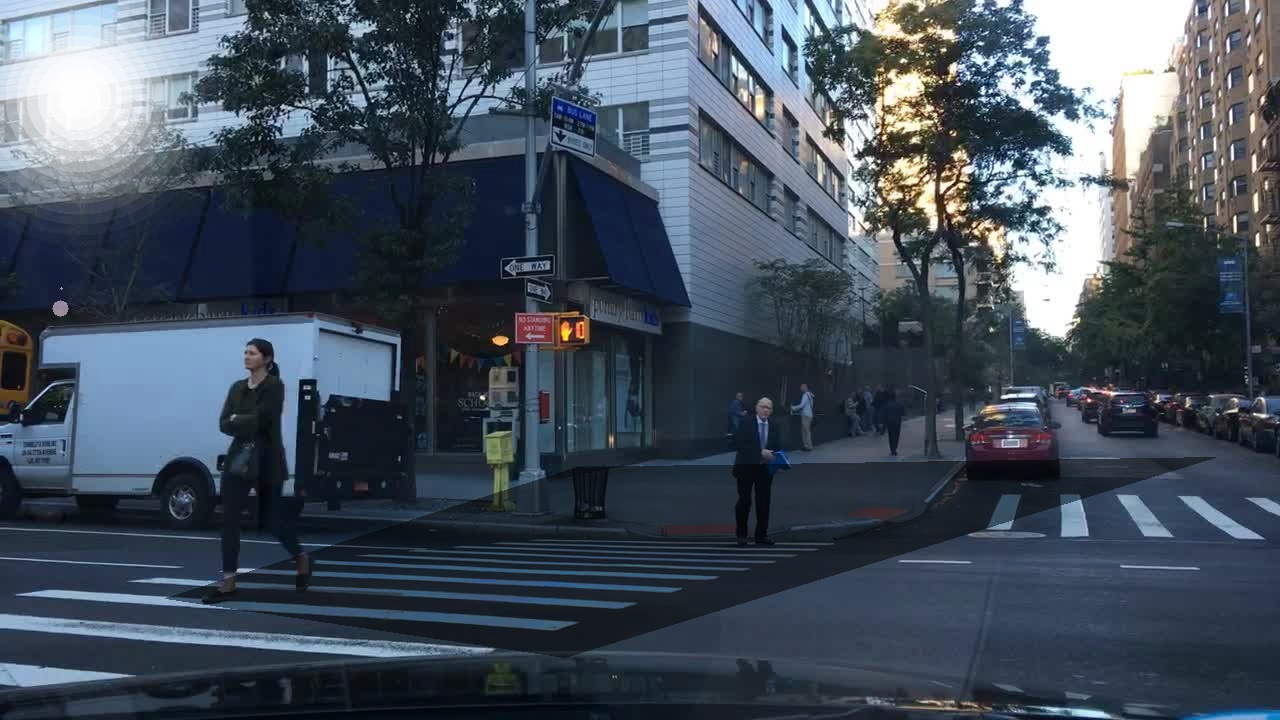}} &
    \raisebox{-.5\height}{\includegraphics[width=0.3\linewidth]{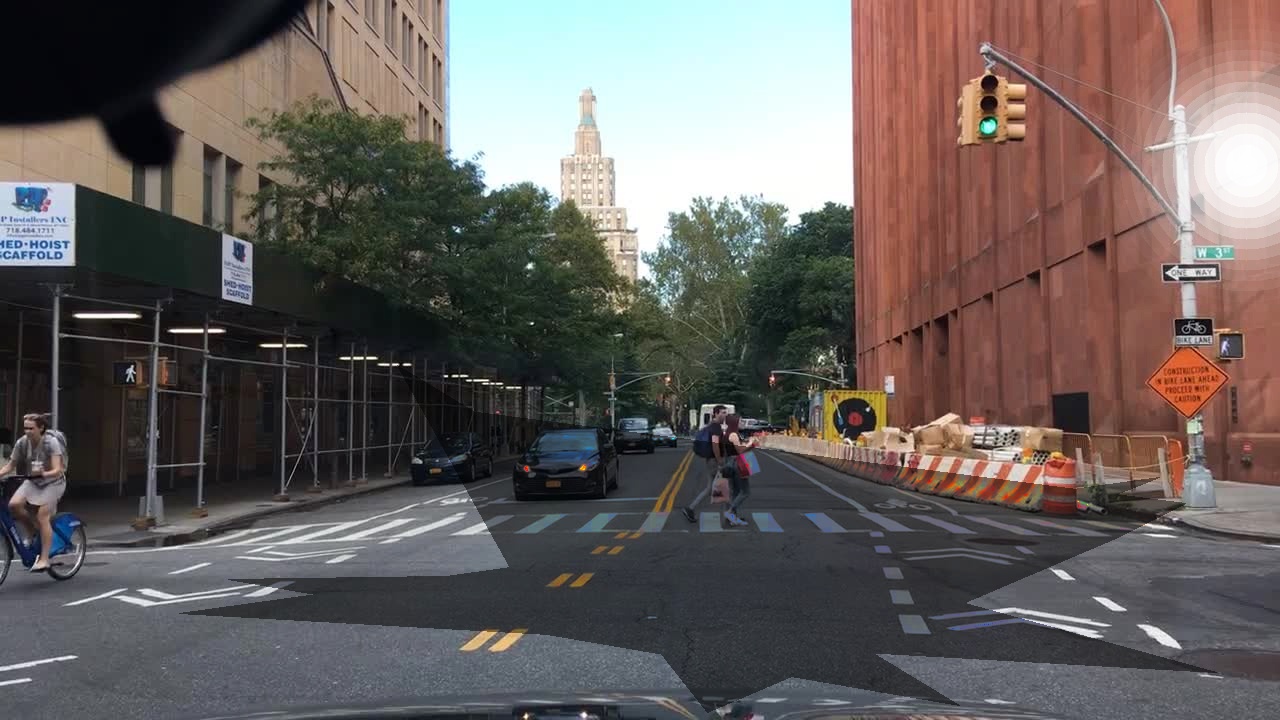}} \\
    
    \rotatebox[origin=c]{90}{Albumentations: Sun 2} &
    \raisebox{-.5\height}{\includegraphics[width=0.3\linewidth]{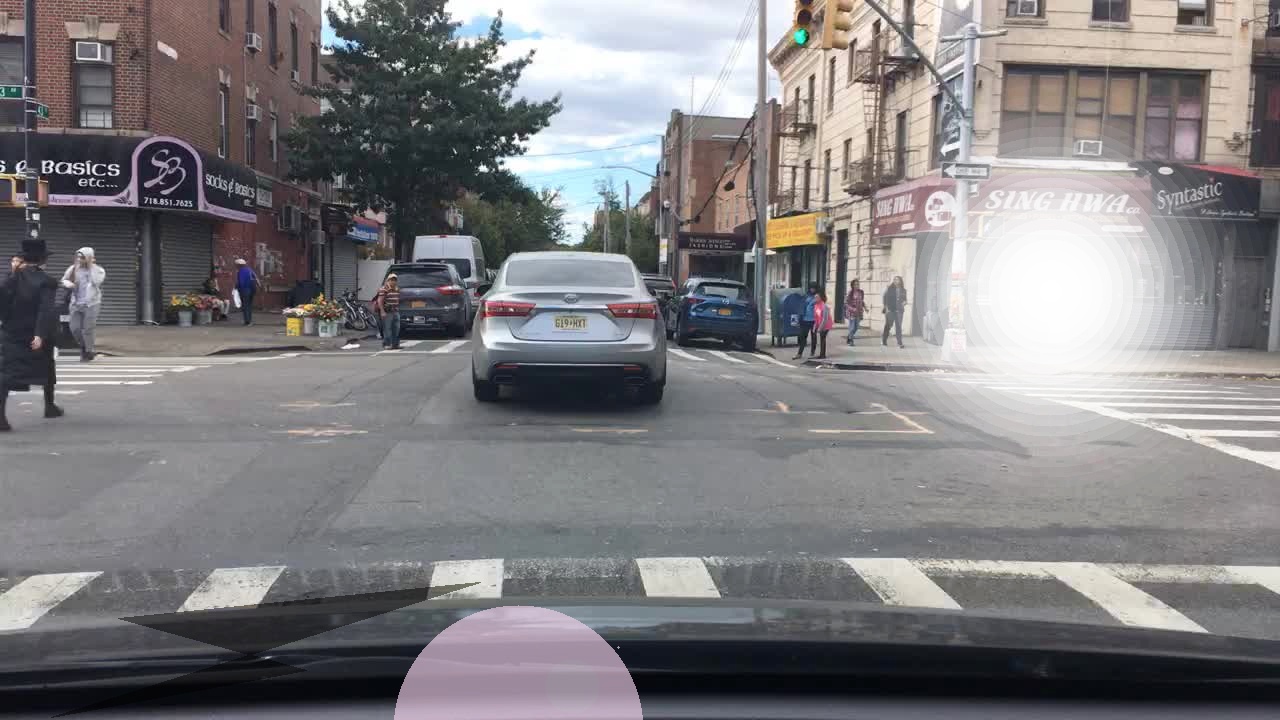}} &
    \raisebox{-.5\height}{\includegraphics[width=0.3\linewidth]{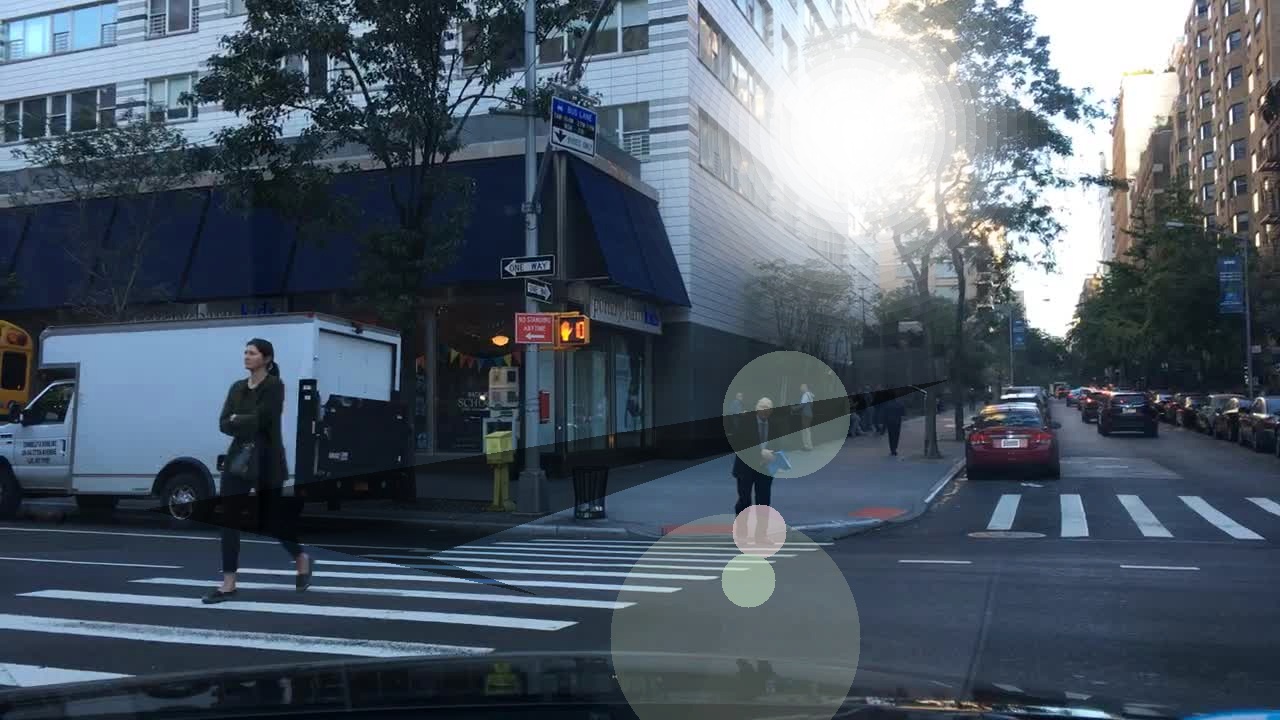}} &
    \raisebox{-.5\height}{\includegraphics[width=0.3\linewidth]{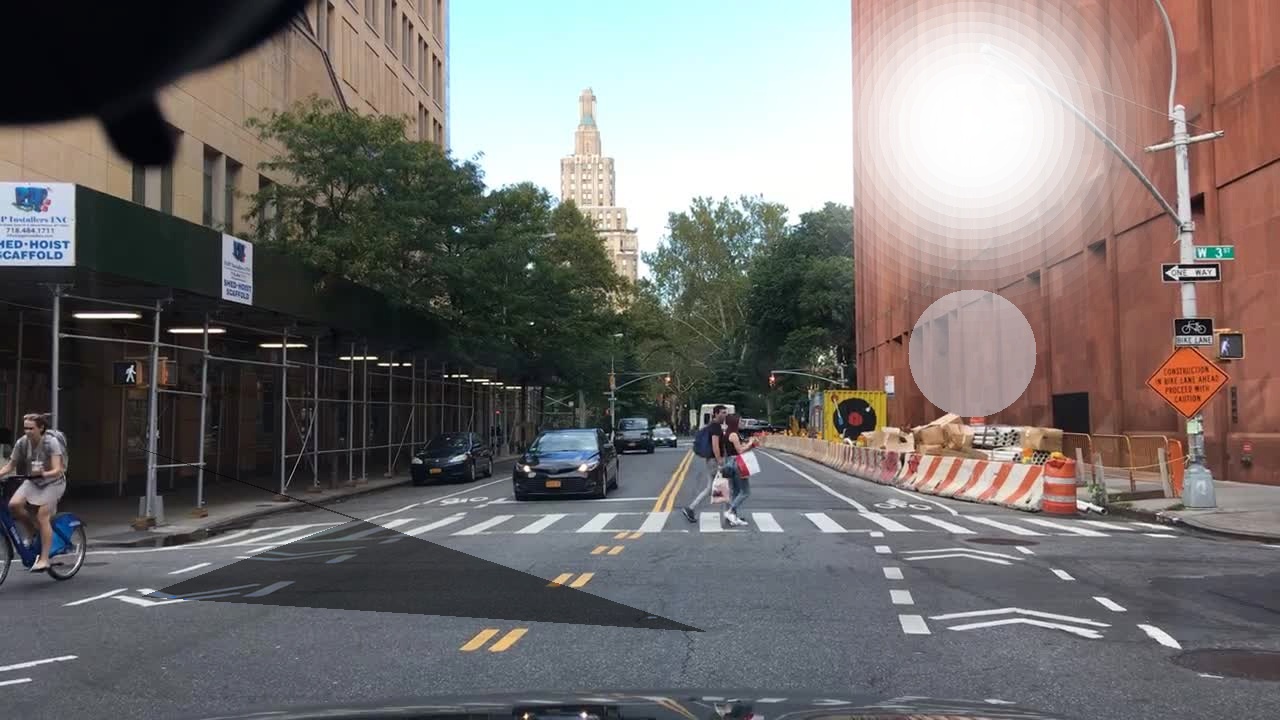}} \\
    
\end{longtable}
\clearpage

\section{Image Quality Scores} \label{sec: big_image_quality_tables}
\subsection{FID}

\begin{longtable}{c|ccccc|cccc|c}
\caption{FID distances for various augmentations across different datasets and weather conditions. Additionally, we list semantic segmentation retraining results with respect to \textsc{ACDC} rain data.} \label{tab:fid_scores} \\
\toprule
& \multicolumn{5}{c|}{BDD} & \multicolumn{4}{c|}{ACDC} & \multicolumn{1}{c}{Retraining Results} \\
\cline{2-11}
Augmentation & Clear & Fog & Overcast & Rain & Snow & Fog & Rain & Snow & Sun & Rain \\
\midrule
\endfirsthead

\toprule
& \multicolumn{5}{c|}{BDD} & \multicolumn{4}{c|}{ACDC} & \multicolumn{1}{c}{Retraining Results} \\
\cline{2-11}
Augmentation & Clear & Fog & Overcast & Rain & Snow & Fog & Rain & Snow & Sun & Rain \\
\midrule
\endhead

\bottomrule
\endfoot

albu\_fog\_1 & 169.47 & 251.66 & 182.55 & 165.14 & 167.52 & 221.59 & 187.59 & 207.52 & 211.16 & 50.45 \\
albu\_fog\_2 & 108.44 & 186.12 & 111.27 & 111.87 & 113.96 & 162.76 & 157.29 & 163.81 & 159.47 & 50.27 \\
albu\_fog\_3 & 224.16 & 303.32 & 236.87 & 223.01 & 223.19 & 261.03 & 234.68 & 251.74 & 254.67 & 51.32 \\
albu\_fog\_4 & 71.95 & 148.87 & 74.13 & 86.19 & 86.72 & 135.59 & 137.94 & 140.24 & 128.76 & 49.85 \\
albu\_fog\_5 & 78.68 & 159.98 & 82.42 & 87.74 & 92.59 & 144.58 & 139.31 & 145.33 & 132.44 & 50.28 \\
albu\_fog\_6 & 104.68 & 185.63 & 109.80 & 110.14 & 113.05 & 168.49 & 149.43 & 169.75 & 150.45 & 50.53 \\
albu\_rain\_1 & 87.68 & 168.39 & 89.55 & 84.47 & 88.54 & 136.30 & 129.79 & 129.17 & 125.54 & 51.07 \\
albu\_rain\_2 & 141.50 & 217.77 & 147.26 & 133.74 & 140.86 & 162.98 & 138.11 & 162.83 & 162.35 & 50.58 \\
albu\_rain\_3 & 193.56 & 264.45 & 202.80 & 175.78 & 189.00 & 207.61 & 171.61 & 202.60 & 206.21 & 48.13 \\
albu\_sun\_1 & 71.80 & 158.43 & 76.12 & 86.34 & 86.03 & 122.38 & 120.41 & 129.30 & 112.76 & 50.87 \\
albu\_sun\_2 & 70.61 & 153.61 & 74.62 & 80.90 & 81.87 & 117.91 & 116.88 & 129.28 & 111.22 & 50.95 \\
albu\_sun\_3 & 76.07 & 156.86 & 85.11 & 85.52 & 87.88 & 120.45 & 116.00 & 128.29 & 117.18 & 50.84 \\
albu\_sun\_4 & 76.76 & 161.88 & 82.99 & 89.87 & 89.81 & 122.94 & 119.88 & 131.74 & 113.75 & 51.12 \\
albu\_sun\_5 & 77.57 & 162.06 & 82.89 & 85.72 & 86.74 & 123.64 & 116.25 & 130.40 & 115.04 & 50.76 \\
albu\_sun\_6 & 79.95 & 162.36 & 86.61 & 88.60 & 92.94 & 124.52 & 118.07 & 132.04 & 118.64 & 50.71 \\
dense\_fog\_1 & 58.11 & 132.21 & 53.07 & 68.19 & 67.17 & 111.32 & 134.36 & 117.29 & 115.09 & 50.71 \\
dense\_fog\_2 & 65.17 & 136.85 & 62.36 & 71.52 & 72.12 & 112.38 & 136.48 & 122.29 & 116.43 & 50.24 \\
dense\_fog\_3 & 72.46 & 142.73 & 71.15 & 79.29 & 79.66 & 117.36 & 140.14 & 127.50 & 126.05 & 50.47 \\
dense\_fog\_4 & 81.28 & 147.54 & 83.19 & 86.32 & 84.64 & 128.21 & 148.05 & 139.74 & 138.17 & 50.15 \\
dense\_fog\_5 & 109.44 & 173.99 & 109.38 & 110.09 & 107.03 & 155.49 & 171.89 & 171.28 & 165.64 & 49.30 \\
overcast\_1 & 49.75 & 132.07 & 47.25 & 67.35 & 66.41 & 107.77 & 123.18 & 113.14 & 97.17 & 50.84 \\
overcast\_2 & 50.73 & 131.16 & 48.99 & 66.59 & 67.31 & 106.24 & 120.47 & 113.63 & 99.26 & 51.00 \\
overcast\_3 & 51.48 & 133.70 & 48.17 & 67.58 & 65.70 & 108.01 & 122.12 & 114.21 & 99.71 & 51.22 \\
overcast\_4 & 52.51 & 130.86 & 49.47 & 67.42 & 65.64 & 110.92 & 119.75 & 114.08 & 101.03 & 50.91 \\
overcast\_5 & 52.45 & 133.06 & 48.37 & 66.49 & 65.19 & 107.47 & 120.51 & 112.69 & 97.98 & 50.31 \\
puddles\_1 & 53.50 & 133.36 & 49.84 & 67.58 & 64.89 & 111.42 & 122.08 & 113.79 & 100.19 & 50.56 \\
puddles\_2 & 53.99 & 132.78 & 53.57 & 66.38 & 65.13 & 108.32 & 127.78 & 112.77 & 101.44 & 50.62 \\
puddles\_3 & 56.20 & 135.95 & 53.75 & 66.34 & 65.09 & 109.96 & 119.26 & 113.28 & 102.95 & 50.41 \\
puddles\_4 & 56.60 & 138.03 & 54.80 & 63.46 & 64.80 & 111.44 & 118.55 & 115.00 & 104.54 & 50.82 \\
puddles\_5 & 57.39 & 140.20 & 57.08 & 65.75 & 65.77 & 114.09 & 117.90 & 112.99 & 107.53 & 50.84 \\
rain\_composition\_1 & 61.69 & 136.33 & 58.75 & 65.52 & 68.11 & 114.09 & 127.73 & 121.56 & 113.79 & 50.76 \\
rain\_composition\_2 & 74.23 & 145.80 & 71.52 & 70.21 & 76.56 & 123.27 & 126.52 & 126.59 & 124.81 & 50.68 \\
rain\_composition\_3 & 96.24 & 160.61 & 96.01 & 83.74 & 91.61 & 138.19 & 141.55 & 136.29 & 146.46 & 49.86 \\
rain\_composition\_4 & 115.12 & 177.82 & 118.78 & 96.37 & 109.44 & 154.15 & 149.75 & 149.71 & 165.27 & 49.64 \\
rain\_composition\_5 & 159.23 & 216.38 & 160.32 & 131.64 & 151.99 & 189.40 & 181.97 & 191.13 & 204.88 & 48.37 \\
rain\_streaks\_1 & 61.08 & 140.91 & 57.14 & 76.70 & 74.65 & 102.86 & 128.36 & 115.54 & 99.18 & 51.68 \\
rain\_streaks\_2 & 63.44 & 146.06 & 61.56 & 79.06 & 77.47 & 103.47 & 125.96 & 113.57 & 100.38 & 52.09 \\
rain\_streaks\_3 & 66.82 & 148.41 & 67.04 & 82.52 & 79.34 & 106.55 & 121.43 & 112.11 & 100.61 & 51.41 \\
rain\_streaks\_4 & 71.70 & 151.76 & 67.98 & 83.40 & 81.75 & 105.44 & 122.18 & 114.22 & 103.80 & 52.02 \\
rain\_streaks\_5 & 75.08 & 157.62 & 74.69 & 87.02 & 82.93 & 105.12 & 123.27 & 114.34 & 102.24 & 51.62 \\
shadow\_sunglare\_1 & 50.52 & 129.67 & 48.14 & 67.00 & 64.55 & 111.68 & 124.24 & 117.70 & 100.94 & 51.20 \\
shadow\_sunglare\_2 & 50.39 & 132.20 & 49.09 & 65.28 & 65.81 & 111.21 & 123.93 & 119.69 & 99.64 & 50.78 \\
shadow\_sunglare\_3 & 50.89 & 130.49 & 49.72 & 66.14 & 64.91 & 110.23 & 123.44 & 116.61 & 102.78 & 50.93 \\
shadow\_sunglare\_4 & 51.29 & 131.76 & 49.63 & 66.81 & 64.90 & 112.01 & 123.73 & 123.02 & 99.58 & 51.31 \\
shadow\_sunglare\_5 & 51.21 & 134.74 & 49.09 & 66.63 & 65.09 & 112.07 & 124.81 & 121.72 & 102.84 & 50.68 \\
wet\_street\_lens\_droplets\_1 & 52.06 & 133.32 & 48.73 & 67.23 & 66.71 & 111.74 & 122.43 & 117.45 & 102.06 & 51.42 \\
wet\_street\_lens\_droplets\_2 & 53.72 & 135.81 & 54.14 & 65.88 & 65.76 & 110.30 & 119.96 & 116.04 & 106.13 & 51.03 \\
wet\_street\_lens\_droplets\_3 & 55.87 & 139.18 & 54.58 & 66.63 & 65.55 & 114.86 & 119.24 & 115.22 & 105.37 & 51.00 \\
wet\_street\_lens\_droplets\_4 & 68.68 & 141.42 & 67.65 & 62.02 & 71.49 & 122.61 & 121.30 & 119.17 & 119.35 & 50.67 \\
wet\_street\_lens\_droplets\_5 & 78.35 & 149.77 & 71.47 & 69.14 & 82.28 & 126.34 & 131.76 & 124.61 & 126.66 & 51.56 \\
\end{longtable}

\subsection{CMMD}

\begin{longtable}{c|ccccc|cccc|c}
\caption{CMMD distances for various augmentations across different datasets and weather conditions. Additionally, we list semantic segmentation retraining results with respect to \textsc{ACDC} rain data.} \label{tab:cmmd_scores} \\
\toprule
& \multicolumn{5}{c|}{BDD} & \multicolumn{4}{c|}{ACDC} & \multicolumn{1}{c}{Retraining Results} \\
\cline{2-11}
Augmentation & Clear & Fog & Overcast & Rain & Snow & Fog & Rain & Snow & Sun & Rain \\
\midrule
\endfirsthead

\toprule
& \multicolumn{5}{c|}{BDD} & \multicolumn{4}{c|}{ACDC} & \multicolumn{1}{c}{Retraining Results} \\
\cline{2-11}
Augmentation & Clear & Fog & Overcast & Rain & Snow & Fog & Rain & Snow & Sun & Rain \\
\midrule
\endhead

\bottomrule
\endfoot

albu\_fog\_1 & 1.10 & 1.28 & 1.22 & 1.31 & 1.64 & 4.14 & 3.94 & 4.25 & 3.96 & 50.45 \\
albu\_fog\_2 & 1.52 & 1.51 & 1.66 & 1.63 & 2.05 & 4.05 & 4.03 & 4.31 & 4.17 & 50.27 \\
albu\_fog\_3 & 1.97 & 1.85 & 2.12 & 2.02 & 2.51 & 4.10 & 4.21 & 4.46 & 4.42 & 51.32 \\
albu\_fog\_4 & 2.45 & 2.28 & 2.66 & 2.55 & 3.01 & 4.06 & 4.39 & 4.62 & 4.58 & 49.85 \\
albu\_fog\_5 & 3.40 & 2.99 & 3.62 & 3.41 & 3.97 & 4.32 & 4.93 & 5.09 & 5.23 & 50.28 \\
albu\_fog\_6 & 3.98 & 3.55 & 4.23 & 4.02 & 4.59 & 4.64 & 5.31 & 5.44 & 5.61 & 50.53 \\
albu\_rain\_1 & 1.98 & 1.82 & 2.06 & 1.75 & 2.32 & 4.19 & 3.84 & 4.07 & 4.45 & 51.07 \\
albu\_rain\_2 & 3.19 & 2.68 & 3.26 & 2.58 & 3.49 & 4.42 & 4.01 & 4.42 & 5.08 & 50.58 \\
albu\_rain\_3 & 3.68 & 3.14 & 3.79 & 3.06 & 4.00 & 4.59 & 4.27 & 4.69 & 5.35 & 48.13 \\
albu\_sun\_1 & 1.06 & 1.39 & 1.17 & 1.38 & 1.69 & 4.63 & 4.31 & 4.72 & 4.27 & 50.87 \\
albu\_sun\_2 & 1.16 & 1.44 & 1.26 & 1.46 & 1.78 & 4.55 & 4.29 & 4.70 & 4.26 & 50.95 \\
albu\_sun\_3 & 1.27 & 1.50 & 1.38 & 1.56 & 1.90 & 4.52 & 4.31 & 4.73 & 4.29 & 50.84 \\
albu\_sun\_4 & 1.21 & 1.51 & 1.31 & 1.52 & 1.84 & 4.64 & 4.35 & 4.76 & 4.31 & 51.12 \\
albu\_sun\_5 & 1.33 & 1.58 & 1.44 & 1.62 & 1.96 & 4.55 & 4.34 & 4.75 & 4.30 & 50.76 \\
albu\_sun\_6 & 1.45 & 1.64 & 1.57 & 1.73 & 2.09 & 4.51 & 4.34 & 4.76 & 4.32 & 50.71 \\
dense\_fog\_1 & 0.94 & 0.90 & 0.99 & 1.11 & 1.51 & 4.28 & 4.23 & 4.43 & 4.48 & 50.71 \\
dense\_fog\_2 & 1.27 & 0.92 & 1.29 & 1.28 & 1.76 & 4.04 & 4.32 & 4.40 & 4.73 & 50.24 \\
dense\_fog\_3 & 1.57 & 1.03 & 1.59 & 1.49 & 2.02 & 3.96 & 4.45 & 4.45 & 4.97 & 50.47 \\
dense\_fog\_4 & 1.80 & 1.14 & 1.81 & 1.66 & 2.23 & 3.92 & 4.55 & 4.49 & 5.15 & 50.15 \\
dense\_fog\_5 & 2.25 & 1.41 & 2.29 & 2.06 & 2.70 & 3.93 & 4.76 & 4.64 & 5.49 & 49.30 \\
overcast\_1 & 0.73 & 1.14 & 0.79 & 1.06 & 1.34 & 4.88 & 4.35 & 4.59 & 4.37 & 50.84 \\
overcast\_2 & 0.86 & 1.24 & 0.93 & 1.17 & 1.46 & 4.88 & 4.39 & 4.60 & 4.43 & 51.00 \\
overcast\_3 & 0.98 & 1.32 & 1.06 & 1.26 & 1.58 & 4.84 & 4.40 & 4.62 & 4.48 & 51.22 \\
overcast\_4 & 1.04 & 1.35 & 1.10 & 1.31 & 1.66 & 4.82 & 4.33 & 4.56 & 4.45 & 50.91 \\
overcast\_5 & 1.07 & 1.36 & 1.13 & 1.34 & 1.69 & 4.82 & 4.35 & 4.56 & 4.47 & 50.31 \\
puddles\_1 & 1.12 & 1.33 & 1.18 & 1.31 & 1.69 & 4.65 & 4.25 & 4.51 & 4.43 & 50.56 \\
puddles\_2 & 1.13 & 1.33 & 1.20 & 1.31 & 1.71 & 4.62 & 4.21 & 4.49 & 4.41 & 50.62 \\
puddles\_3 & 1.15 & 1.33 & 1.23 & 1.32 & 1.73 & 4.58 & 4.17 & 4.46 & 4.39 & 50.41 \\
puddles\_4 & 1.17 & 1.33 & 1.24 & 1.32 & 1.75 & 4.55 & 4.14 & 4.44 & 4.38 & 50.82 \\
puddles\_5 & 1.17 & 1.32 & 1.24 & 1.32 & 1.74 & 4.53 & 4.11 & 4.42 & 4.35 & 50.84 \\
rain\_composition\_1 & 3.19 & 2.57 & 3.21 & 2.40 & 3.34 & 3.87 & 3.73 & 3.93 & 4.91 & 50.76 \\
rain\_composition\_2 & 3.68 & 3.14 & 3.79 & 3.06 & 4.00 & 4.59 & 4.27 & 4.69 & 5.35 & 50.68 \\
rain\_composition\_3 & 4.00 & 3.34 & 4.12 & 3.41 & 4.47 & 4.74 & 4.41 & 4.78 & 5.38 & 49.86 \\
rain\_composition\_4 & 4.12 & 3.47 & 4.24 & 3.55 & 4.61 & 4.76 & 4.44 & 4.81 & 5.41 & 49.64 \\
rain\_composition\_5 & 4.18 & 3.62 & 4.34 & 3.68 & 4.74 & 4.78 & 4.48 & 4.83 & 5.44 & 48.37 \\
rain\_streaks\_1 & 2.65 & 2.29 & 2.71 & 2.31 & 2.95 & 3.85 & 3.90 & 4.02 & 4.53 & 51.68 \\
rain\_streaks\_2 & 2.90 & 2.47 & 2.96 & 2.48 & 3.15 & 3.88 & 3.98 & 4.06 & 4.69 & 52.09 \\
rain\_streaks\_3 & 3.12 & 2.62 & 3.15 & 2.64 & 3.34 & 3.94 & 4.07 & 4.11 & 4.84 & 51.41 \\
rain\_streaks\_4 & 3.29 & 2.74 & 3.34 & 2.78 & 3.49 & 3.97 & 4.14 & 4.15 & 4.97 & 52.02 \\
rain\_streaks\_5 & 3.47 & 2.86 & 3.51 & 2.90 & 3.65 & 4.00 & 4.22 & 4.19 & 5.10 & 51.62 \\
shadow\_sunglare\_1 & 0.20 & 0.65 & 0.25 & 0.53 & 0.79 & 4.73 & 4.23 & 4.57 & 4.29 & 51.20 \\
shadow\_sunglare\_2 & 0.19 & 0.66 & 0.25 & 0.52 & 0.78 & 4.74 & 4.20 & 4.55 & 4.26 & 50.78 \\
shadow\_sunglare\_3 & 0.20 & 0.68 & 0.27 & 0.54 & 0.79 & 4.74 & 4.17 & 4.54 & 4.22 & 50.93 \\
shadow\_sunglare\_4 & 0.25 & 0.73 & 0.33 & 0.59 & 0.83 & 4.69 & 4.14 & 4.50 & 4.18 & 51.31 \\
shadow\_sunglare\_5 & 0.32 & 0.78 & 0.41 & 0.66 & 0.90 & 4.62 & 4.09 & 4.45 & 4.13 & 50.68 \\
wet\_street\_lens\_droplets\_1 & 1.08 & 0.97 & 1.13 & 1.17 & 1.63 & 4.09 & 4.03 & 4.29 & 4.34 & 51.42 \\
wet\_street\_lens\_droplets\_2 & 1.57 & 1.10 & 1.60 & 1.42 & 2.03 & 3.79 & 4.02 & 4.23 & 4.61 & 51.03 \\
wet\_street\_lens\_droplets\_3 & 2.10 & 1.43 & 2.15 & 1.87 & 2.55 & 3.70 & 4.19 & 4.35 & 4.88 & 51.00 \\
wet\_street\_lens\_droplets\_4 & 2.50 & 1.67 & 2.56 & 2.15 & 2.93 & 3.64 & 4.24 & 4.39 & 5.09 & 50.67 \\
wet\_street\_lens\_droplets\_5 & 3.17 & 2.19 & 3.27 & 2.80 & 3.64 & 3.72 & 4.54 & 4.66 & 5.49 & 51.56 \\
\end{longtable}

\subsection{Contrastive-FID/CMMD on ACDC}

\begin{longtable}{c|cccc|cccc|cccc}
\caption{C-FID/C-CMMD scores on \textsc{ACDC}, as well as class predictions of the multi-weather classifier on every augmentation set.} \label{tab:contrastive_scores} \\
\toprule
& \multicolumn{4}{c|}{C-FID} & \multicolumn{4}{c|}{C-CMMD} & \multicolumn{4}{c}{Classification Results (Weather Classifier)} \\
\cline{2-13}
Augmentation & Fog & Rain & Snow & Sun & Fog & Rain & Snow & Sun & Fog & Rain & Snow & Sun \\
\midrule
\endfirsthead

\toprule
& \multicolumn{4}{c|}{C-FID} & \multicolumn{4}{c|}{C-CMMD} & \multicolumn{4}{c}{Classification Results (Weather Classifier)} \\
\cline{2-13}
Augmentation & Fog & Rain & Snow & Sun & Fog & Rain & Snow & Sun & Fog & Rain & Snow & Sun \\
\midrule
\endhead

\bottomrule
\endfoot

albu\_fog\_1 & -0.26 & 0.41 & -0.01 & -0.08 & -0.07 & 0.14 & -0.17 & 0.11 & 5 & 114 & 78 & 603 \\
albu\_fog\_2 & -0.05 & 0.09 & -0.07 & 0.03 & 0.09 & 0.11 & -0.16 & -0.03 & 14 & 140 & 72 & 574 \\
albu\_fog\_3 & -0.16 & 0.27 & -0.02 & -0.07 & 0.20 & 0.08 & -0.15 & -0.11 & 16 & 192 & 58 & 534 \\
albu\_fog\_4 & 0.00 & -0.07 & -0.13 & 0.21 & 0.34 & 0.02 & -0.18 & -0.14 & 199 & 74 & 126 & 401 \\
albu\_fog\_5 & -0.12 & 0.03 & -0.14 & 0.24 & 0.53 & -0.03 & -0.15 & -0.26 & 317 & 124 & 63 & 296 \\
albu\_fog\_6 & -0.21 & 0.27 & -0.24 & 0.24 & 0.53 & -0.05 & -0.14 & -0.25 & 348 & 138 & 36 & 278 \\
albu\_rain\_1 & -0.18 & 0.01 & 0.03 & 0.15 & -0.05 & 0.31 & 0.07 & -0.28 & 25 & 180 & 108 & 487 \\
albu\_rain\_2 & -0.16 & 0.53 & -0.15 & -0.14 & 0.06 & 0.47 & 0.05 & -0.47 & 16 & 320 & 75 & 389 \\
albu\_rain\_3 & -0.20 & 0.59 & -0.11 & -0.18 & 0.11 & 0.43 & 0.03 & -0.47 & 27 & 418 & 38 & 317 \\
albu\_sun\_1 & -0.04 & 0.03 & -0.25 & 0.30 & -0.13 & 0.16 & -0.20 & 0.20 & 1 & 190 & 59 & 550 \\
albu\_sun\_2 & 0.03 & 0.07 & -0.32 & 0.27 & -0.09 & 0.15 & -0.21 & 0.18 & 0 & 172 & 55 & 573 \\
albu\_sun\_3 & 0.00 & 0.15 & -0.24 & 0.11 & -0.05 & 0.14 & -0.23 & 0.16 & 3 & 168 & 65 & 564 \\
albu\_sun\_4 & -0.03 & 0.07 & -0.29 & 0.29 & -0.10 & 0.15 & -0.21 & 0.19 & 0 & 205 & 57 & 538 \\
albu\_sun\_5 & -0.07 & 0.17 & -0.28 & 0.22 & -0.06 & 0.14 & -0.22 & 0.18 & 0 & 207 & 40 & 553 \\
albu\_sun\_6 & -0.04 & 0.18 & -0.26 & 0.16 & -0.02 & 0.13 & -0.23 & 0.15 & 2 & 165 & 57 & 576 \\
dense\_fog\_1 & 0.29 & -0.44 & 0.08 & 0.15 & 0.07 & 0.12 & -0.07 & -0.11 & 289 & 67 & 364 & 80 \\
dense\_fog\_2 & 0.34 & -0.43 & -0.01 & 0.19 & 0.33 & 0.05 & -0.03 & -0.30 & 570 & 12 & 206 & 12 \\
dense\_fog\_3 & 0.35 & -0.35 & 0.01 & 0.05 & 0.50 & 0.01 & 0.01 & -0.41 & 726 & 2 & 72 & 0 \\
dense\_fog\_4 & 0.32 & -0.26 & -0.03 & 0.01 & 0.62 & -0.02 & 0.03 & -0.48 & 766 & 2 & 31 & 1 \\
dense\_fog\_5 & 0.27 & -0.14 & -0.12 & 0.01 & 0.79 & -0.05 & 0.06 & -0.57 & 797 & 0 & 3 & 0 \\
overcast\_1 & 0.09 & -0.42 & -0.10 & 0.54 & -0.27 & 0.18 & -0.04 & 0.16 & 44 & 158 & 293 & 305 \\
overcast\_2 & 0.14 & -0.35 & -0.13 & 0.43 & -0.25 & 0.17 & -0.02 & 0.13 & 45 & 153 & 317 & 285 \\
overcast\_3 & 0.11 & -0.36 & -0.11 & 0.45 & -0.21 & 0.17 & -0.03 & 0.09 & 40 & 146 & 347 & 267 \\
overcast\_4 & 0.02 & -0.28 & -0.09 & 0.41 & -0.23 & 0.19 & -0.02 & 0.08 & 49 & 158 & 419 & 174 \\
overcast\_5 & 0.08 & -0.36 & -0.11 & 0.48 & -0.22 & 0.18 & -0.01 & 0.07 & 53 & 160 & 432 & 155 \\
puddles\_1 & 0.02 & -0.33 & -0.07 & 0.47 & -0.16 & 0.20 & -0.04 & 0.03 & 70 & 161 & 401 & 168 \\
puddles\_2 & 0.16 & -0.48 & -0.01 & 0.44 & -0.16 & 0.21 & -0.05 & 0.02 & 76 & 175 & 389 & 160 \\
puddles\_3 & 0.05 & -0.26 & -0.07 & 0.33 & -0.16 & 0.22 & -0.05 & 0.01 & 75 & 191 & 377 & 157 \\
puddles\_4 & 0.03 & -0.21 & -0.09 & 0.30 & -0.15 & 0.23 & -0.06 & 0.00 & 75 & 200 & 376 & 149 \\
puddles\_5 & -0.03 & -0.16 & 0.00 & 0.21 & -0.16 & 0.24 & -0.06 & 0.00 & 87 & 207 & 356 & 150 \\
rain\_composition\_1 & 0.18 & -0.26 & -0.07 & 0.19 & 0.10 & 0.16 & -0.10 & -0.14 & 371 & 83 & 297 & 49 \\
rain\_composition\_2 & 0.07 & -0.04 & -0.04 & 0.02 & 0.39 & 0.14 & -0.06 & -0.39 & 656 & 15 & 125 & 4 \\
rain\_composition\_3 & 0.07 & -0.03 & 0.13 & -0.16 & 0.63 & 0.09 & -0.06 & -0.49 & 765 & 1 & 33 & 1 \\
rain\_composition\_4 & 0.01 & 0.13 & 0.13 & -0.26 & 0.77 & 0.09 & -0.05 & -0.59 & 789 & 1 & 10 & 0 \\
rain\_composition\_5 & 0.05 & 0.22 & 0.01 & -0.25 & 0.95 & 0.06 & -0.05 & -0.65 & 798 & 0 & 2 & 0 \\
rain\_streaks\_1 & 0.34 & -0.53 & -0.14 & 0.50 & 0.23 & 0.18 & 0.05 & -0.40 & 240 & 75 & 415 & 70 \\
rain\_streaks\_2 & 0.29 & -0.48 & -0.10 & 0.42 & 0.28 & 0.17 & 0.09 & -0.46 & 309 & 54 & 399 & 38 \\
rain\_streaks\_3 & 0.14 & -0.37 & -0.07 & 0.38 & 0.30 & 0.17 & 0.13 & -0.50 & 382 & 40 & 352 & 26 \\
rain\_streaks\_4 & 0.23 & -0.35 & -0.10 & 0.29 & 0.34 & 0.16 & 0.15 & -0.53 & 479 & 25 & 280 & 16 \\
rain\_streaks\_5 & 0.23 & -0.39 & -0.11 & 0.35 & 0.38 & 0.15 & 0.18 & -0.57 & 567 & 11 & 216 & 6 \\
shadow\_sunglare\_1 & 0.07 & -0.34 & -0.14 & 0.50 & -0.23 & 0.21 & -0.10 & 0.15 & 8 & 163 & 104 & 525 \\
shadow\_sunglare\_2 & 0.09 & -0.33 & -0.20 & 0.56 & -0.26 & 0.23 & -0.10 & 0.17 & 6 & 157 & 93 & 544 \\
shadow\_sunglare\_3 & 0.11 & -0.33 & -0.11 & 0.41 & -0.27 & 0.24 & -0.11 & 0.19 & 3 & 153 & 84 & 560 \\
shadow\_sunglare\_4 & 0.09 & -0.30 & -0.27 & 0.60 & -0.27 & 0.23 & -0.11 & 0.19 & 2 & 139 & 80 & 579 \\
shadow\_sunglare\_5 & 0.12 & -0.30 & -0.21 & 0.49 & -0.26 & 0.23 & -0.11 & 0.19 & 1 & 124 & 78 & 597 \\
wet\_street\_lens\_droplets\_1 & 0.06 & -0.29 & -0.14 & 0.45 & -0.27 & 0.24 & -0.08 & 0.15 & 35 & 216 & 260 & 289 \\
wet\_street\_lens\_droplets\_2 & 0.10 & -0.23 & -0.10 & 0.26 & -0.24 & 0.26 & -0.09 & 0.10 & 36 & 239 & 234 & 291 \\
wet\_street\_lens\_droplets\_3 & -0.04 & -0.19 & -0.05 & 0.32 & -0.21 & 0.28 & -0.10 & 0.06 & 38 & 259 & 229 & 274 \\
wet\_street\_lens\_droplets\_4 & -0.07 & -0.02 & 0.05 & 0.04 & -0.10 & 0.51 & 0.02 & -0.33 & 26 & 385 & 147 & 242 \\
wet\_street\_lens\_droplets\_5 & 0.03 & -0.13 & 0.09 & 0.02 & 0.25 & 0.41 & 0.18 & -0.65 & 117 & 417 & 201 & 65 \\
\end{longtable}

\end{landscape}
\clearpage 

\end{document}


\onecolumn

\appendix
\section{Results for \textsc{mnist}} \label{sec: A}

\begin{figure*}[h]
\begin{center}
\centerline{%
      \includegraphics[width=0.5\textwidth]{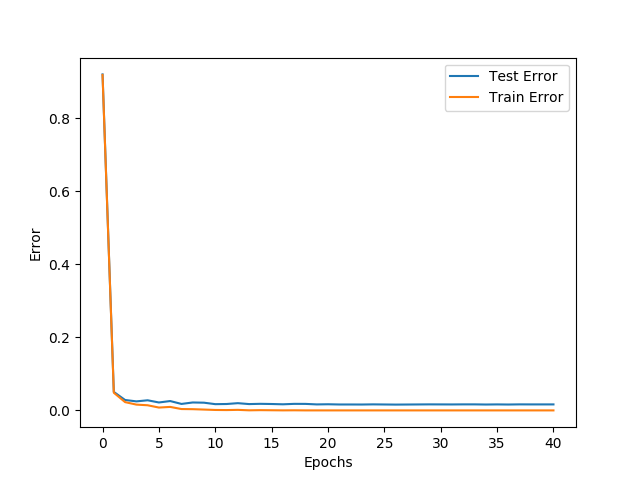}%
    \includegraphics[width=0.5\textwidth]{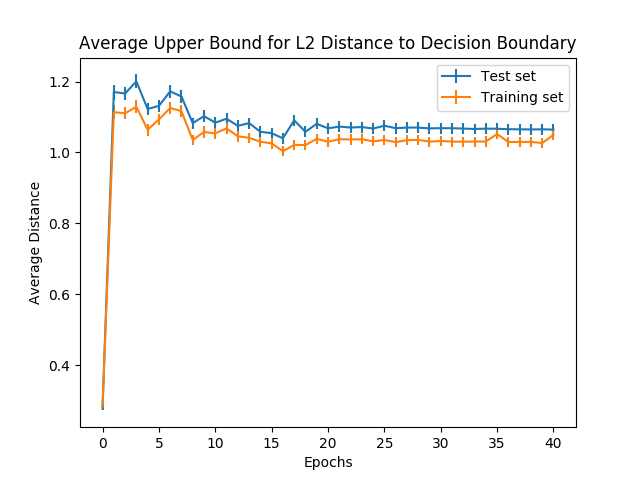}%
      }%
\end{center}
   \caption{In (a) on the left, we observe that the train and test error both converge quickly to almost $0$. \newline
   In (b) we see that even though there is little room for improvement in the test error, the upper bound to the 
   distance to the decision boundary is decreasing in a small but visible 
   fashion after the first epoch supporting observation 1.} 
 \label{fig: mnist_results_vanilla}
\end{figure*}

\begin{figure*}[h]
\begin{center}
\centerline{%
      \includegraphics[width=0.5\textwidth]{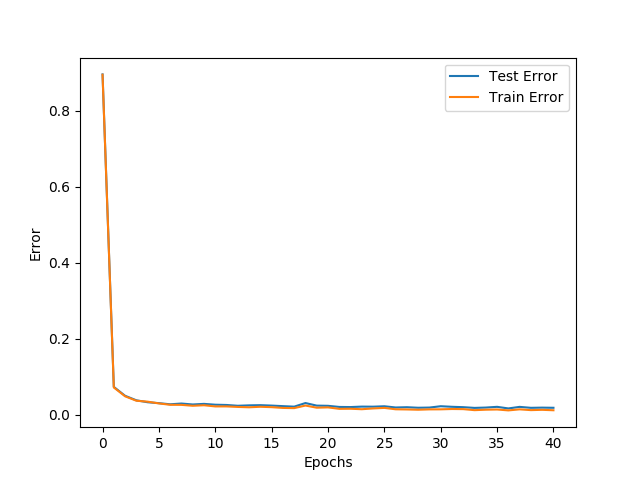}%
    \includegraphics[width=0.5\textwidth]{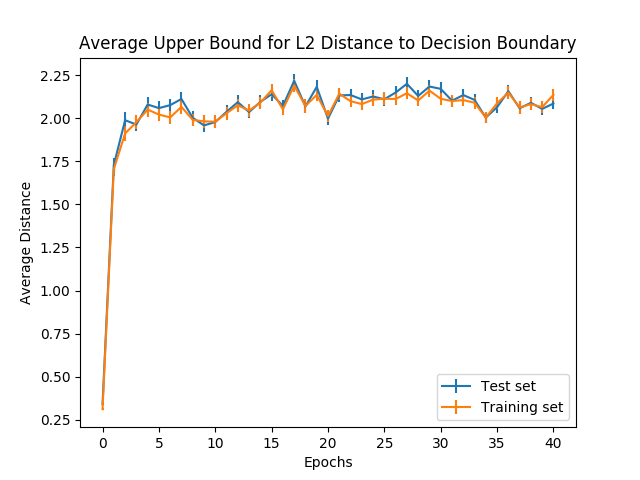}%
      }%
\end{center}
   \caption{(a) The train and test errors also converge quickly to near optimal performance for adversarial training on \textsc{mnist}. \newline
   (b) $d_2^{avg}$ is rather increasing during adversarial training and nearly twice as high as for standard training which is evidence for observation 2.} 
 \label{fig: mnist_results_adversarial_1}
\end{figure*}

\begin{figure*}[h]
\begin{center}
\centerline{%
      \includegraphics[width=0.5\textwidth]{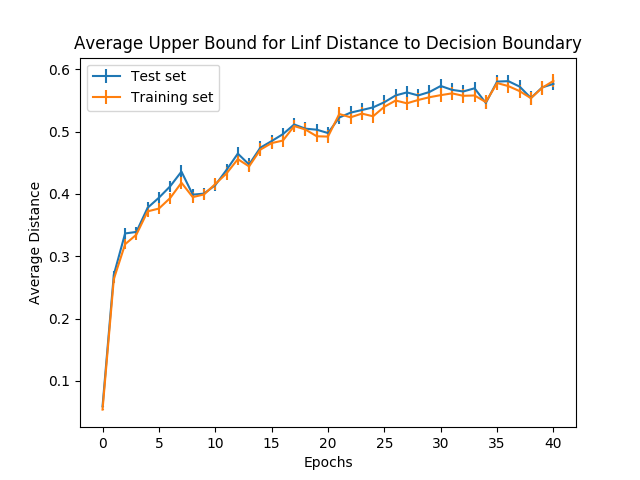}%
      }%
\end{center}
   \caption{$d_{\infty}^{avg}$ is steadily increasing for PGD adversarial training.} 
 \label{fig: mnist_results_adversarial_2}
\end{figure*}

\begin{figure*}[h]
\begin{center}
\centerline{%
      \includegraphics[width=0.5\textwidth]{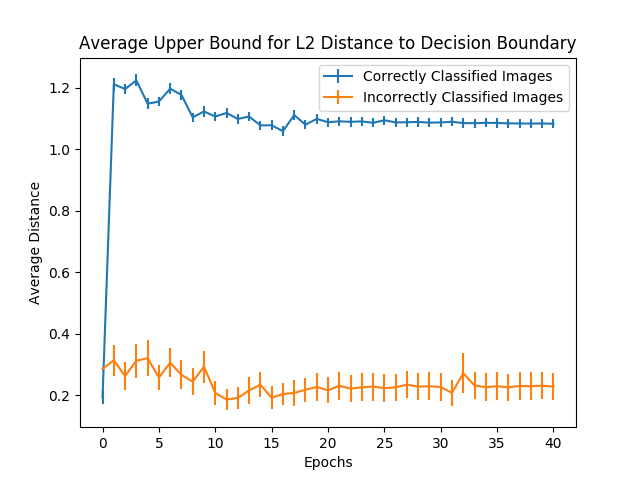}%
    \includegraphics[width=0.5\textwidth]{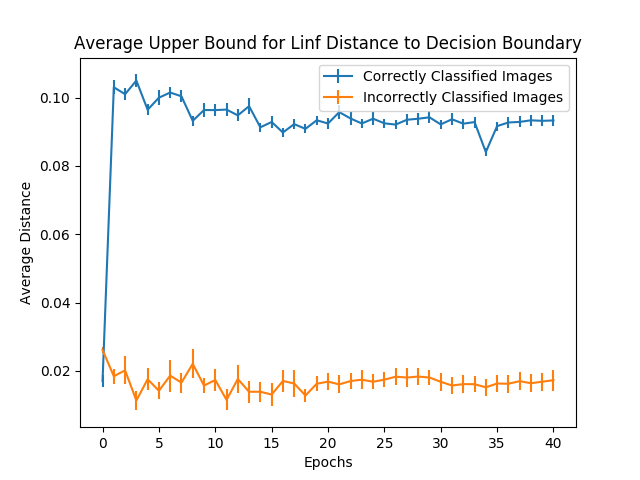}%
      }%
\end{center}
   \caption{(a) and (b): For a dense network on \textsc{Fashion Mnist} with standard training, 
   $d_2^{avg}$ and $d_{\infty}^{avg}$ are much higher for correctly classified data points than for incorrectly classified ones supporting observation 3.} 
 \label{fig: mnist_results_correct_incorrect}
\end{figure*}

For \textsc{mnist} we used a dense network architecture in contrast to the \textsc{cnn} architecture which we used for \textsc{fashion-mnist}.
In Figure~\ref{fig: mnist_results_vanilla} we see that observation 1 is supported even though the network quickly achieves near optimal test error.
Figure~\ref{fig: mnist_results_adversarial_1} supports observation 2 showing that $d_2^{avg}$ is nearly twice as high for adversarial training 
as for standard training. In Figure~\ref{fig: mnist_results_adversarial_2} we even see a steady increase for $d_{\infty}^{avg}$ during 
adversarial training.
Finally, Figure~\ref{fig: mnist_results_correct_incorrect} is evidence for observation 3. Notice that the standard error generally is larger for incorrectly classified 
samples than for correctly classified ones, since there are far fewer of them due to the small error on train and test set.

\begin{figure*}[h]
\begin{center}
\centerline{%
   \includegraphics[width=0.35 \textwidth]{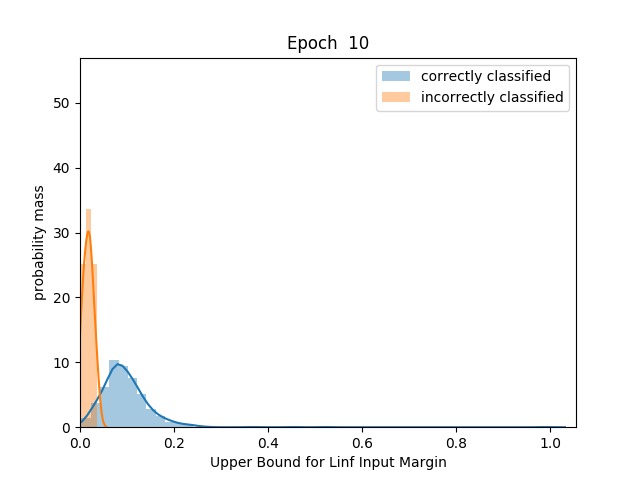}%
    \includegraphics[width=0.35 \textwidth]{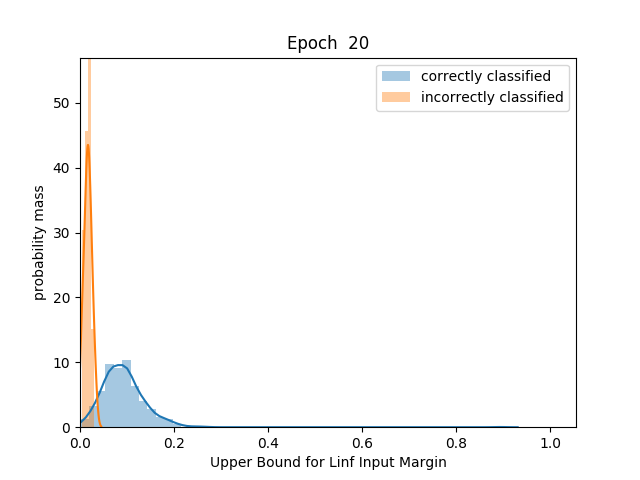}%
    \includegraphics[width=0.35 \textwidth]{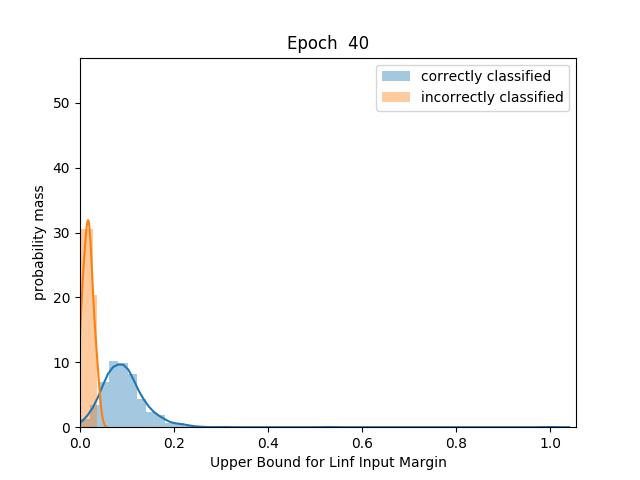}%
    }%
\centerline{%
\includegraphics[width=0.35 \textwidth]{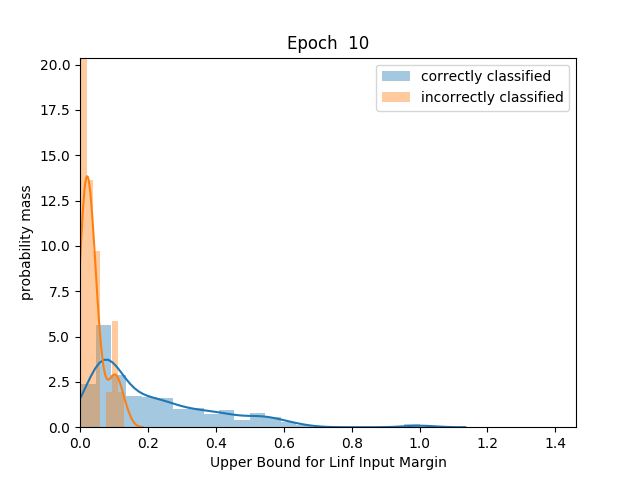}%
\includegraphics[width=0.35 \textwidth]{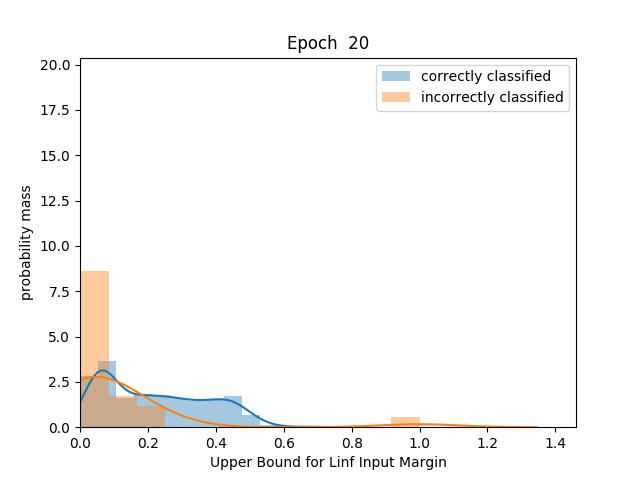}%
\includegraphics[width=0.35 \textwidth]{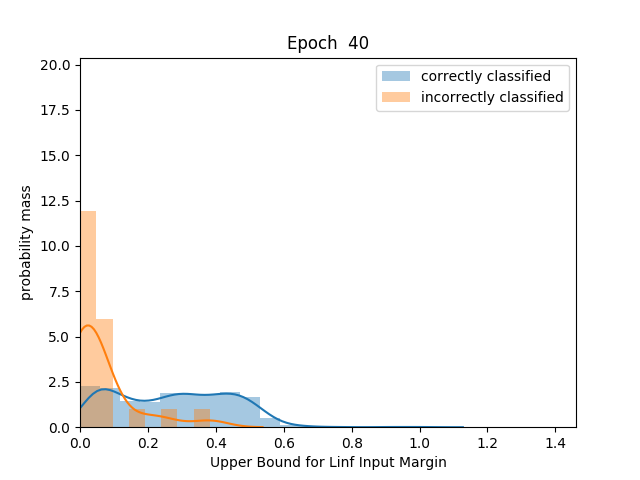}%
}%
\end{center}
   \caption{Sample $d_{\infty}^{avg}$ distributions for $1000$ training images and the 
   training epochs $10, 20$ and $40$: (1) First row: $d_{\infty}^{avg}(x)$ distribution for the 
   classically trained dense network on \textsc{mnist}; (2) Second row: $d_{\infty}(x)^{avg}$ 
   distribution for the adversarially trained dense network on \textsc{mnist}.} 
 \label{fig: mnist_distributions}
\end{figure*}
\clearpage

\section{Work in Progress Notice}
This document is still work in progress and we are planning to release a discussion of the network 
architectures as well the results for \textsc{Cifar-10} in a future version.